\documentclass{article}

\usepackage[preprint]{preprint}

\usepackage[utf8]{inputenc} 
\usepackage[T1]{fontenc}    
\usepackage{hyperref}       
\usepackage{url}            
\usepackage{booktabs}       
\usepackage{amsfonts}       
\usepackage{nicefrac}       
\usepackage{microtype}      
\usepackage{xcolor}         

\usepackage{graphicx}
\usepackage{subcaption}
\usepackage{natbib}
\usepackage{tikz}
\usepackage{siunitx}
\usepackage{soul}
\usepackage{balance}
\usepackage{titletoc}

\usepackage{amsthm,amsfonts,amssymb,amsmath,bbm}

\usepackage[capitalize,noabbrev]{cleveref}
\crefname{equation}{}{} 
\renewcommand{\paragraph}[1]{\textbf{#1}~}

\hypersetup{
    colorlinks=true,
    linkcolor={red!50!black},
    citecolor={blue!50!black},
    urlcolor={blue!80!black}
}

\newcommand{\task}[1][]{\mathbf{t}_{#1}}
\newcommand{\taskdist}[1][]{\mathcal{T}_{#1}}
\newcommand{\dMMSE}[1][]{dMMSE$_{#1}$}
\newcommand{\q}[2]{q_{#1}(#2)}
\newcommand{\tcrit}[1][]{t^{\crit}_{#1}}

\newcommand{\crit}{\operatorname{crit}}

\newcommand\papertitle{%
    Dynamics of Transient Structure in In-Context Linear Regression Transformers
}

\icmltitlerunning{\papertitle}
\begin{document}

\twocolumn[
\icmltitle{\papertitle}

\begin{icmlauthorlist}
\icmlauthor{Liam Carroll}{timaeus,gradient}
\icmlauthor{Jesse Hoogland}{timaeus}
\icmlauthor{Matthew Farrugia-Roberts}{oxford}
\icmlauthor{Daniel Murfet}{uom}
\end{icmlauthorlist}

\icmlaffiliation{timaeus}{Timaeus}
\icmlaffiliation{gradient}{Gradient Institute}
\icmlaffiliation{oxford}{Department of Computer Science, University of Oxford}
\icmlaffiliation{uom}{School of Mathematics and Statistics, the University of Melbourne}

\icmlcorrespondingauthor{Liam Carroll}{lemmykc@gmail.com}

\icmlkeywords{Science of deep learning, training dynamics, transient structure, singular learning theory}

\vskip 0.3in
]

\printAffiliationsAndNotice{}

\begin{abstract}
Modern deep neural networks display striking examples of rich internal computational structure. Uncovering principles governing the development of such structure is a priority for the science of deep learning. In this paper, we explore the transient ridge phenomenon: when transformers are trained on in-context linear regression tasks with intermediate task diversity, they initially behave like ridge regression before specializing to the tasks in their training distribution. This transition from a general solution to a specialized solution is revealed by joint trajectory principal component analysis. Further, we draw on the theory of Bayesian internal model selection to suggest a general explanation for the phenomena of transient structure in transformers, based on an evolving tradeoff between loss and complexity. We empirically validate this explanation by measuring the model complexity of our transformers as defined by the local learning coefficient.
\end{abstract}

\section{Introduction}

Why do neural networks transition between qualitatively different modes of computation during training?
This phenomenon has been studied for decades in artificial and biological neural networks \citep{baldi1989neural,rogers2004semantic}.
Recent work on transformers has uncovered particularly salient examples of transitions between two well-characterized alternative ways of approximating the data distribution.
For instance, \citet{power2022grokking} show a ``grokking'' transition from an initial \emph{memorizing} solution to a \emph{generalizing} solution while training transformers to perform modular arithmetic.
Conversely, \citet{singh2024transient} show a transition from a ``transient'' generalizing solution to a memorizing solution while training transformers for in-context classification.

In this paper, we study a similar transition from generalization to memorization in transformers trained for in-context linear regression.
Following \citet{raventós2023pretraining}, we construct sequences with latent regression vectors (\emph{tasks}) sampled uniformly from a fixed set of size $M$ (the \emph{task diversity}).
In this setting, \citet{raventós2023pretraining} showed that fully trained transformers may behaviorally approximate either of two distinct in-context learning algorithms:
\begin{enumerate}
    \item \emph{Discrete minimum mean squared error (dMMSE):}
        The posterior mean given a uniform prior over the $M$ tasks
        (implies memorizing the $M$ tasks in some fashion).
    \item \emph{Ridge regression (ridge):}
        The posterior mean given a Gaussian prior from which the fixed tasks were initially sampled (independent of $M$, generalizes to new tasks).
\end{enumerate}
Moreover, \citet[\S6.1]{panwar2024bayesianprism} showed that for intermediate $M$ values, the out-of-distribution loss of a given transformer is non-monotonic, suggesting that these transformers initially approach ridge before diverting towards dMMSE. We term this phenomenon \emph{transient ridge}.

In this paper, we extend the brief analysis of \citet[\S6.1]{panwar2024bayesianprism} and investigate the dynamics of transient ridge in detail, contributing the following.
\begin{itemize}
    \item 
        In \cref{section:pca}, we replicate transient ridge and we comparatively analyze the in-distribution function-space trajectories of our transformers using \emph{joint trajectory principal component analysis}, revealing generalization--memorization as a principal axis of development and clarifying how the task diversity affects the dynamics.
    \item
        In \cref{section:slt}, we explain transient ridge as the transformer navigating \emph{a tradeoff between loss and complexity that evolves over training}, akin to Bayesian internal model selection \citetext{\citealp[\S7.6]{greybook}; \citealp{chen2023tms1}}, and we validate this explanation by estimating the complexity of the competing solutions using the local learning coefficient \citep{quantifdegen}.
\end{itemize}
These results expand our understanding of the transient ridge phenomenon and highlight the evolving loss/complexity tradeoff as a promising principle for understanding similar transience phenomena.
\Cref{section:discussion} discusses limitations and directions for further investigation.

\section{Related work}\label{section:related-work}

In this section, we review empirical and theoretical work on the topic of the emergence and transience of computational structure in deep learning.

\paragraph{Internal computational structure.}
Modern deep learning has shown striking examples of the emergence of internal computational structure in deep neural networks, such as
    syntax trees in transformers trained on natural language \citep{hewitt-manning-2019-structural},
    conceptual chess knowledge in AlphaZero \citep{mcgrath2022acquisition},
    and various results from mechanistic interpretability \citep[e.g.,][]{olah2020zoom,cammarata2020thread,elhage2021mathematical}.

It is known that properties of the data distribution influence the emergence of computational structure.
For example, \citet{chan2022data} studied an in-context classification and identified data properties that are necessary for transformers to develop in-context learning abilities.
\citet{raventós2023pretraining} studied in-context linear regression \citep{garg2022what,akyurek2023learningalgorithmincontextlearning,von2023transformers, bai2024transformers} and showed that changing the \emph{task diversity} of the training distribution can change the in-context learning algorithm approximated by the fully-trained transformer.

\paragraph{Transient structure.}
In some cases, \emph{multiple} interesting computational structures emerge throughout training, with different ones determining model outputs at different times.
A well-known example is the ``grokking'' transition, in which transformers learning modular arithmetic initially memorize the mappings from the training set, before eventually generalizing to unseen examples \citep{power2022grokking} using an internal addition algorithm \citep{nanda2023progress}.

Conversely, \citet{singh2024transient} showed that transformers trained for in-context classification \citep{chan2022data} can gradually shift from predicting based on contextual examples to predicting memorized labels, losing the ability to generalize to new mappings.
\citet{singh2024transient} termed this phenomenon ``transient in-context learning.''

Similarly, for in-context linear regression, \citet[\S 6.1]{panwar2024bayesianprism} observed transformers initially achieving low out-of-distribution generalization loss (indicating that they approximate ridge regression) before eventually specializing to a memorized set of tasks.
In an attempt to unify terminology, we call this phenomenon ``transient ridge.''
Compared to \citet[\S6.1]{panwar2024bayesianprism}, our work is novel in that it offers a more in-depth empirical analysis of this phenomenon, and we also offer an explanation of the phenomenon.

Additional examples of ``transient structure'' have recently been observed in settings including
    language modeling \citep{icl1},
    in-context Markovian sequence modeling \citep{edelman2024statisticalinductionheads,park2024competition}, and 
    in-context modular arithmetic \citep{he2024learningtogrok}.

\paragraph{Explaining transient in-context learning.}
There have been attempts to explain transience in the in-context classification setting originally studied by \citet{singh2024transient}.
\citet{nguyen2024differential} offer a simplified model in which in-context learning is acquired more rapidly than in-weight learning, and targeted regularization of the induction mechanism can cause it to later give way to in-weight learning.

\citet{chan2024understanding} give an explanation based on regret bounds for in-context and in-weight learning.
In their model, in-context learning emerges because it is initially more accurate than in-weight learning for rare classes. 
Once the model sees more data for a class, in-weight learning becomes more accurate than in-context learning, due to limitations in their proposed induction mechanism.

Compared to these models, we offer a higher-level explanation of the general phenomenon of transient structure in terms of principles governing the preference for one solution over another at different points in training.
We study this explanation in the setting of in-context linear regression, but it is also applicable in other settings.

\paragraph{Explaining transient structure.}
There have been several attempts to explain transient structure in more general terms.
If the memorizing solution achieves lower loss than the transient generalizing solution, the ultimate preference for memorization is not surprising \citep{singh2024transient,park2024competition}.
The question remains, why would a generalizing solution arise in the first place if it is not as accurate as the memorizing solution \citep{singh2024transient}?

\citet{panwar2024bayesianprism} speculate that the initial emergence of the generalizing solution could be due to an inductive bias towards \emph{simplicity}. However, this still leaves the question, given that a less-accurate generalizing solution does emerge, why would it then fade later in training \citep{singh2024transient}?

Our work integrates these two perspectives. Rather than prioritizing accuracy \emph{or} simplicity, we postulate a tradeoff between accuracy and simplicity that \emph{evolves over training}.
This explains the emergence of a simpler, less accurate generalizing solution (ridge) \emph{and} its subsequent replacement by a complex, more accurate memorizing solution (dMMSE).

\paragraph{Internal model selection in deep learning.}
Recent work has studied the relevance of \emph{internal model selection} in Bayesian inference to deep learning.
\citet{chen2023tms1} showed that, when small autoencoders transition between different encoding schemes during training \citep{elhage2022superposition}, such transitions are consistent with Bayesian inference.
\citet{icl1} and \citet{lang1} found that the same theory can be used to detect the formation of internal structure, such as induction circuits \citep{elhage2021mathematical,olsson2022context} in small language models.
Ours is the first work to analyze a transition between two transformer solutions in detail from this perspective.

\section{In-context linear regression}\label{section:setting}

In this section, we introduce the in-context linear regression setting and the idealized dMMSE and ridge solutions, largely following \citet{raventós2023pretraining}.

\subsection{Nested multi-task data distributions}
\label{section:setting:data}

Given a latent regression vector, or \emph{task}, $\task \in \mathbb{R}^D$, we define a conditional distribution $q(S|\task)$ of sequences of i.i.d.\ pairs
    \begin{equation*}
    S
    = (x_1, y_1, \ldots, x_K, y_K)
    \in (\mathbb{R}^D \times \mathbb{R})^K
    \end{equation*}
where $x_k \sim q(x) = \mathcal{N}(0,I_D)$
and $y_k \sim q(y | x_k, \task) = \mathcal{N}(\task^\top x_k, \sigma^2)$.
We set $K=16$, $D=8$, and $\sigma^2 = 0.125$.

We then define an unconditional \emph{data distribution} of sequences
    $q(S) = q(S|\task)q(\task)$,
where $q(\task)$ is one of several \emph{task distributions} described below.
We sample a dataset of size $N$,
    $\mathcal{D} = \{S^i\}_{i=1}^N \sim q(S)$,
by first sampling $\task^i \sim q(\task)$
and then sampling $S^i \sim q(S|\task^i)$
for $i=1,\ldots,N$.

We define a task distribution $\q{M}{\task}$ for each \emph{task diversity} $M \in \mathbb{N} \cup \{\infty\}$ as follows.
We fix an unbounded i.i.d.\ sequence
    $\task[1], \task[2], \ldots \sim \mathcal{N}(0,I_D)$.
For $M \in \mathbb{N}$ we define
\begin{equation*}
\taskdist[M] = \{\task[1], \dots, \task[M]\}
\quad\text{and}\quad
\q{M}{\task} = \mathrm{Uniform}(\taskdist[M]).
\end{equation*}
We further define
    $\taskdist[\infty] = \{\task[1], \ldots\}$
and
    $\q{\infty}{\task} = \mathcal{N}(0,I_D)$.
We denote by $\q{M}{S}$ the data distribution formed from $\q{M}{\task}$, and by $\mathcal{D}^{(M)}$ a corresponding dataset.

Note that, in a departure from \citet{raventós2023pretraining}, the task sets $\taskdist[1] \subseteq \taskdist[2] \subseteq \cdots \subseteq \taskdist[\infty]$ are nested by construction.
In particular, the \emph{root task} $\task[1]$ is included at every $M$, allowing us to compare all models by their behavior on $q(S|\task[1])$.

\subsection{Mean squared error objective}
\label{section:setting:loss}

Given a sequence $S$, denote by $S_{\leq k}$ the \emph{context} subsequence $(x_1, y_1, \ldots,x_{k-1}, y_{k-1}, x_k)$ with \emph{label} $y_k$.
Let $f$ be a function mapping contexts to predicted labels.
Given a dataset $\{S^i\}_{i=1}^N \sim q(S)$ we define the \emph{per-token empirical loss}
\begin{equation}\label{eq:lnkw_linear}
    \ell_{N,k}(f)
    =
    \frac{1}{N} \sum_{i=1}^N  (f(S^i_{\leq k}) - y_k^i)^2
    .
\end{equation}
Averaging over context lengths we obtain the \textit{empirical loss}
\begin{equation}\label{eq:ln}
    \ell_N(f) = \frac{1}{K} \sum_{k=1}^{K} \ell_{N,k}(f)
    .
\end{equation}
The corresponding \textit{population loss} $\ell(f)$ is defined by taking the expectation over the data distribution $q(S)$,
\begin{equation}
    \ell(f)
    =
    \mathbb{E}_{S \sim q} \left[
        \frac{1}{K} \sum_{k=1}^K (f(S_{\leq k}) - y_k^i)^2
    \right]
    .
\end{equation}
For a function $f(\cdot, w)$ implemented by a transformer with parameter $w$, we denote the losses $\ell_{N,k}(w)$, $\ell_{N}(w)$, and $\ell(w)$. For task diversity $M$ we use a superscript $\ell^M$.

\subsection{Idealized in-context linear regression predictors}
\label{section:dmmse-vs-ridge}

Given a context $S_{\le k}$ there are many possible algorithms that could be chosen to predict $\hat{y}_k$.
\citet{raventós2023pretraining} studied the following two predictors:

\paragraph{Predictor 1 \textnormal{(dMMSE)}.}
For $M \in \mathbb{N}$ and $k = 1, \ldots, K$, the \emph{discrete minimum mean squared error} predictor, \dMMSE[M], is the function
    $g^M_k : (\mathbb{R}^D \times \mathbb{R})^k \times \mathbb{R}^D \to \mathbb{R}$
such that
\begin{equation}\label{eq:dmmse}
    g_k^M(x_1, y_1, \ldots, x_k)
    = 
    \left(\hat{\task}_{k}^{M}\right)^\top x_k
\end{equation}
where the \dMMSE[M] \emph{task estimate}
    $\hat{\task}^{M}_k \in \mathbb{R}^D$
is given by
\begin{equation*}
    \hat{\task}_{k}^{M}
    = \frac{
        \sum_{m=1}^M
        \exp\left(
            -\frac{1}{2\sigma^2}
            \sum_{j=1}^{k-1} \left(y_j - \task[m]^\top x_j \right)^2
        \right)
        \task[m]
    }{
        \sum_{m=1}^M
        \exp\left(
            -\frac{1}{2\sigma^2}
            \sum_{j=1}^{k-1} (y_j - \task[m]^\top x_j)^2
        \right)
    }
    .
\end{equation*}
Note that the \dMMSE[M] task estimate and therefore the prediction explicitly depends on the tasks $\taskdist[M] = \{\task[1], \ldots, \task[M]\}$.

\paragraph{Predictor 2 \textnormal{(ridge)}.}
For $k = 1, \ldots, K$, the ridge predictor
    $g^{\infty}_k : (\mathbb{R}^D \times \mathbb{R})^k \times \mathbb{R}^D \to \mathbb{R}$
is given by
\begin{equation}\label{eq:ridge}
    g^{\infty}_k(x_1, y_1, \ldots, x_k)
    =
    \left(\hat{\task}_{k}^{\infty}\right)^\top x_k
\end{equation}
where if $k=1$ the task estimate is $\hat{\task}_{k}^{\infty} = \mathbf{0}$, otherwise the task estimate $\hat{\task}_{k}^{\infty}$ is given by $L_2$-regularized least-squares regression on the examples in the context with the regularization parameter set to $\sigma^2$,
\begin{equation*}
    \hat{\task}_{k}^{\infty}
    =
    \left(X^\top X + \sigma^2 I_D\right)^{-1} X^\top Y,
\end{equation*}
where
    $X = (x_1^\top, \dots, x_{k-1}^\top)$ and 
    $Y = (y_1, \dots, y_{k-1})$.

\paragraph{Optimality of predictors.}
\citet{raventós2023pretraining} showed that for finite task diversity $M \in \mathbb{N}$, given data distribution $\q{M}{S}$, the minimum mean squared error predictions are given by
    equation~\cref{eq:dmmse}, that is, \dMMSE[M],
whereas for infinite task diversity, given data distribution $\q{\infty}{S}$, the minimum mean squared error predictions are given by
    equation~\cref{eq:ridge}, that is, ridge.

Moreover, note that for a fixed context $x_1, y_1, \ldots, x_k$, we have that as $M\to\infty$,
$
    \hat{\task}_{k}^{M}
    \xrightarrow{\text{a.s.}}\,
    \hat{\task}_{k}^{\infty}
$.
It follows that ridge is an \emph{approximately} optimal predictor for $\q{M}{S}$ given a large finite task diversity $M$.
However, for all finite task diversities $M$ it remains possible to reduce expected loss on $\q{M}{S}$ by specializing to the tasks in $\taskdist[M]$ (at the cost of increased loss on sequences constructed from other tasks).

\paragraph{Consistency of predictors.}
The task estimates $\hat{\task}_{k}^{M}, \hat{\task}_{k}^{\infty}$ are both asymptotically consistent assuming unbounded sequences drawn based on a realizable task $\task \in \taskdist[M]$.
However, the task estimates will differ for all $k$ (due to the different priors, $\q{M}{\task}$ and $\q{\infty}{\task}$), particularly for early tokens and especially in the under-determined regime
    $k \leq D$.

\section{The transient ridge phenomenon}\label{section:pca}

In this section, we replicate the transient ridge phenomenon observed by \citet[\S6.1]{panwar2024bayesianprism} by training transformers at a range of task diversity parameters and evaluating their performance on out-of-distribution (OOD) sequences.

We then apply the general technique of \emph{joint trajectory PCA}:
We use principal component analysis (PCA) to decompose the collective function-space trajectories of the transformers, producing a low-dimensional representation of their behavioral development.
Without having to specify the idealized predictors, we recover the difference between dMMSE and ridge as correlated to the second principal component, and show that in the lead up to the task diversity threshold trajectories are increasingly drawn towards ridge.

\subsection{Transformer training}

We train transformers on nested multi-task data distributions with varying task diversity
    (\cref{section:setting:data})
under the mean squared error objective
    (\cref{section:setting:loss})
to see when they behaviorally approximate dMMSE or ridge
    (\cref{section:dmmse-vs-ridge}).
We use a $2$-layer transformer with $d=2.65$~million parameters (details in \cref{appendix:transformer-details}; more architectures in \cref{appendix:architecture-comparison}).

We train with each of a set $\mathcal{M}$ of task diversities ranging from $M = 1$ to $M = 2^{15}$ and also including $\infty$.
Each run generates a trajectory
    $w_t^M \subseteq \mathbb{R}^d$
through parameter space for training steps $t=0, \ldots, T$, from which we subsample checkpoints $\mathcal{C} \subseteq \{0, \ldots, T\}$ using a union of linear and logarithmic intervals (\cref{appendix:checkpoint-sampling}).
For notational ease, we sometimes denote the function $f(\cdot, w_t^M)$ as $f_M(\cdot, w_t)$.

\subsection{Joint trajectory principal component analysis}

An established method for studying the development of structure and function in a system is to analyze its trajectory in configuration space.
\citet{amadei1993essential} developed the technique of applying PCA to such trajectories, called \emph{essential dynamics} or simply \emph{trajectory PCA}.
It is argued that important features of trajectories appear in the \emph{essential subspace} spanned by the leading principal components \citep{briggman2005optical,ahrens2012brainwide}, though interpreting PCA of time series requires care
    \citetext{%
        cf.,~\citealp{shinn2023phantom,antognini2018pca};
        also~\cref{appendix:pca-perils}%
    }.

Trajectory PCA has seen diverse applications in molecular biology \citep{amadei1993essential,meyer2006essential,hayward2008normal}, neuroscience \citep{briggman2005optical,cunningham2014dimensionality}, and deep learning \citep{olsson2022context,mao2024manifold}.
We adapt a multi-trajectory variant \citep{briggman2005optical} to study the \emph{collective} behavioral dynamics of our family of transformer models trained with different task diversities. In particular, we simultaneously perform PCA on the trajectories of all models through a finite-dimensional subspace of function space.
Our detailed methodology for this \emph{joint trajectory PCA} is as follows.

\paragraph{Joint encoding of transformer trajectories.}
Given a parameter $w$, we can view the transformer as mapping each sequence $S$ to a vector of predictions for its $K$ subsequences,
\begin{equation*}
    f(S, w)
    = \big( f(S_{\leq1}, w), \ldots, f(S_{\leq K}, w) \big)
    \in \mathbb{R}^K.
\end{equation*}
We fix a finite dataset
    $\mathcal{D}^{(1)} = \{S^i\}_{i=1}^B \sim \q{1}{S}$
of $B=512$ input sequences (recalling that the root task~$\task[1]$ is shared by all task sets, so is the natural task to use to compare in-distribution behavior).
We concatenate the outputs of $f(\cdot, w)$ for each input $S^i$ into one long row vector,
\begin{equation*}
    f(\mathcal{D}^{(1)}, w)
    = \big( f(S^1, w), \ldots, f(S^B, w) \big)
    \in \mathbb{R}^{BK},
\end{equation*}
representing the function $f(\cdot, w)$ as a point in a finite-dimensional subspace of function space.

We apply this construction to each transformer checkpoint
    $\{w_t^M\}_{M \in \mathcal{M}, t \in \mathcal{C}}$.
For each $M \in \mathcal{M}$, we aggregate the row vectors from each checkpoint into a matrix
    $F_M \in \mathbb{R}^{|\mathcal{C}| \times BK}$
and then stack each $F_M$ vertically into
    $F_{\mathcal{M}} \in \mathbb{R}^{|\mathcal{M}| |\mathcal{C}| \times BK}$:
\begin{equation*}
    F_M
    =
    \begin{bmatrix}
        f(\mathcal{D}^{(1)}, w^M_{t_1}) \\
        f(\mathcal{D}^{(1)}, w^M_{t_2}) \\
        \vdots \\
        f(\mathcal{D}^{(1)}, w^M_{t_{|\mathcal{C}|}})
    \end{bmatrix}
    \text{\ for\ }
    M \in \mathcal{M},
    \quad
    F_{\mathcal{M}}
    =
    \begin{bmatrix}
        F_1 \\
        \vdots \\
        F_{2^{15}} \\
        F_{\infty}
    \end{bmatrix}.
\end{equation*}

\paragraph{Principal component analysis.} \label{sec:PCA_of_F_M}
We apply PCA to the joint matrix $F_{\mathcal{M}}$.
Supposing $F_{\mathcal{M}}$ has been mean-centered, it has a singular value decomposition
    $F_{\mathcal{M}} = U \Lambda V^{\top}$
where
    $U \in \mathbb{R}^{|\mathcal{M}| |\mathcal{C}| \times |\mathcal{M}| |\mathcal{C}|}$ has left singular vectors as columns,
    $\Lambda \in \mathbb{R}^{|\mathcal{M}| |\mathcal{C}| \times BK}$ is a diagonal matrix of ordered positive singular values,
and
    $V \in \mathbb{R}^{BK \times BK}$ has right singular vectors as its columns.
For $v \leq BK$, let $V_v$ denote the \emph{loading matrix} given by the first $v$ columns of $V$. The span of these columns forms the $v$-dimensional \emph{(joint) essential subspace}.

\paragraph{Projecting into the essential subspace.}
The corresponding projection from feature space into the essential subspace is given by
    $\pi_v: \mathbb{R}^{BK} \to \mathbb{R}^v$
where
    $\pi_v(y) = y V_v$.
The developmental trajectory of each model is then represented as a curve
    $\gamma_M: \mathcal{C} \to \mathbb{R}^v$ 
defined by
\begin{equation*}
    \gamma_M(t) = \pi_v(f(\mathcal{D}^{(1)}, w^M_t))\,.
\end{equation*}
For any principal component dimension $i\leq v$ we call the $i$\textsuperscript{th} component function $\gamma_M^i : \mathcal{C} \to \mathbb{R}$ a \emph{PC-over-time curve.}

The \dMMSE[M] and ridge predictors defined in equations~\cref{eq:dmmse} and~\cref{eq:ridge} can likewise be encoded as row vectors and projected into the essential subspace.
For $M \in \mathbb{N} \cup \{\infty\}$ let
\begin{equation*}
    G_M = \big(
        g^M_1(S^1_{\leq 1}),
        g^M_2(S^1_{\leq 2}),
        \ldots,
        g^M_K(S^B_{\leq K})
    \big) \in \mathbb{R}^{BK}.
\end{equation*}
Then each predictor projects to a single point in the essential subspace, $\pi_v(G_M)$.
Note that we do not include the points $G_1$, \ldots, $G_{\infty}$ in the data prior to performing PCA.

\begin{figure*}[t!]
    \centering
    \includegraphics[width=\linewidth]{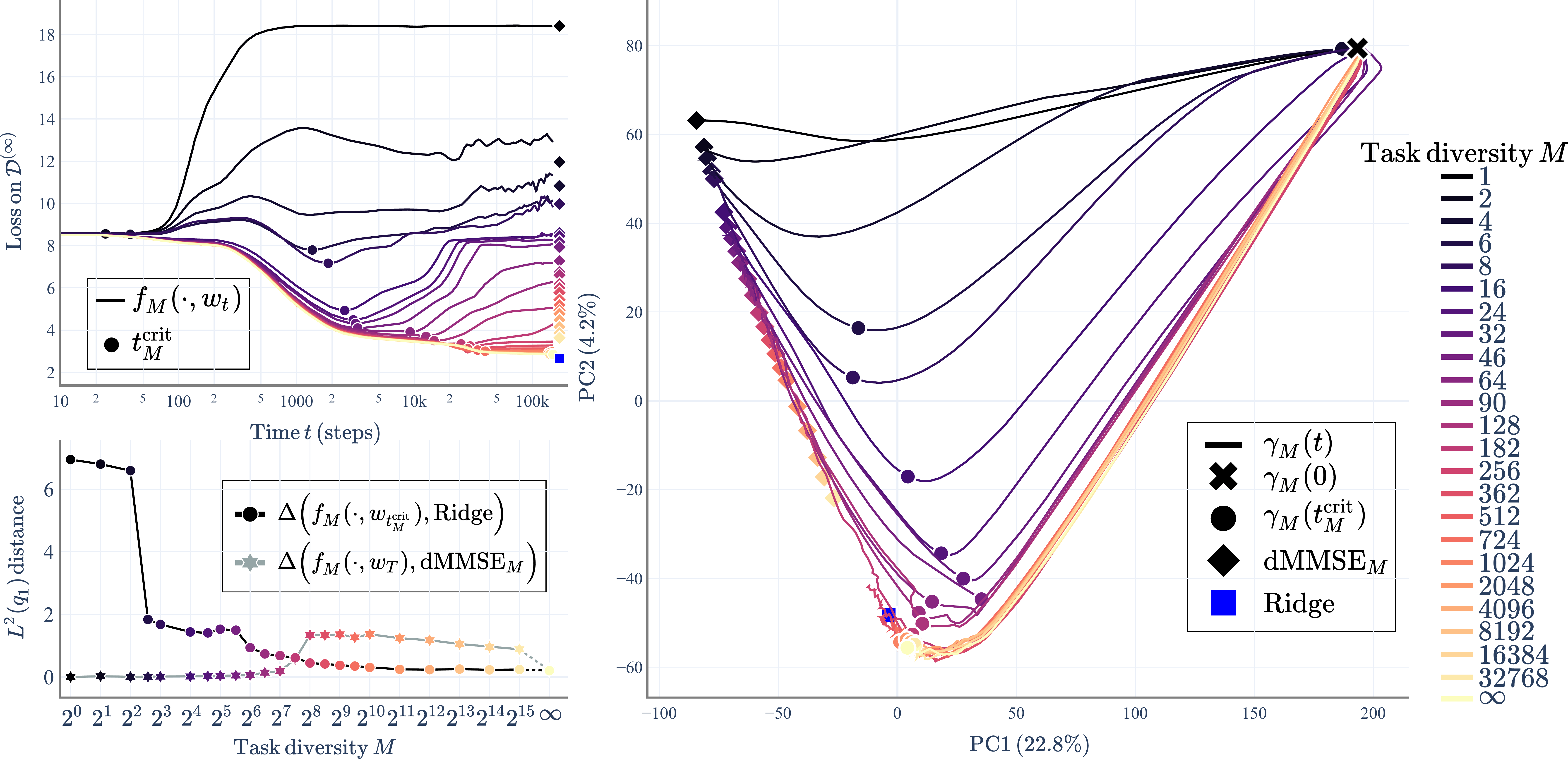}
    \caption{\label{fig:ood-and-pca}%
    \textbf{Behavioral dynamics of the transient ridge phenomenon.}
    \emph{(Top left):}
        OOD loss over training on sequences sampled with a Gaussian task distribution for task diversities $M \in \mathcal{M}$.
        For intermediate $M$ we see non-monotonicity caused by the transient ridge phenomenon, or ``forgetting'' as observed by \citet[\S6.1]{panwar2024bayesianprism}.
        We define $\tcrit[M]$ as the step at which the OOD loss is minimized for $M$
        (\cref{appendix:tcrit}).
        We mark this step with a circle in the other plots.
    \emph{(Right):}
        We project each transformer's trajectory $\{f(\cdot, w_t^M)\}_{t \in \mathcal{C}}$ to a curve $\gamma_M(t)$ in the essential subspace computed by joint trajectory PCA.
        We project \dMMSE[M] (diamonds) and ridge (square) into the same subspace.
        For intermediate task diversity $M$, the development is deflected towards ridge on its way towards \dMMSE[M].
    \emph{(Bottom left):}
        In-distribution function-space distances $\Delta(\cdot, \text{\dMMSE[M]/Ridge})$ clarify which fully-trained transformers (stars) approximate \dMMSE[M], and which transformers approximate ridge at $\tcrit[M]$ (circles).
    \emph{(Note):} loss and PC curves are lightly smoothed, see \cref{section:gaussian_smoothing} for raw data.
    }
\end{figure*}

\subsection{Experimental results}\label{section:ood-results}

\Cref{fig:ood-and-pca} shows OOD loss on a fixed test set $\mathcal{D}^{(\infty)} \sim \q{\infty}{S}$ and the result of 2-dimensional joint trajectory PCA ($27.5\%$ explained variance).
\Cref{appendix:ood} shows in-distribution loss.
\Cref{appendix:pca} extends to 4-dimensional PCA, explores the effect of checkpoint distributions, and shows that results are insensitive to the choice of batch size $B \geq 16$.

\paragraph{Essential subspace.}
Strikingly, PC2 correlates with an axis of behavioral difference between \dMMSE[M] (for increasing $M$) and ridge.
\Cref{appendix:per-token-retreat} confirms that the predictions on earlier tokens, where dMMSE and ridge differ more, load more heavily on PC2 than those for later tokens do.
PC2 also correlates with OOD loss, while PC1 appears to correlate with loss on $\q{1}{S}$ and with a notion of ``development time'' (see \cref{fig:D00-TPCs-grid} and \cref{appendix:PC1-dev-time}).

\paragraph{Task diversity threshold.}
As in \citet{raventós2023pretraining},
    at low task diversity (in our case $M \leq 128$), fully-trained transformers behaviorally approximate \dMMSE[M],
while
    above a \emph{task diversity threshold} ($M \geq 362$), they converge to a point that behaviorally approximates ridge.
Trajectories $M \in \{182, 256\}$ converge somewhere between.

\paragraph{Transient ridge.}
Replicating \citet[\S6.1]{panwar2024bayesianprism}, we see that for \emph{intermediate $M$} in the lead-up to the task diversity threshold, the OOD loss is non-monotonic.
For $M \in \{16,24,\ldots,128\}$, loss decreases towards that of ridge, then increases to that of \dMMSE[M]. We see a partial dip for $M \in \{6, 8\}$ and a partial rise for $M \in \{182,256\}$.

Trajectory PCA reveals that this non-monotonicity coincides with changes in the development of \emph{in-distribution} behavior.
For low $M$, the transformers proceed directly to \dMMSE[M] in the essential subspace.
As $M$ increases (until the task diversity threshold), the trajectories are increasingly deflected from this straight path into one that \emph{transits via approximating ridge}.
Beyond the task diversity threshold, the trajectories proceed directly to ridge and do not depart.

This trajectory PCA result suggests that the presence of the approximate ridge solution in the optimization landscape is in some sense influencing the development of internal structures in the transformers.
Moreover, as $M$ increases, as the \dMMSE[M] solution changes, the strength of the influence of the ridge solution increases.
In the next section, we attempt to understand the nature of this influence.

\section{Evolving loss/complexity tradeoff}\label{section:slt}

In this section, we model the transient ridge phenomenon as the result of the transformer navigating an evolving tradeoff between loss and complexity as it undergoes additional training. We draw on the theory of Bayesian internal model selection to qualitatively predict the nature of the tradeoff.
We then empirically validate this model of the phenomenon by quantifying the complexity of the fully-trained transformers using the associated complexity measure.

\subsection{Learning solutions of increasing complexity}

The learning dynamics of many systems follow a pattern of progressing from solutions of \emph{high loss} but \emph{low complexity} to solutions of \emph{low loss} but \emph{high complexity}.
This pattern has been studied in detail in certain models including
    deep linear networks \citep[e.g.,][]{baldi1989neural,saxe2019mathematical,gissin2019implicit,jacot2021saddle},
    multi-index models \citep{abbeSGDLearningNeural2023},
and
    image models \citep{kalimeris2019increasingcomplexity},
each with their own notion of ``complexity.''

Unfortunately, we lack results describing how such a progression should play out, or what complexity measure to use, for general deep learning.
Therefore, we turn to singular learning theory
    \citep[SLT;][]{greybook,watanabe2018}---%
    a framework for studying statistical models with degenerate information geometry, including neural networks
    \citep{hagiwara1993,watanabe2007almost,goodpaper}---%
in which a similar loss/complexity tradeoff has been studied in general terms in the setting of Bayesian inference. 

\subsection{Bayesian internal model selection}
\label{section:internal-model-selection}

In Bayesian inference, SLT shows that the solutions around which the posterior concentrates are determined by a balance of loss and complexity.
Moreover, the ideal balance changes as the number of samples increases, driving a progression from simple but inaccurate solutions to accurate but complex solutions
    \citetext{\citealp[\S7.6]{greybook}; \citealp{chen2023tms1}}.
The leading-order complexity measure is the \emph{local learning coefficient}
    \citep[LLC;][]{quantifdegen},
which can be understood as a degeneracy-aware effective parameter count.
We outline this \emph{internal model selection} principle below.

\paragraph{Bayesian posterior.}
Consider a neural network parameter space
    $\mathcal{W} \subseteq \mathbb{R}^d$.
Let
    $\varphi$
be a nonzero prior over $\mathcal{W}$ and
    $\ell_n : \mathcal{W} \to \mathbb{R}$
an empirical loss (the average negative log likelihood) on $n$ samples.
Then the Bayesian posterior probability of a neighborhood
    $\mathcal{U} \subseteq \mathcal{W}$
given $n$ samples is
\begin{equation*}
    p_n(\mathcal{U})
    =
    \frac{Z_n(\mathcal{U})}{Z_n(\mathcal{W})}
\end{equation*}
where $Z_n(\mathcal{X})$ is the marginal likelihood of
    $\mathcal{X} \subseteq \mathcal{W}$,
\begin{equation*}
    Z_n(\mathcal{X})
    =
    \int_{\mathcal{X}}\exp(-n\ell_n(w))\varphi(w)\,dw.
\end{equation*}

\paragraph{Bayesian posterior log-odds.}
Consider two neighborhoods
    $\mathcal{U}, \mathcal{V} \subseteq \mathcal{W}$.
The preference of the Bayesian posterior for $\mathcal{U}$ over $\mathcal{V}$ can be summarized in the \emph{posterior log-odds},
\begin{equation}\label{eq:posterior-odds}
    \log\frac{p_n(\mathcal{U})}{p_n(\mathcal{V})}
    =
    \log Z_n(\mathcal{U}) - \log Z_n(\mathcal{V}),
\end{equation}
which is positive to the extent that $p_n$ prefers $\mathcal{U}$ over $\mathcal{V}$.

\paragraph{Watanabe's free energy formula.}
SLT gives an asymptotic expansion of the Bayesian local free energy $-\log Z_n(\cdot)$.
Let $u \in \mathcal{W}$ be a \emph{solution}, that is, a local minimum of the expected negative log likelihood, and let $\mathcal{U}$ be a closed ball around $u$, in which $u$ is a maximally degenerate global minimum.
Then, under certain technical conditions on the model, we have the following asymptotic expansion in $n$
    \citetext{\citealp[Theorem 11]{watanabe2018}; \citealp{quantifdegen}}:
\begin{equation}\label{eq:free-energy-formula}
    -\log Z_n(\mathcal{U})
    = \ell_n(u) \cdot n
    + \lambda(u) \cdot \log n
    + O_p(\log\log n)
\end{equation}
where $\lambda(u)$ is the LLC and the lower-order terms include various other contributions, such as from the prior.

\paragraph{The loss/complexity tradeoff.}
If $v \in \mathcal{V}$ is a competing solution (with its own neighborhood), then \cref{eq:posterior-odds} and \cref{eq:free-energy-formula} give
\begin{equation}\label{eq:tradeoff}
    \log\frac{p_n(\mathcal{U})}{p_n(\mathcal{V})}
    = \Delta\ell_n \cdot n 
      + \Delta\lambda \cdot \log n
      + O_p(\log\log n)
\end{equation}
where $\Delta\ell_n = \ell_n(v) - \ell_n(u)$ and $\Delta\lambda = \lambda(v) - \lambda(u)$.

Equation~\cref{eq:tradeoff} describes an evolving tradeoff between loss and complexity as follows.
Assume the lower-order terms from each free energy expansion cancel.
Then
    if $\Delta\ell_n < 0$ ($u$ has higher loss than $v$)
    and $\Delta\lambda > 0$ ($u$ has lower LLC than $v$),
the sign of the log-odds depends on $n$.
The Bayesian posterior will prefer
    $\mathcal{U}$ (around the simple but inaccurate solution)
until
    $\log(n)/n < (- \Delta \ell_n)/\Delta\lambda$,
after which it will prefer
    $\mathcal{V}$ (around the accurate but complex solution).

\paragraph{From Bayesian inference to deep learning.}
Neural networks are typically trained by stochastic gradient-based optimization, not Bayesian inference.
Nevertheless, as described in \cref{section:related-work}, recent work suggests that some qualitatively similar evolving tradeoff governs the development of structure in deep learning over \emph{training time} \citep{chen2023tms1,icl1,lang1}.

This suggests that some as-yet-unknown principle of ``dynamic\footnotemark{} internal model selection''---%
    in which the loss and the LLC play leading roles%
---underpins the structure of the optimization landscape, in turn influencing the trajectories followed by stochastic gradient-based optimization.
Based on this motivation, we apply equation~\cref{eq:tradeoff} to qualitatively predict the transient ridge phenomenon in terms of the differences in loss and LLC of the transformers that approximately implement the \dMMSE[M] and ridge predictors.
\footnotetext{
    In the sense of nonlinear dynamics \citep[cf., e.g.,][]{strogatz}, where it is well-established that degeneracy in the geometry of critical points of a governing potential influences system trajectories.
}

\subsection{Explaining the transient ridge phenomenon}\label{section:prediction}

\begin{figure}[t!]
    \centering
    \includegraphics
        [width=\columnwidth]
        {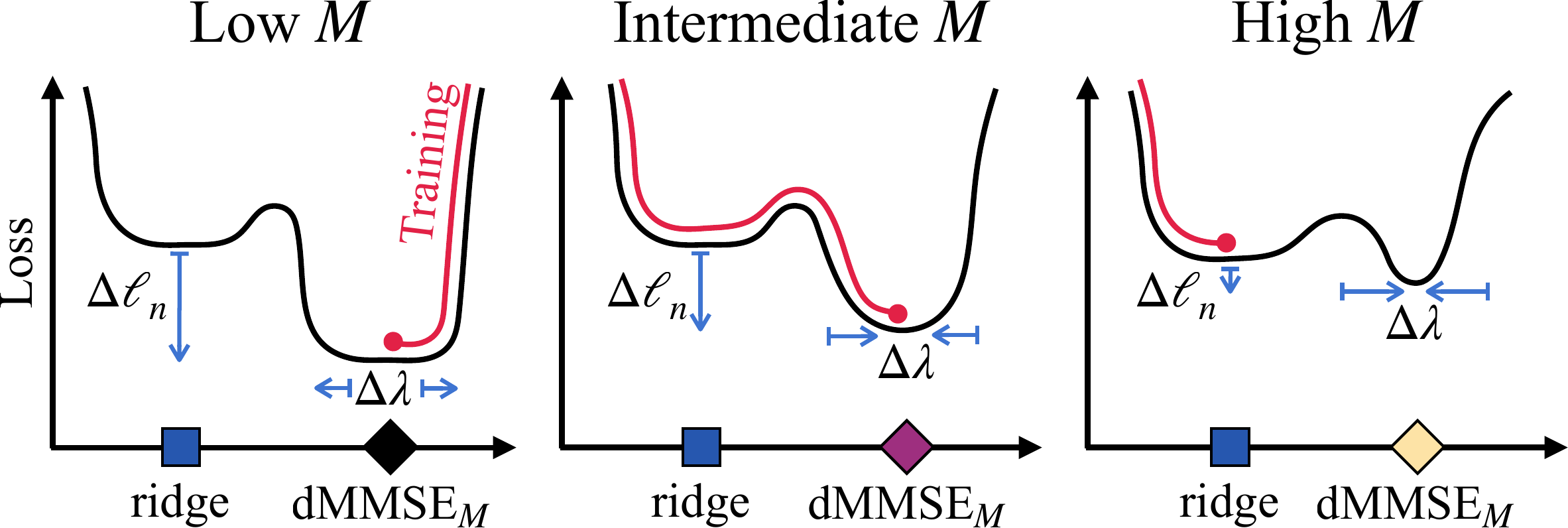}
    \caption{\label{fig:cartoon}%
        \textbf{Transient ridge in the loss landscape.}
        Conceptual illustration of transient ridge arising as the result of an evolving tradeoff between loss and LLC (complexity, illustrated as sharpness).
        As $M$ increases, we expect the loss gap between \dMMSE[M] and ridge to shrink and the LLC of \dMMSE[M] to grow, creating transience for intermediate $M$.
    }
\end{figure}

We offer the following explanation for the dynamics of transformers trained on in-context linear regression data with task diversity $M \in \mathbb{N}$. Let $u_\infty \in \mathcal{U}$ and $v_M \in \mathcal{V}_M$ be transformer parameters approximately implementing ridge and \dMMSE[M] respectively, along with their neighborhoods.
\begin{enumerate}
    \item \emph{Low $M$:}
        We expect $v_M$ to have much lower loss \emph{and} LLC than $u_\infty$.
        As equation~\cref{eq:tradeoff} never favors $\mathcal{U}$, training should proceed directly to $v_M$.

    \item \emph{Intermediate $M$:}
        We expect $v_M$ to have lower loss but higher LLC than $u_\infty$.
        As equation~\cref{eq:tradeoff} initially favors $\mathcal{U}$ but eventually favors $\mathcal{V}_M$, training should proceed first towards $u_\infty$ before pivoting towards $v_M$ (after a number of training steps that increases with $M$).

    \item \emph{High $M$:}
        We expect $v_M$ to have slightly lower loss but much higher LLC than $u_\infty$.
        As equation~\cref{eq:tradeoff} only favors $\mathcal{V}_M$ at very high $n$, trajectories should proceed to $u_\infty$ and should not depart by the end of training.
\end{enumerate}
See \cref{fig:cartoon} for a conceptual illustration.
For $M$ values that fall between these three prototypical cases, the posterior preference is less sharp. Therefore we expect to see gradual shifts the dynamics over the range of $M$ values.

\subsection{Empirical validation of the explanation}

The above explanation is consistent with the findings of \cref{section:pca}.
It remains to validate that the trends in the loss and LLC are as expected.
In this section, we outline our experiments estimating the loss and LLC of $u_\infty$ and $v_M$.

\begin{figure}[t!]
    \centering
    \includegraphics[width=\columnwidth]{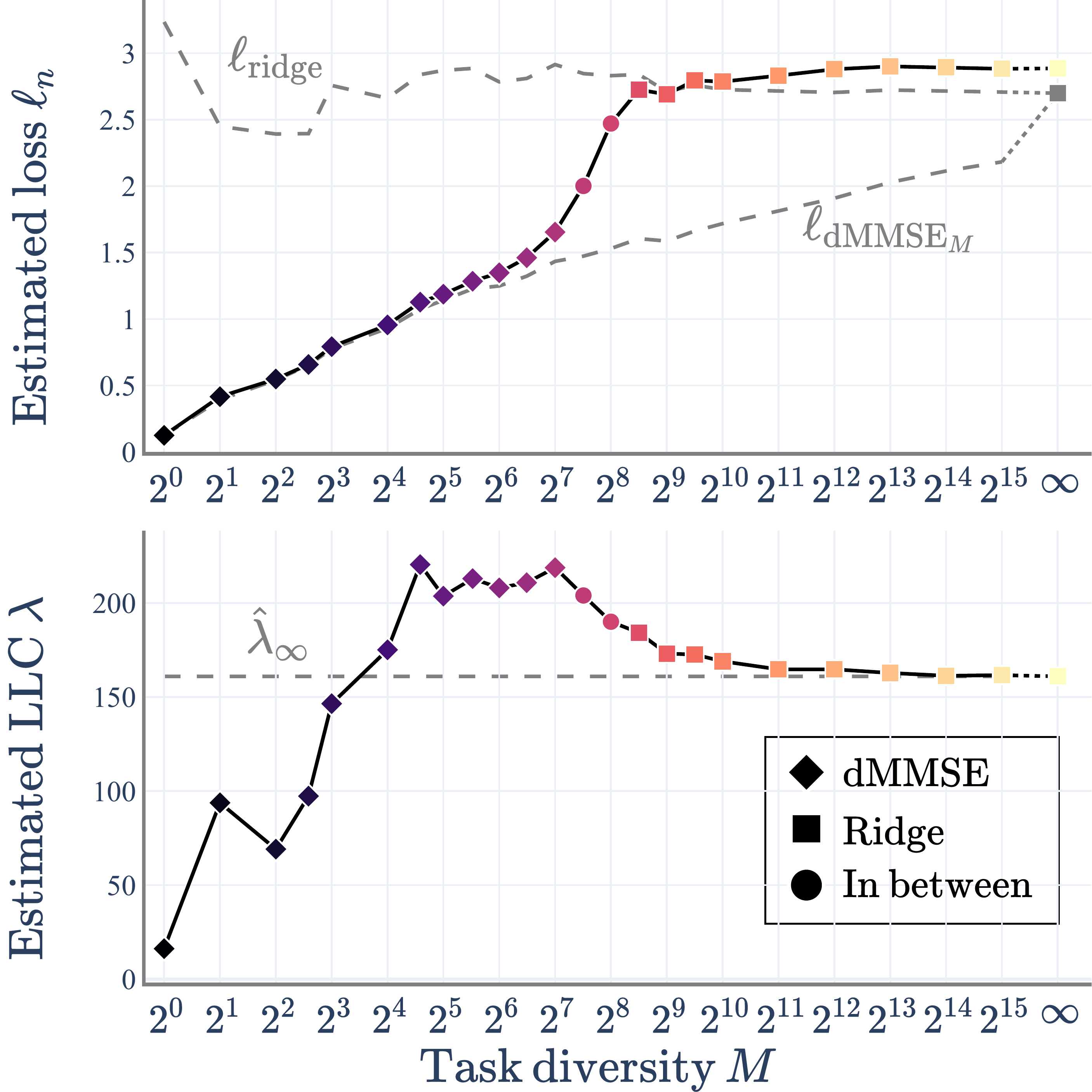}
    \caption{\label{fig:clean-explanation}%
    \textbf{Loss and LLC estimates match predictions.}
    \emph{(Top):}
        Estimated loss with respect to data distribution $q_M(S)$ for the idealized predictors and fully-trained transformers.
        The gap between \dMMSE[M] and ridge decreases with $M$, and trained transformers approximate this loss on either side of the task diversity threshold (diamonds for \dMMSE[M], squares for ridge).
    \emph{(Bottom):}
        Estimated LLC for fully-trained transformers. Large-$M$ LLCs converge to the LLC of ridge (dashed line). Small-$M$ LLCs, representing the LLC of \dMMSE[M], cross this line as $M$ increases.
    }
\end{figure}

\paragraph{Estimating loss.}
We first estimate the loss of $u_\infty$ and $v_M$ by directly evaluating the idealized predictors~\cref{eq:dmmse} and~\cref{eq:ridge}.
Alternatively, noting that the transformer cannot necessarily realize these predictors, we evaluate the end-of-training parameters (representing $v_M$ for low $M$ or $u$ for high $M$).
\Cref{fig:clean-explanation}(top) confirms that the loss gap between idealized predictors shrinks with increasing $M$, and the transformers achieve similar loss to their respective algorithms.

\paragraph{Estimating LLC.}
The LLC is architecture-dependent, so we can't meaningfully measure the LLC of the idealized predictors, only our fully-trained transformers (representing $v_M$ for low $M$ or $u_\infty$ for high $M$).
Following \citet{quantifdegen}, we estimate the LLC of a parameter $w_*^M$ as the average increase in empirical loss $\ell_n$ for nearby parameters,
\begin{equation}\label{eq:llc-estimator}
    \hat\lambda(w_*^M)
    =
    n \beta\left(
        \mathbb{E}_{w \mid w_*^M, \gamma}^\beta\bigl[\ell_n(w)\bigr]-\ell_n(w_*^M)
    \right),
\end{equation}
where $n$ is a sample size, $\beta$ is an inverse temperature, $\gamma$ is a localization strength parameter, and $\mathbb{E}_{w \mid w_*^M, \gamma}^\beta$ is expectation over the localized Gibbs posterior
\begin{equation*}
    p\left(w ; w_*^M, \beta, \gamma\right)
    \propto
    \exp \left(
        -n \beta \ell_n(w)-\frac{\gamma}{2}\left\|w-w_*^M\right\|_2^2
    \right).
\end{equation*}
We sample from this posterior with stochastic gradient Langevin dynamics
    \citep[SGLD;][]{wellingBayesianLearningStochastic2011}.
\Cref{appendix:llc} gives further details on LLC estimation, sampling with SGLD, and hyperparameter calibration.

\Cref{fig:clean-explanation}(bottom) shows LLC estimates $\hat \lambda(w_*^M)$ for fully-trained transformers.
High-$M$ LLCs converge to $\hat \lambda_\infty$, which we take as $\lambda(u_\infty)$, the LLC of the ridge solution.
For low-$M$ transformers that converge to \dMMSE[M], we take $\hat \lambda(w_*^M)$ as $\lambda(v_M)$.
As expected, this LLC increases with $M$, and crosses $\hat \lambda_\infty$ during the onset of transience.
Surprisingly, the estimated LLC of \dMMSE[M] plateaus above $M=32$, suggesting that the approximation achieved by the fully-trained transformers may be incomplete.

\section{Limitations and future work}\label{section:discussion}

Our findings support an understanding of the transient ridge phenomenon as driven by an evolving loss/complexity tradeoff, governed by principles that are yet to be fully discovered but qualitatively resemble Bayesian internal model selection.
In this section, we enumerate remaining gaps in this understanding, representing future steps towards a comprehensive understanding of neural network development.

\subsection{Transformers versus idealized predictors}

Our analysis is based primarily on in-distribution behavior, and it is not clear that our transformers can or do faithfully approximate the idealized predictors for all input sequences.
Moreover, it is unclear whether the solutions governing training dynamics are necessarily the parameters to which transformers converge (we consider an alternative interpretation in \cref{appendix:clean-v-messy}).
Future work could seek a more detailed understanding of transformer solutions arising in practice, for example using mechanistic interpretability.

\subsection{The role of lower-order terms}
\label{section:limitations:lower-order-terms}

In \cref{section:internal-model-selection}, we make the simplifying assumption that the lower-order terms from each expansion cancel. However if these terms are not equal then their difference enters the posterior log odds, influencing the evolution of the posterior, especially for low $n$.
SLT has studied these terms \citep[cf., e.g.,][]{quantifdegen}, but they are not as well-studied as the LLC.
Future work could deepen our theoretical understanding of these lower-order terms or our empirical understanding of their role in internal model selection.

\subsection{Dynamic versus Bayesian internal model selection}

Of course, our primary motivation is to study neural network development, rather than Bayesian internal model selection per se.
While we have contributed further evidence that the loss and the LLC play a leading role in a principle of ``dynamic internal model selection'' that governs neural network development, the precise form of this principle and the precise roles of the loss and the LLC remain to be determined. 
Our work highlights this as a promising direction for future empirical and theoretical work.

\subsection{Why does transience \emph{stop?}}

Bayesian internal model selection suggests that the ridge solution should \emph{always eventually} give way to a more complex but more accurate \dMMSE[M] solution.
In practice, replicating \citet{raventós2023pretraining}, we see a clear \emph{task diversity threshold} above which transformers never leave ridge.
This could be due to capacity constraints, under-training, neuroplasticity loss, or a concrete difference between Bayesian and ``dynamic'' internal model selection.
\cref{appendix:end-of-retreat} offers a preliminary analysis, but reaches no firm conclusion, leaving this an open question for future work.

\subsection{Beyond in-context linear regression}

As outlined in \cref{section:related-work}, the phenomenon of a transient generalizing solution giving way to a memorizing solution over training has now been observed in a range of sequence modeling settings beyond our setting of in-context linear regression
    \citetext{%
        \citealp{singh2024transient};
        \citealp{icl1};
        \citealp{edelman2024statisticalinductionheads};
        \citealp{he2024learningtogrok};
        \citealp{park2024competition}%
    }.
There's also the reverse phenomenon of a transition from a memorizing solution to an equal-loss but simpler generalizing solution \citep[``grokking;''][]{power2022grokking,nanda2023progress}.

These settings represent rich subjects for future empirical work investigating the principles governing the development neural networks. In the case of grokking, we note that the particular loss/complexity tradeoff outlined in \cref{section:slt} does not account for transitions that \emph{decrease} complexity, though such transitions can be described within the Bayesian internal model selection framework by taking lower-order terms into account (cf.~\cref{section:limitations:lower-order-terms}).

\section{Conclusion}

This paper contributes an in-depth study of the training dynamics of transformers in the settings of in-context linear regression with variable task diversity.
We adapt the technique of trajectory principal component analysis from molecular biology and neuroscience and deploy it to expand our empirical understanding of the developmental dynamics of our transformers, and the variation in these dynamics with task diversity, revealing the choice between memorization and generalization as a principal axis of development.

Moreover, we adopt the perspective of singular learning theory to offer an explanation of these dynamics as an evolving tradeoff between loss and complexity (as measured by the local learning coefficient), finding evidence that these elements play a leading role in governing the development of our transformers, akin to their role in governing the development of the posterior in Bayesian internal model selection.
These findings open the door to future research aiming to uncover the true principles governing the development of internal structure in deep learning.

\section*{Impact statement}

The emergence of computational structure in deep neural networks is not only a fascinating scientific and mathematical phenomenon.
This structure determines a model's out-of-distribution generalization behavior, and in turn its safety, robustness, and alignment properties.
As society races ahead to develop ever more complex neural networks and integrate them ever more deeply into our digital and physical world, understanding the principles governing neural network development is a priority for the science of deep learning.
This work aims to contribute towards improving our scientific understanding of neural network development, which is an integral part of (though not alone sufficient for) ensuring that future technological advances in the field of deep learning have robustly positive impact.

\section*{Acknowledgments}
\addcontentsline{toc}{section}{Acknowledgments}

We thank 
    Edmund Lau,
    George Wang,
and 
    Susan Wei
for helpful conversations.
Google's TPU Research Cloud program supported some of our experiments with Cloud TPUs.


\urlstyle{same}
\bibliographystyle{icml2025}
\bibliography{main}

\begin{thebibliography}{60}
\providecommand{\natexlab}[1]{#1}
\providecommand{\url}[1]{\texttt{#1}}
\expandafter\ifx\csname urlstyle\endcsname\relax
  \providecommand{\doi}[1]{doi: #1}\else
  \providecommand{\doi}{doi: \begingroup \urlstyle{rm}\Url}\fi

\bibitem[Abbe et~al.(2023)Abbe, Adser{\`a}, and
  Misiakiewicz]{abbeSGDLearningNeural2023}
Abbe, E., Adser{\`a}, E.~B., and Misiakiewicz, T.
\newblock {SGD} learning on neural networks: Leap complexity and
  saddle-to-saddle dynamics.
\newblock In \emph{Proceedings of Thirty Sixth Conference on Learning Theory},
  pp.\  2552--2623. PMLR, 2023.

\bibitem[Ahrens et~al.(2012)Ahrens, Li, Orger, Robson, Schier, Engert, and
  Portugues]{ahrens2012brainwide}
Ahrens, M.~B., Li, J.~M., Orger, M.~B., Robson, D.~N., Schier, A.~F., Engert,
  F., and Portugues, R.
\newblock Brain-wide neuronal dynamics during motor adaptation in zebrafish.
\newblock \emph{Nature}, 485\penalty0 (7399):\penalty0 471--477, May 2012.
\newblock ISSN 0028-0836.
\newblock \doi{10.1038/nature11057}.

\bibitem[Akyürek et~al.(2023)Akyürek, Schuurmans, Andreas, Ma, and
  Zhou]{akyurek2023learningalgorithmincontextlearning}
Akyürek, E., Schuurmans, D., Andreas, J., Ma, T., and Zhou, D.
\newblock What learning algorithm is in-context learning? investigations with
  linear models, 2023.
\newblock Preprint \href{https://arxiv.org/abs/2211.15661}{arXiv:2211.15661}
  [cs.LG].

\bibitem[Amadei et~al.(1993)Amadei, Linssen, and
  Berendsen]{amadei1993essential}
Amadei, A., Linssen, A.~B., and Berendsen, H.~J.
\newblock Essential dynamics of proteins.
\newblock \emph{Proteins: Structure, Function, and Bioinformatics}, 17\penalty0
  (4):\penalty0 412--425, 1993.

\bibitem[Antognini \& Sohl-Dickstein(2018)Antognini and
  Sohl-Dickstein]{antognini2018pca}
Antognini, J. and Sohl-Dickstein, J.
\newblock {PCA} of high dimensional random walks with comparison to neural
  network training.
\newblock In \emph{Advances in Neural Information Processing Systems},
  volume~31, 2018.

\bibitem[Bai et~al.(2024)Bai, Chen, Wang, Xiong, and Mei]{bai2024transformers}
Bai, Y., Chen, F., Wang, H., Xiong, C., and Mei, S.
\newblock Transformers as statisticians: Provable in-context learning with
  in-context algorithm selection.
\newblock \emph{Advances in Neural Information Processing Systems}, 36, 2024.

\bibitem[Baldi \& Hornik(1989)Baldi and Hornik]{baldi1989neural}
Baldi, P. and Hornik, K.
\newblock Neural networks and principal component analysis: Learning from
  examples without local minima.
\newblock \emph{Neural Networks}, 2\penalty0 (1):\penalty0 53--58, 1989.

\bibitem[Briggman et~al.(2005)Briggman, Abarbanel, and
  Kristan~Jr]{briggman2005optical}
Briggman, K.~L., Abarbanel, H.~D., and Kristan~Jr, W.
\newblock Optical imaging of neuronal populations during decision-making.
\newblock \emph{Science}, 307\penalty0 (5711):\penalty0 896--901, 2005.

\bibitem[Cammarata et~al.(2020)Cammarata, Olah, Schubert, Goh, Petrov, and
  Carter]{cammarata2020thread}
Cammarata, N., Olah, C., Schubert, L., Goh, G., Petrov, M., and Carter, S.
\newblock Thread: Circuits.
\newblock \emph{Distill}, 2020.
\newblock URL \url{https://distill.pub/2020/circuits}.

\bibitem[Chan et~al.(2024)Chan, Chen, György, and
  Schuurmans]{chan2024understanding}
Chan, B., Chen, X., György, A., and Schuurmans, D.
\newblock Toward understanding in-context vs. in-weight learning, 2024.
\newblock Preprint \href{https://arxiv.org/abs/2410.23042}{arXiv:2410.23042}
  [cs.LG].

\bibitem[Chan et~al.(2022)Chan, Santoro, Lampinen, Wang, Singh, Richemond,
  McClelland, and Hill]{chan2022data}
Chan, S., Santoro, A., Lampinen, A., Wang, J., Singh, A., Richemond, P.,
  McClelland, J., and Hill, F.
\newblock Data distributional properties drive emergent in-context learning in
  transformers.
\newblock \emph{Advances in Neural Information Processing Systems},
  35:\penalty0 18878--18891, 2022.

\bibitem[Chen et~al.(2023)Chen, Lau, Mendel, Wei, and Murfet]{chen2023tms1}
Chen, Z., Lau, E., Mendel, J., Wei, S., and Murfet, D.
\newblock Dynamical versus {Bayesian} phase transitions in a toy model of
  superposition, 2023.
\newblock Preprint \href{https://arxiv.org/abs/2310.06301}{arXiv:2310.06301}
  [cs.LG].

\bibitem[Cunningham \& Yu(2014)Cunningham and Yu]{cunningham2014dimensionality}
Cunningham, J.~P. and Yu, B.~M.
\newblock Dimensionality reduction for large-scale neural recordings.
\newblock \emph{Nature Neuroscience}, 17\penalty0 (11):\penalty0 1500--1509,
  2014.

\bibitem[Edelman et~al.(2024)Edelman, Tsilivis, Edelman, Malach, and
  Goel]{edelman2024statisticalinductionheads}
Edelman, E., Tsilivis, N., Edelman, B.~L., Malach, E., and Goel, S.
\newblock The evolution of statistical induction heads: In-context learning
  markov chains.
\newblock In \emph{The Thirty-eighth Annual Conference on Neural Information
  Processing Systems}, 2024.

\bibitem[Elhage et~al.(2021)Elhage, Nanda, Olsson, Henighan, Joseph, Mann,
  Askell, Bai, Chen, Conerly, DasSarma, Drain, Ganguli, Hatfield-Dodds,
  Hernandez, Jones, Kernion, Lovitt, Ndousse, Amodei, Brown, Clark, Kaplan,
  McCandlish, and Olah]{elhage2021mathematical}
Elhage, N., Nanda, N., Olsson, C., Henighan, T., Joseph, N., Mann, B., Askell,
  A., Bai, Y., Chen, A., Conerly, T., DasSarma, N., Drain, D., Ganguli, D.,
  Hatfield-Dodds, Z., Hernandez, D., Jones, A., Kernion, J., Lovitt, L.,
  Ndousse, K., Amodei, D., Brown, T., Clark, J., Kaplan, J., McCandlish, S.,
  and Olah, C.
\newblock A mathematical framework for transformer circuits.
\newblock \emph{Transformer Circuits Thread}, 2021.

\bibitem[Elhage et~al.(2022)Elhage, Hume, Olsson, Schiefer, Henighan, Kravec,
  Hatfield-Dodds, Lasenby, Drain, Chen, Grosse, McCandlish, Kaplan, Amodei,
  Wattenberg, and Olah]{elhage2022superposition}
Elhage, N., Hume, T., Olsson, C., Schiefer, N., Henighan, T., Kravec, S.,
  Hatfield-Dodds, Z., Lasenby, R., Drain, D., Chen, C., Grosse, R., McCandlish,
  S., Kaplan, J., Amodei, D., Wattenberg, M., and Olah, C.
\newblock Toy models of superposition.
\newblock \emph{Transformer Circuits Thread}, 2022.

\bibitem[Garg et~al.(2022)Garg, Tsipras, Liang, and Valiant]{garg2022what}
Garg, S., Tsipras, D., Liang, P., and Valiant, G.
\newblock What can transformers learn in-context? a case study of simple
  function classes.
\newblock In \emph{Advances in Neural Information Processing Systems}, 2022.

\bibitem[Gelman \& Rubin(1992)Gelman and Rubin]{gelmanrubin}
Gelman, A. and Rubin, D.~B.
\newblock Inference from iterative simulation using multiple sequences.
\newblock \emph{Statistical Science}, 7:\penalty0 457--472, January 1992.
\newblock \doi{10.1214/ss/1177011136}.

\bibitem[Gelman et~al.(2003)Gelman, Carlin, Stern, and
  Rubin]{gelman2003bayesian}
Gelman, A., Carlin, J.~B., Stern, H.~S., and Rubin, D.~B.
\newblock \emph{Bayesian Data Analysis}.
\newblock Chapman and Hall/CRC, 2003.

\bibitem[Geweke(1992)]{geweke1992evaluating}
Geweke, J.
\newblock Evaluating the accuracy of sampling-based approaches to the
  calculations of posterior moments.
\newblock \emph{Bayesian Statistics}, 4:\penalty0 641--649, 1992.

\bibitem[Gissin et~al.(2020)Gissin, Shalev-Shwartz, and
  Daniely]{gissin2019implicit}
Gissin, D., Shalev-Shwartz, S., and Daniely, A.
\newblock The implicit bias of depth: How incremental learning drives
  generalization.
\newblock In \emph{International Conference on Learning Representations}, 2020.

\bibitem[Hagiwara et~al.(1993)Hagiwara, Toda, and Usui]{hagiwara1993}
Hagiwara, K., Toda, N., and Usui, S.
\newblock On the problem of applying {AIC} to determine the structure of a
  layered feedforward neural network.
\newblock In \emph{1993 International Joint Conference on Neural Networks},
  volume~3, pp.\  2263--2266. IEEE, 1993.
\newblock \doi{10.1109/IJCNN.1993.714176}.

\bibitem[Hayward \& De~Groot(2008)Hayward and De~Groot]{hayward2008normal}
Hayward, S. and De~Groot, B.~L.
\newblock Normal modes and essential dynamics.
\newblock \emph{Molecular Modeling of Proteins}, pp.\  89--106, 2008.

\bibitem[He et~al.(2024)He, Doshi, Das, and Gromov]{he2024learningtogrok}
He, T., Doshi, D., Das, A., and Gromov, A.
\newblock Learning to grok: Emergence of in-context learning and skill
  composition in modular arithmetic tasks.
\newblock In \emph{The Thirty-eighth Annual Conference on Neural Information
  Processing Systems}, 2024.

\bibitem[Hess(2000)]{hess2000diffusion}
Hess, B.
\newblock Similarities between principal components of protein dynamics and
  random diffusion.
\newblock \emph{Physical Review E}, 62:\penalty0 8438--8448, Dec 2000.
\newblock \doi{10.1103/PhysRevE.62.8438}.

\bibitem[Hewitt \& Manning(2019)Hewitt and
  Manning]{hewitt-manning-2019-structural}
Hewitt, J. and Manning, C.~D.
\newblock {A} structural probe for finding syntax in word representations.
\newblock In \emph{Proceedings of the 2019 Conference of the North {A}merican
  Chapter of the Association for Computational Linguistics: Human Language
  Technologies, Volume 1 (Long and Short Papers)}, pp.\  4129--4138.
  Association for Computational Linguistics, 2019.

\bibitem[Hoogland et~al.(2024)Hoogland, Wang, Farrugia-Roberts, Carroll, Wei,
  and Murfet]{icl1}
Hoogland, J., Wang, G., Farrugia-Roberts, M., Carroll, L., Wei, S., and Murfet,
  D.
\newblock The developmental landscape of in-context learning, 2024.
\newblock Preprint \href{https://arxiv.org/abs/2402.02364}{arXiv:2402.02364}
  [cs.LG].

\bibitem[Jacot et~al.(2021)Jacot, Ged, {\c{S}}im{\c{s}}ek, Hongler, and
  Gabriel]{jacot2021saddle}
Jacot, A., Ged, F., {\c{S}}im{\c{s}}ek, B., Hongler, C., and Gabriel, F.
\newblock Saddle-to-saddle dynamics in deep linear networks: Small
  initialization training, symmetry, and sparsity, 2021.
\newblock Preprint \href{https://arxiv.org/abs/2106.15933}{arXiv:2106.15933}
  [stat.ML].

\bibitem[Kalimeris et~al.(2019)Kalimeris, Kaplun, Nakkiran, Edelman, Yang,
  Barak, and Zhang]{kalimeris2019increasingcomplexity}
Kalimeris, D., Kaplun, G., Nakkiran, P., Edelman, B., Yang, T., Barak, B., and
  Zhang, H.
\newblock {SGD} on neural networks learns functions of increasing complexity.
\newblock In \emph{Advances in Neural Information Processing Systems},
  volume~32, 2019.

\bibitem[Karpathy(2022)]{nanogpt}
Karpathy, A.
\newblock {NanoGPT}, 2022.
\newblock URL \url{https://github.com/karpathy/nanoGPT}.

\bibitem[Kingma \& Ba(2014)Kingma and Ba]{kingma2014adam}
Kingma, D.~P. and Ba, J.
\newblock Adam: A method for stochastic optimization, 2014.
\newblock Published as a conference paper at ICLR 2015. Preprint
  \href{https://arxiv.org/abs/1412.6980}{arXiv:1412.6980} [cs.LG].

\bibitem[Lau et~al.(2025)Lau, Furman, Wang, Murfet, and Wei]{quantifdegen}
Lau, E., Furman, Z., Wang, G., Murfet, D., and Wei, S.
\newblock The local learning coefficient: A singularity-aware complexity
  measure.
\newblock In \emph{The 28th International Conference on Artificial Intelligence
  and Statistics}, 2025.
\newblock To appear. Preprint
  \href{https://arxiv.org/abs/2308.12108}{arXiv:2308.12108} [stat.ML].

\bibitem[Mao et~al.(2024)Mao, Griniasty, Teoh, Ramesh, Yang, Transtrum, Sethna,
  and Chaudhari]{mao2024manifold}
Mao, J., Griniasty, I., Teoh, H.~K., Ramesh, R., Yang, R., Transtrum, M.~K.,
  Sethna, J.~P., and Chaudhari, P.
\newblock The training process of many deep networks explores the same
  low-dimensional manifold.
\newblock \emph{Proceedings of the National Academy of Sciences}, 121\penalty0
  (12), 2024.

\bibitem[McGrath et~al.(2022)McGrath, Kapishnikov, Toma{\v{s}}ev, Pearce,
  Wattenberg, Hassabis, Kim, Paquet, and Kramnik]{mcgrath2022acquisition}
McGrath, T., Kapishnikov, A., Toma{\v{s}}ev, N., Pearce, A., Wattenberg, M.,
  Hassabis, D., Kim, B., Paquet, U., and Kramnik, V.
\newblock Acquisition of chess knowledge in {AlphaZero}.
\newblock \emph{Proceedings of the National Academy of Sciences}, 119\penalty0
  (47):\penalty0 e2206625119, 2022.

\bibitem[Meyer et~al.(2006)Meyer, Ferrer-Costa, P{\'e}rez, Rueda, Bidon-Chanal,
  Luque, Laughton, and Orozco]{meyer2006essential}
Meyer, T., Ferrer-Costa, C., P{\'e}rez, A., Rueda, M., Bidon-Chanal, A., Luque,
  F.~J., Laughton, C.~A., and Orozco, M.
\newblock Essential dynamics: a tool for efficient trajectory compression and
  management.
\newblock \emph{Journal of Chemical Theory and Computation}, 2\penalty0
  (2):\penalty0 251--258, 2006.

\bibitem[Nanda et~al.(2023)Nanda, Chan, Lieberum, Smith, and
  Steinhardt]{nanda2023progress}
Nanda, N., Chan, L., Lieberum, T., Smith, J., and Steinhardt, J.
\newblock Progress measures for grokking via mechanistic interpretability.
\newblock In \emph{The Eleventh International Conference on Learning
  Representations}, 2023.

\bibitem[Neyshabur et~al.(2017)Neyshabur, Bhojanapalli, Mcallester, and
  Srebro]{neyshabur2017exploring}
Neyshabur, B., Bhojanapalli, S., Mcallester, D., and Srebro, N.
\newblock Exploring generalization in deep learning.
\newblock In \emph{Advances in Neural Information Processing Systems},
  volume~30. Curran Associates, Inc., 2017.

\bibitem[Nguyen \& Reddy(2024)Nguyen and Reddy]{nguyen2024differential}
Nguyen, A. and Reddy, G.
\newblock Differential learning kinetics govern the transition from
  memorization to generalization during in-context learning, 2024.
\newblock Preprint \href{https://arxiv.org/abs/2412.00104}{arXiv:2412.00104}
  [cs.LG].

\bibitem[Olah et~al.(2020)Olah, Cammarata, Schubert, Goh, Petrov, and
  Carter]{olah2020zoom}
Olah, C., Cammarata, N., Schubert, L., Goh, G., Petrov, M., and Carter, S.
\newblock Zoom in: An introduction to circuits.
\newblock \emph{Distill}, 5\penalty0 (3):\penalty0 e00024.001, March 2020.
\newblock URL \url{https://distill.pub/2020/circuits/zoom-in}.

\bibitem[Olsson et~al.(2022)Olsson, Elhage, Nanda, Joseph, DasSarma, Henighan,
  Mann, Askell, Bai, Chen, Conerly, Drain, Ganguli, Hatfield-Dodds, Hernandez,
  Johnston, Jones, Kernion, Lovitt, Ndousse, Amodei, Brown, Clark, Kaplan,
  McCandlish, and Olah]{olsson2022context}
Olsson, C., Elhage, N., Nanda, N., Joseph, N., DasSarma, N., Henighan, T.,
  Mann, B., Askell, A., Bai, Y., Chen, A., Conerly, T., Drain, D., Ganguli, D.,
  Hatfield-Dodds, Z., Hernandez, D., Johnston, S., Jones, A., Kernion, J.,
  Lovitt, L., Ndousse, K., Amodei, D., Brown, T., Clark, J., Kaplan, J.,
  McCandlish, S., and Olah, C.
\newblock In-context learning and induction heads.
\newblock \emph{Transformer Circuits Thread}, 2022.
\newblock URL
  \url{https://transformer-circuits.pub/2022/in-context-learning-and-induction-heads/}.

\bibitem[Panwar et~al.(2024)Panwar, Ahuja, and Goyal]{panwar2024bayesianprism}
Panwar, M., Ahuja, K., and Goyal, N.
\newblock In-context learning through the {Bayesian} prism.
\newblock In \emph{The Twelfth International Conference on Learning
  Representations}, 2024.

\bibitem[Park et~al.(2024)Park, Lubana, Pres, and Tanaka]{park2024competition}
Park, C.~F., Lubana, E.~S., Pres, I., and Tanaka, H.
\newblock Competition dynamics shape algorithmic phases of in-context learning,
  2024.
\newblock Preprint \href{https://arxiv.org/abs/2412.01003}{arXiv:2412.01003}
  [cs.LG].

\bibitem[Phuong \& Hutter(2022)Phuong and Hutter]{phuong2022formal}
Phuong, M. and Hutter, M.
\newblock Formal algorithms for transformers, 2022.
\newblock Preprint \href{https://arxiv.org/abs/2207.09238}{arXiv:2207.09238}
  [cs.LG].

\bibitem[Power et~al.(2022)Power, Burda, Edwards, Babuschkin, and
  Misra]{power2022grokking}
Power, A., Burda, Y., Edwards, H., Babuschkin, I., and Misra, V.
\newblock Grokking: Generalization beyond overfitting on small algorithmic
  datasets, 2022.
\newblock Preprint \href{https://arxiv.org/abs/2201.02177}{arXiv:2201.02177}
  [cs.LG].

\bibitem[Ravent\'{o}s et~al.(2023)Ravent\'{o}s, Paul, Chen, and
  Ganguli]{raventós2023pretraining}
Ravent\'{o}s, A., Paul, M., Chen, F., and Ganguli, S.
\newblock Pretraining task diversity and the emergence of non-bayesian
  in-context learning for regression.
\newblock In \emph{Advances in Neural Information Processing Systems},
  volume~36, pp.\  14228--14246. Curran Associates, Inc., 2023.

\bibitem[Rogers \& McClelland(2004)Rogers and McClelland]{rogers2004semantic}
Rogers, T.~T. and McClelland, J.~L.
\newblock \emph{Semantic Cognition: A Parallel Distributed Processing
  Approach}.
\newblock MIT Press, 2004.

\bibitem[Saxe et~al.(2019)Saxe, McClelland, and Ganguli]{saxe2019mathematical}
Saxe, A.~M., McClelland, J.~L., and Ganguli, S.
\newblock A mathematical theory of semantic development in deep neural
  networks.
\newblock \emph{Proceedings of the National Academy of Sciences}, 116\penalty0
  (23):\penalty0 11537--11546, 2019.

\bibitem[Shinn(2023)]{shinn2023phantom}
Shinn, M.
\newblock Phantom oscillations in principal component analysis.
\newblock \emph{Proceedings of the National Academy of Sciences}, 120\penalty0
  (48):\penalty0 e2311420120, 2023.

\bibitem[Singh et~al.(2024)Singh, Chan, Moskovitz, Grant, Saxe, and
  Hill]{singh2024transient}
Singh, A., Chan, S., Moskovitz, T., Grant, E., Saxe, A., and Hill, F.
\newblock The transient nature of emergent in-context learning in transformers.
\newblock \emph{Advances in Neural Information Processing Systems}, 36, 2024.

\bibitem[Strogatz(1994)]{strogatz}
Strogatz, S.~H.
\newblock \emph{Nonlinear Dynamics and Chaos: With Applications to Physics,
  Biology, Chemistry, and Engineering}.
\newblock Perseus Books, Reading, Massachusetts, 1994.
\newblock ISBN 978-0-201-54344-5.

\bibitem[Vats \& Knudson(2020)Vats and Knudson]{vats2020revisiting}
Vats, D. and Knudson, C.
\newblock Revisiting the {Gelman--Rubin} diagnostic, 2020.
\newblock Preprint \href{https://arxiv.org/abs/1812.09384}{arXiv:1812.09384}
  [stat.CO].

\bibitem[von Oswald et~al.(2023)von Oswald, Niklasson, Randazzo, Sacramento,
  Mordvintsev, Zhmoginov, and Vladymyrov]{von2023transformers}
von Oswald, J., Niklasson, E., Randazzo, E., Sacramento, J., Mordvintsev, A.,
  Zhmoginov, A., and Vladymyrov, M.
\newblock Transformers learn in-context by gradient descent.
\newblock In \emph{Proceedings of the 40th International Conference on Machine
  Learning}, volume 202 of \emph{Proceedings of Machine Learning Research},
  pp.\  35151--35174. PMLR, 2023.

\bibitem[Wang et~al.(2024)Wang, Hoogland, van Wingerden, Furman, and
  Murfet]{lang1}
Wang, G., Hoogland, J., van Wingerden, S., Furman, Z., and Murfet, D.
\newblock Differentiation and specialization of attention heads via the refined
  local learning coefficient, 2024.
\newblock Preprint \href{https://arxiv.org/abs/2410.02984}{arXiv:2410.02984}
  [cs.LG].

\bibitem[Wang(2008)]{wang2008karhunenloeve}
Wang, L.
\newblock \emph{Karhunen-Loeve Expansions and Their Applications}.
\newblock PhD thesis, London School of Economics and Political Science, 2008.
\newblock URL \url{http://etheses.lse.ac.uk/2950/}.

\bibitem[Watanabe(2007)]{watanabe2007almost}
Watanabe, S.
\newblock Almost all learning machines are singular.
\newblock In \emph{IEEE Symposium on Foundations of Computational
  Intelligence}, pp.\  383--388. IEEE, 2007.
\newblock \doi{10.1109/FOCI.2007.371500}.

\bibitem[Watanabe(2009)]{greybook}
Watanabe, S.
\newblock \emph{Algebraic Geometry and Statistical Learning Theory}.
\newblock Cambridge University Press, 2009.

\bibitem[Watanabe(2018)]{watanabe2018}
Watanabe, S.
\newblock \emph{Mathematical Theory of {Bayesian} Statistics}.
\newblock CRC Press, Taylor and Francis group, USA, 2018.

\bibitem[Wei et~al.(2023)Wei, Murfet, Gong, Li, Gell-Redman, and
  Quella]{goodpaper}
Wei, S., Murfet, D., Gong, M., Li, H., Gell-Redman, J., and Quella, T.
\newblock Deep learning is singular, and that's good.
\newblock \emph{IEEE Transactions on Neural Networks and Learning Systems},
  34\penalty0 (12):\penalty0 10473--10486, 2023.
\newblock \doi{10.1109/TNNLS.2022.3167409}.

\bibitem[Welling \& Teh(2011)Welling and
  Teh]{wellingBayesianLearningStochastic2011}
Welling, M. and Teh, Y.~W.
\newblock Bayesian learning via stochastic gradient {Langevin} dynamics.
\newblock In \emph{Proceedings of the 28th International Conference on Machine
  Learning}, 2011.

\bibitem[Xu(2024)]{xu2024hyperparameter}
Xu, A.~K.
\newblock Hyperparameter tuning for local learning coefficient estimation.
\newblock Master's thesis, The University of Melbourne, May 2024.

\end{thebibliography}
\addcontentsline{toc}{section}{References}

\balance


\clearpage
\appendix
\onecolumn
\counterwithin{figure}{section}
\addcontentsline{toc}{part}{Appendix}
\part*{Appendix}

\startcontents[appendix]
\vspace{-2ex}
\printcontents[appendix]{l}{1}{\setcounter{tocdepth}{2}}

\clearpage
\section{Transformer training details}\label{appendix:transformer-details}

\subsection{Architecture}\label{appendix:lr-architecture}

We use the same in-context linear regression transformer architecture as \citet{icl1}---a two-layer transformer modeled after NanoGPT \citetext{\citealp{nanogpt}; see also \citealp{phuong2022formal}} and \citet{raventós2023pretraining} (but with fewer layers).
In more detail, the architecture is a pre-layer-norm decoder-only transformer with a learnable positional embedding. For the primary architecture used in the main body and related appendices, we set $L=2$ layers, $H=4$ attention heads per layer, $d_{\mathrm{embed}}=512$ embedding dimensions and $d_{\mathrm{MLP}}=512$ MLP dimensions, yielding a transformer with the aforementioned $d=2.65$~million trainable parameters.
Further details are given in \cref{tab:lr-hyperparameters}.

We note that \citet{raventós2023pretraining} used $D = 8, K = 16, \sigma^2 = 0.25, L = 8, d_{\text{embed}} = 128, H = 2$.
That is, our models are smaller and wider than those of \citet{raventós2023pretraining}, so some results are not directly comparable.

\subsection{Tokenization}\label{appendix:lr-tokenization}

To run sequences
    $S = (x_1, y_1, \ldots, x_K, y_K)$
through the transformer and produce a sequence of predicted labels
    $\hat y_1, \ldots, \hat y_K$
for each subsequence $S_{\leq 1}, \ldots, S_{\leq K}$ requires an initial encoding or ``tokenization'' step and a final ``projection'' step.
\begin{itemize}
    \item 
        The sequence $S$ is first encoded as a sequence of tokens in $\mathbb{R}^{D+1}$ using the tokenization function $\tau : (\mathbb{R}^D \times \mathbb{R})^K \to (\mathbb{R}^{D+1})^{2K}$:
        \begin{equation*}
        \tau(x_1, y_1, x_2, y_2, \ldots, x_K, y_K) = \left(
            \begin{pmatrix}
                0\\
                \vline \\
                x_1 \\
                \vline
            \end{pmatrix},
            \begin{pmatrix}
                y_1 \\
                0 \\
                \vdots \\
                0    
            \end{pmatrix},
            \begin{pmatrix}
                0\\
                \vline \\
                x_2 \\
                \vline
            \end{pmatrix},
            \begin{pmatrix}
                y_2 \\
                0 \\
                \vdots \\
                0    
            \end{pmatrix},
            \ldots,
            \begin{pmatrix}
                0\\
                \vline \\
                x_{K} \\
                \vline
            \end{pmatrix},
            \begin{pmatrix}
                y_K\\
                0 \\
                \vdots\\
                0
            \end{pmatrix}
        \right).
        \end{equation*}
        Note that this tokenization includes the final $y_K$ even though this is never used as part of the context for a prediction (it is used only as a label).
    
    \item 
        The transformer architecture takes a token sequence in $(\mathbb{R}^{D+1})^{2K}$ as input and outputs a vector of the same shape.
        To extract the predictions $\hat y_k$ for each $k$, we read out the first component of every other token using a projection function $\pi_Y : (\mathbb{R}^{D+1})^{2K} \to \mathbb R^{K}$,
        \begin{equation*}
            \pi_Y
            \left(
                \begin{pmatrix}
                    \hat y_1\\
                    \vdots
                \end{pmatrix},
                \begin{pmatrix}
                    \ .\ \,\\
                    \vdots  
                \end{pmatrix},
                \begin{pmatrix}
                    \hat y_2 \\
                    \vdots  
                \end{pmatrix},
                \begin{pmatrix}
                    \ .\ \,\\
                    \vdots  
                \end{pmatrix},
                \ldots,
                \begin{pmatrix}
                    \hat y_{K}\\
                    \vdots
                \end{pmatrix},
                \begin{pmatrix}
                    \ .\ \,\\
                    \vdots
                \end{pmatrix}
            \right)
            =
            (\hat y_1, \hat y_2, \ldots, \hat y_K).
        \end{equation*}
        We use these extracted predictions in computing the loss as per \cref{section:setting}.
        Note that transformer's outputs for dimensions that are not extracted by $\pi_Y$ are not subject to training (nor are they subject to evaluation).
\end{itemize}
Evaluating the loss of the transformer on all masked context predictions and not just the final token is an important design choice for our results. Given a sufficiently large context, many regression algorithms will arrive at the same prediction on late tokens. By training and then tracking the transformer's early token predictions too, its functional outputs are more representative of its internal structure, giving a richer picture of their essential dynamics with the joint trajectory PCA (see also \cref{appendix:per-token-retreat}).

\subsection{Training}

We train each transformer for $T=150$k steps.
We employ a learning rate scheduler that increases linearly from 0 to $0.003$ over the first $50$k steps and remains constant thereafter. We train without explicit regularization and use the Adam optimizer \citep{kingma2014adam}, with batch-size $B_{\text{train}} = 1024$. Each training run took 3--4 TPU hours with TPUs provided by Google TPU Research Cloud.
Further details are given in \cref{tab:lr-hyperparameters}.

Each model was initialized at the same point in parameter space, $w_0^{M_1} = w_0^{M_2}$ for each $M_1, M_2 \in \mathcal{M}$, drawn once according to default settings in PyTorch. All models were trained to convergence, except for $M = {182, 256}$ whose loss was still decreasing very slowly at the end of training. (We used a fixed training duration across all models for practical comparison).

We note that \citet{raventós2023pretraining} used $B_{\text{train}} = 256, T = 500K$, meaning we train with a larger batch size but for fewer steps. However, the total number of training samples (153M) is comparable to their setting (128M).

\subsection{Checkpoint distribution}\label{appendix:checkpoint-sampling}

We sample checkpoint indices $\mathcal{C} \subseteq \{0, \ldots, T\}$, where $|\mathcal{C}| = 2203$. In particular, we use a combination of linear and log spaced checkpoints, $\mathcal{C} = \mathcal{C}_{\mathrm{linear}} \cup \mathcal{C}_{\mathrm{log}}$. Setting $T = 150$k and $N=1000$ we define 
\begin{equation*}
    \mathcal{C}_{\mathrm{linear}} = \{0, 100, 200, \dots, T\}
    \qquad
    \text{and}
    \qquad
    \mathcal{C}_{\mathrm{log}} = \{ \lfloor T^{\frac{j}{N-1}} \rfloor : j = 0, 1, ..., N-1\}
    .
\end{equation*}
This set comprises 1,500 checkpoints spaced linearly at 100-step intervals, with the remaining checkpoints logarithmically spaced.

This sampling strategy allows us to observe both early-stage rapid changes and later-stage gradual developments throughout the training process.
We investigate the effect of combining log and linear steps on trajectory PCA in \cref{section:ed_checkpoint_sampling}.

\begin{table}[bth]
\centering
\caption{Summary of hyperparameters for the primary transformer of the main body.}
\begin{tabular}{llll}
\toprule
\textbf{Hyperparameter} & \textbf{Category} & \textbf{Description/Notes} & \textbf{Values} \\
\midrule
$N$ & Data & Total \# of training samples & $153{,}600{,}000$ \\
$B_{\mathrm{train}}$ & Data & Batch size during training & 1024 \\
$T$ & Data & \# of training steps & $150{,}000$ \\
$D$ & Data & Dimension of linear regression task (task size) & 8 \\
$K$ & Data & Maximum in-context examples & 16 \\
\(\sigma^2\) & Data & Variance of noise in data generation & $0.125$ \\
$L$ & Model & \# of layers in the model & 2 \\
$H$ & Model & \# of attention heads per layer & 4 \\
\(d_{\mathrm{mlp}}\) & Model & Size of the hidden layer in MLP & 512 \\
\(d_{\mathrm{embed}}\) & Model & Embedding size & 512 \\
\(\mathrm{seed}\) & Misc & Training run seed & 1 \\
Optimizer Type & Optimizer & Type of optimizer & Adam \\
\(\eta\) & Optimizer & Maximum learning rate & 0.003 \\
\(\lambda_\mathrm{wd}\) & Optimizer & Weight Decay & 0 \\
\(\beta_{1,2}\) & Optimizer & Betas & (0.9, 0.999) \\
Scheduler Type & Scheduler & Type of learning rate scheduler & Increase then constant \\
Strategy & Scheduler & Strategy for annealing the learning rate & Linear \\
\% start & Scheduler & Percentage of the cycle when learning rate is increasing & $33.3\%$ \\
\bottomrule
\end{tabular}
\label{tab:lr-hyperparameters}
\end{table}

\clearpage
\section{Transformer evaluation details}\label{appendix:ood}

The full results of our evaluation are shown in \cref{fig:evals}.

\subsection{Evaluation data}

We evaluate the performance of our transformers on several distributions:
\begin{enumerate}
    \item[1.]
        We evaluate the root task performance of our transformers on a dataset
            $\mathcal{D}^{(1)} \sim \q{1}{S}$.
        Note that $\q{1}{S}$ is \emph{not} the training distribution for models with task diversity $M > 1$, but, by construction, the sequences from this distribution are still \emph{in} the training distribution for those models in the sense that task $\task[1]$ has positive support under $\q{M}{\task[t]}$.
    \item[$M$.]
        We also evaluate in-distribution performance of the model trained with task diversity $M$ on a dataset
            $\mathcal{D}^{(M)} \sim \q{M}{S}$.
        Note that this evaluation dataset is different for each model, and it is the same distribution from which the training dataset for that model was sampled.
    \item[$\infty$.]
        As discussed in \cref{section:ood-results}, we evaluate the out-of-distribution (OOD) performance of our transformers on a dataset
            $\mathcal{D}^{(\infty)} \sim \q{\infty}{S}$.
        Note that this is technically in-distribution data (rather than OOD data) for the model trained with task diversity $M=\infty$.
\end{enumerate}
The specific sequences in $\mathcal{D}^{(1)}$, $\mathcal{D}^{(M)}$, and $\mathcal{D}^{(\infty)}$ are sampled independently from training sequences and therefore have (almost surely) not been seen by the models during training.
Each evaluation dataset contains $B=512$ samples. 

\subsection{Critical times}
\label{appendix:tcrit}

\begin{figure}[h!]
    \centering
    \begin{subfigure}[c]{0.59\linewidth}
        \centering
        \vspace{2.5ex}
        \includegraphics[width=\textwidth]{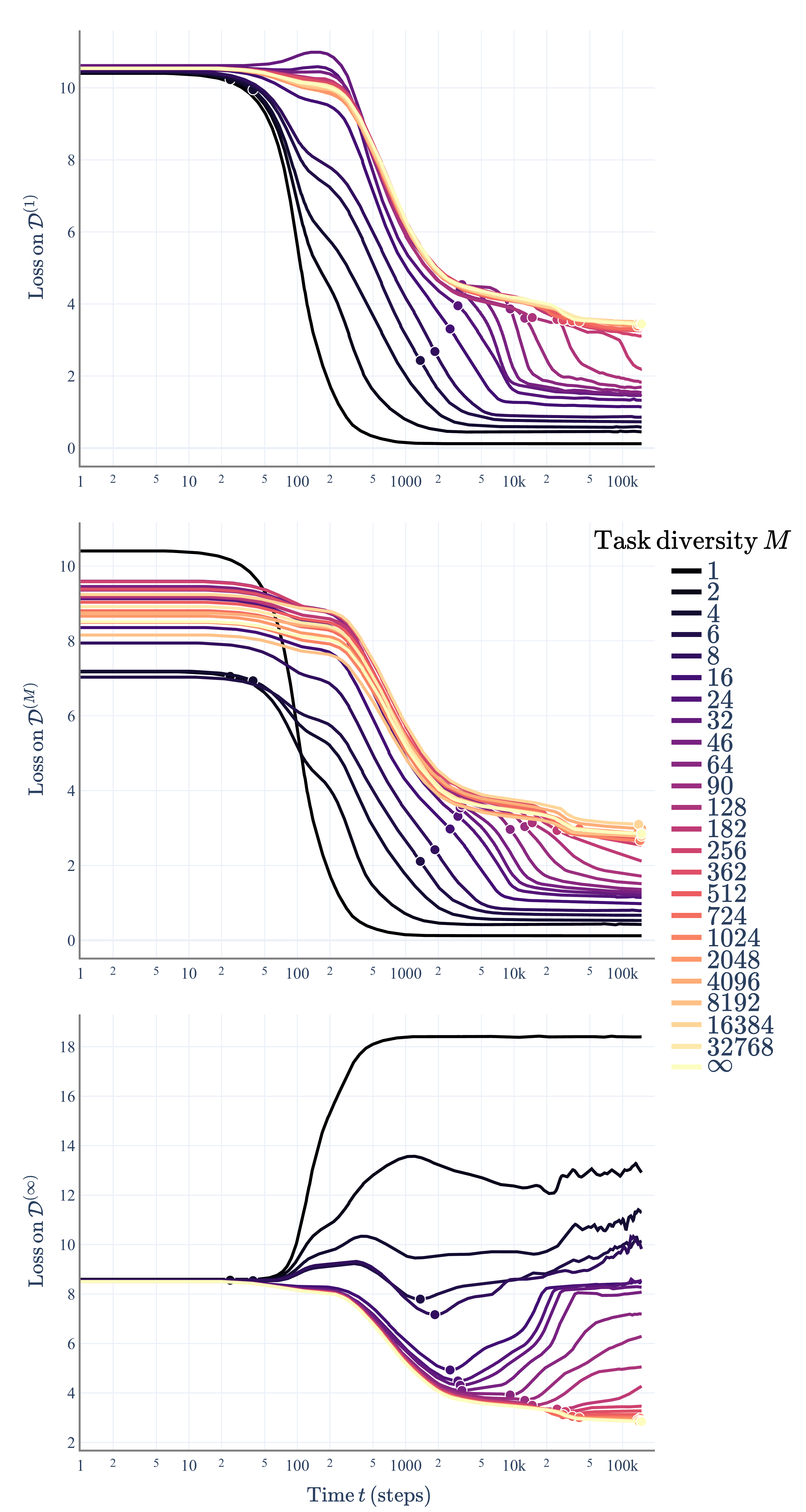}
        \vspace{2.5ex}
        \caption{\label{fig:id-and-ood}\normalsize%
            \emph{(Top):}
                Root task distribution test loss.
            \emph{(Middle):}
                In-distribution test loss.
            \emph{(Bottom):}
                Out-of-distribution test loss (same as \cref{fig:ood-and-pca}). The time $t_M^{\text{crit}}$ is marked with a circle, consistent with the main body. 
        }
    \end{subfigure}
    \hfill
    \begin{subfigure}[c]{0.35\linewidth}
        \centering
            \begin{tabular}{rr}
            \toprule
            \textbf{Task diversity }$M$ & \textbf{Critical time }$t^{\text{crit}}_M$ \\
            \midrule
                  $1$ &   $0.00$k \\
                  $2$ &   $0.02$k \\
                  $4$ &   $0.04$k \\
                  $6$ &   $1.36$k \\
                  $8$ &   $1.86$k \\
                 $16$ &   $2.57$k \\
                 $24$ &   $3.03$k \\
                 $32$ &   $3.18$k \\
                 $46$ &   $3.30$k \\
                 $64$ &   $9.20$k \\
                 $90$ &  $12.54$k \\
                $128$ &  $14.70$k \\
                $182$ &  $24.90$k \\
                $256$ &  $29.70$k \\
                $362$ &  $28.30$k \\
                $512$ &  $34.40$k \\
                $724$ &  $39.90$k \\
               $1024$ & $145.30$k \\
               $2048$ & $138.00$k \\
               $4096$ & $150.00$k \\
               $8192$ & $150.00$k \\
              $16384$ & $141.00$k \\
              $32768$ & $150.00$k \\
             $\infty$ & $150.00$k \\
            \bottomrule
        \end{tabular}
        \caption{\label{tab:t_crit_values}\normalsize%
            Training checkpoints at which loss on $\mathcal{D}^{\infty}$ is minimized for each $f_M$. The curves are lightly smoothed as explained in \cref{section:gaussian_smoothing}. A visual depiction of this table can be seen in \cref{fig:tcrit_comparison}.
        }
    \end{subfigure}
    \caption{Model evaluation and table of critical times for the primary architecture. Each dataset $\mathcal{D}^{(M)}$ is fixed for all timesteps, with $B=512$ samples in each. 
    }
    \label{fig:evals}
\end{figure}

For a given model $f_M(\cdot, w_t)$, we define $\tcrit[M] \in \mathcal{C}$ as the step at which the loss $\ell_B^{\infty}(f_M)$ on $\mathcal{D}^{(\infty)}$ is minimized.
We give the values of $\tcrit[M]$ for the primary architecture in \cref{tab:t_crit_values}.

\clearpage
\section{Trajectory PCA details}\label{appendix:pca}

As described in \cref{section:pca}, we perform PCA on finite-dimensional in-distribution approximations of the functions implemented by each transformer at various checkpoints over training.
Note that this methodology can be used to study joint trajectory PCA of \emph{any} family of trajectories that need not be variations of a training distribution; for example, different random seeds of the same training run. The same technique could also be used to study out-of-distribution behavior by varying the input distribution for which the functional outputs are being measured.

\subsection{Lissajous phenomena of trajectory PCA}\label{appendix:pca-perils}

Caution is required when interpreting PCA results of timeseries. It is a well known fact that the principal components of Brownian motion (i.e.\ a stationary diffusion process) are the Fourier modes, proven in the high-dimensional discrete case in \citet{antognini2018pca}, and in the continuous case in \citet{hess2000diffusion, shinn2023phantom} building on results in Karhunen--Loève theory \citep{wang2008karhunenloeve}. This means that when these trajcetories are plotted in $(\mathrm{PC}j, \mathrm{PC}k)$ space for some $j,k \leq v$, they appear as Lissajous curves. 
In practice this means that ``turning points'' in trajectories projected onto their first few principal components (such as the local minima on the PC1 vs.\ PC2 plots that characterize transient ridge) are not \emph{necessarily} caused by meaningful signals in the data, and should ideally be corroborated by independent sources of information. In our case, we interpret this trend as meaningful only since it corroborates the trends in the OOD loss.

\subsection{Interpretation of PC1 and PC2}\label{appendix:PC1-dev-time}

The most significant axis of variation in timeseries data typically reflects the \emph{development time} of the underlying process - capturing both periods of stasis and rapid change. In the canonical Brownian motion case, this manifests as a monotonic half-cosine first principal component, while more general autocorrelative processes can yield deformed versions of these Fourier modes \citep{shinn2023phantom}. For neural networks trained via SGD, this development time is naturally dictated by gradient descent steps along the loss landscape.

\begin{figure}
    \centering
    \includegraphics[width=1\linewidth]{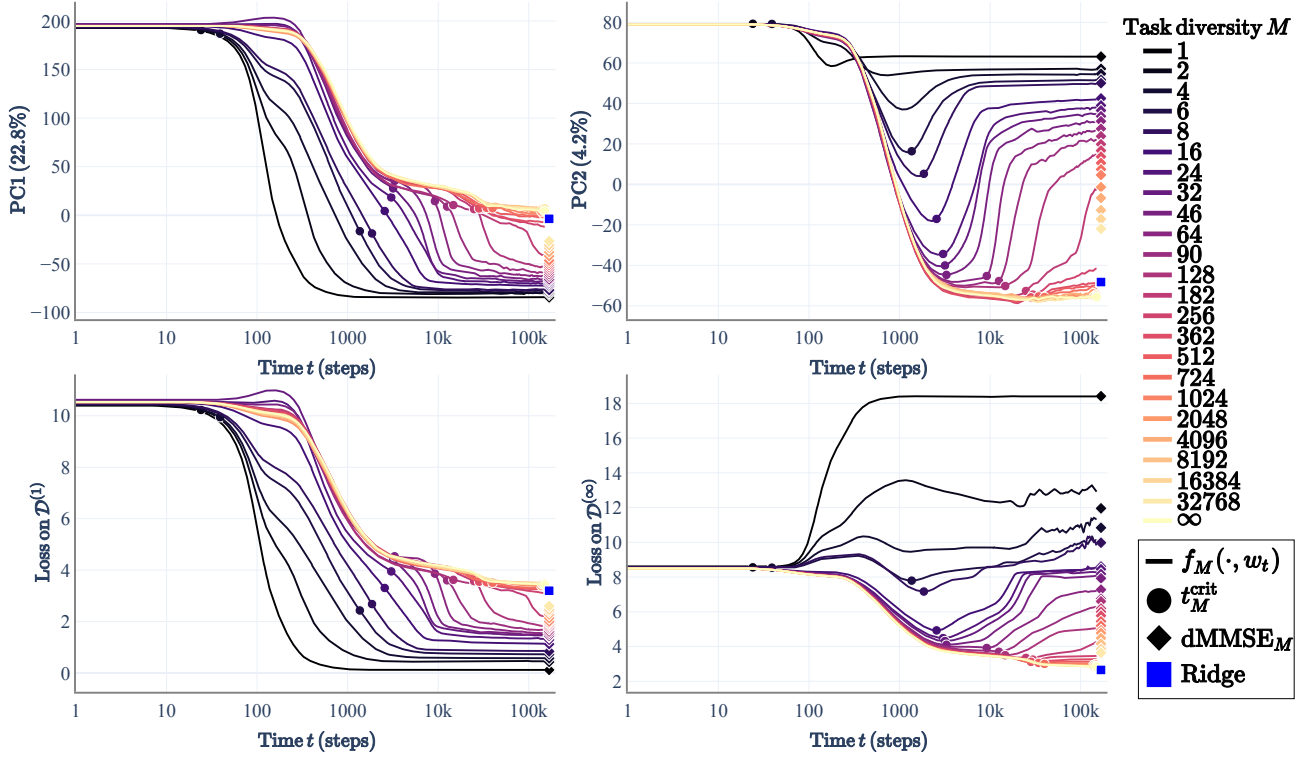}
    \caption{\textbf{Alignment of PC trajectories with loss dynamics.}
    \emph{(Top):} 
    First two principal component trajectories $\gamma_M(t) = (\gamma_M^1(t), \gamma_M^2(t))$---\emph{(left)} and \emph{(right)} respectively---over training time $t$, corresponding to the data of \cref{fig:ood-and-pca}. 
    \emph{(Bottom):} 
    Corresponding loss curves $\ell_B^{1}$ \emph{(left)} and $\ell_B^{\infty}$ \emph{(right)}. There is a striking correlation between each PC1 curve its corresponding root-distribution loss $\ell_B^{1}$, which we provide an explanation of in \cref{appendix:PC1-dev-time}. Transient ridge is seen in the PC2 curves $\gamma_M^2(t)$, whose critical points coincide with the minima of each $\ell_B^{\infty}$ curve which defines $\tcrit$ as in \cref{appendix:tcrit}}.
    \label{fig:D00-TPCs-grid}
\end{figure}

Indeed, since our PCA results in \cref{fig:ood-and-pca} are built from outputs on $\mathcal{D}^{(1)}$, we see in the first column of \cref{fig:D00-TPCs-grid} that PC1 strongly correlates with each model's loss on this dataset. Thus, these loss curves effectively capture the underlying development time that PCA detects, providing a clear interpretation of PC1.

PC2, capturing the second most significant axis of variation, correlates with the distinction between dMMSE and ridge, as evidenced in \cref{fig:D00-TPCs-grid} where the turning point of each $\ell^{\infty}_B$ curve coincides with its corresponding PC2 curve. These PC2-over-time curves provide finer-grained insight into when a model trends toward dMMSE compared to the OOD loss, with deviations depicted more dramatically. We further investigate this interpretation of PC2 in \cref{appendix:per-token-retreat}.

\subsection{Extending to \texorpdfstring{$v=4$}{v=4} components}\label{apx:pca-4}

\begin{figure}
    \centering
    \includegraphics[width=1\linewidth]{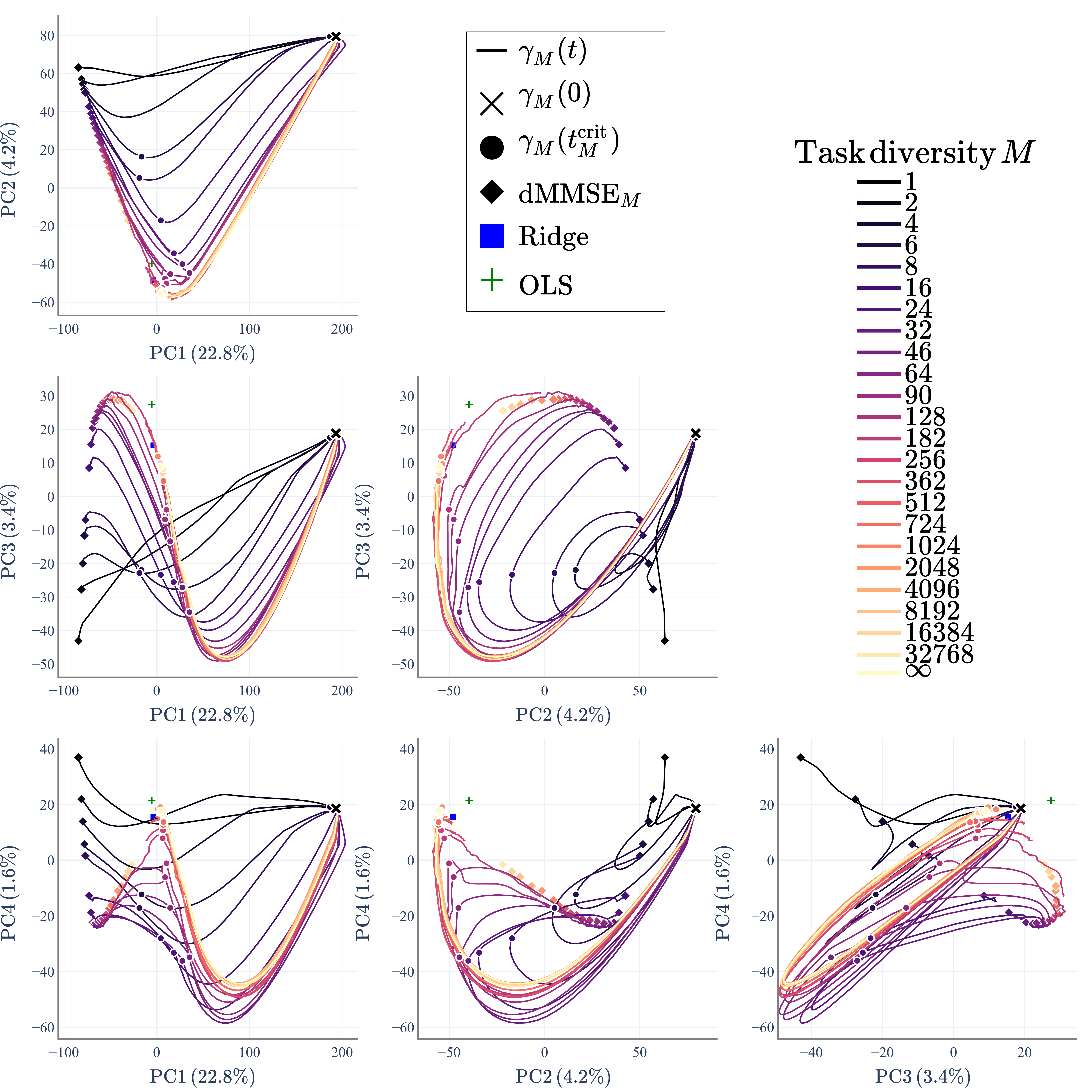}
    \caption{The extension of \cref{fig:ood-and-pca} to all pairs spanned by the first $v=4$ principal components. Note the starkness of the transient ridge phenomenon present in $(\mathrm{PC}2, \mathrm{PC}4)$ \emph{(bottom row, middle column)}. In general, although these shapes arise as deformed Lissajous curves as explained in \cref{appendix:PC1-dev-time}, the deviations between models are meaningful and warrant further investigation. The full PC-over-time curves are shown in \cref{fig:D00-TPCs-smoothed_unsmoothed}.}
    \label{fig:D00-PC-space-4}
\end{figure}

In \cref{fig:D00-PC-space-4} and the right of \cref{fig:D00-TPCs-smoothed_unsmoothed} we investigate the trajectory PCA with $v=4$ principal components. Transient ridge is visible in $(\mathrm{PC}2, \mathrm{PC}4)$ with a sharp deviation occurring in the $f_{128}$ trajectory.

\subsection{Effect of checkpoint distribution}\label{section:ed_checkpoint_sampling}

\begin{figure}
    \centering
    \includegraphics[width=1\linewidth]{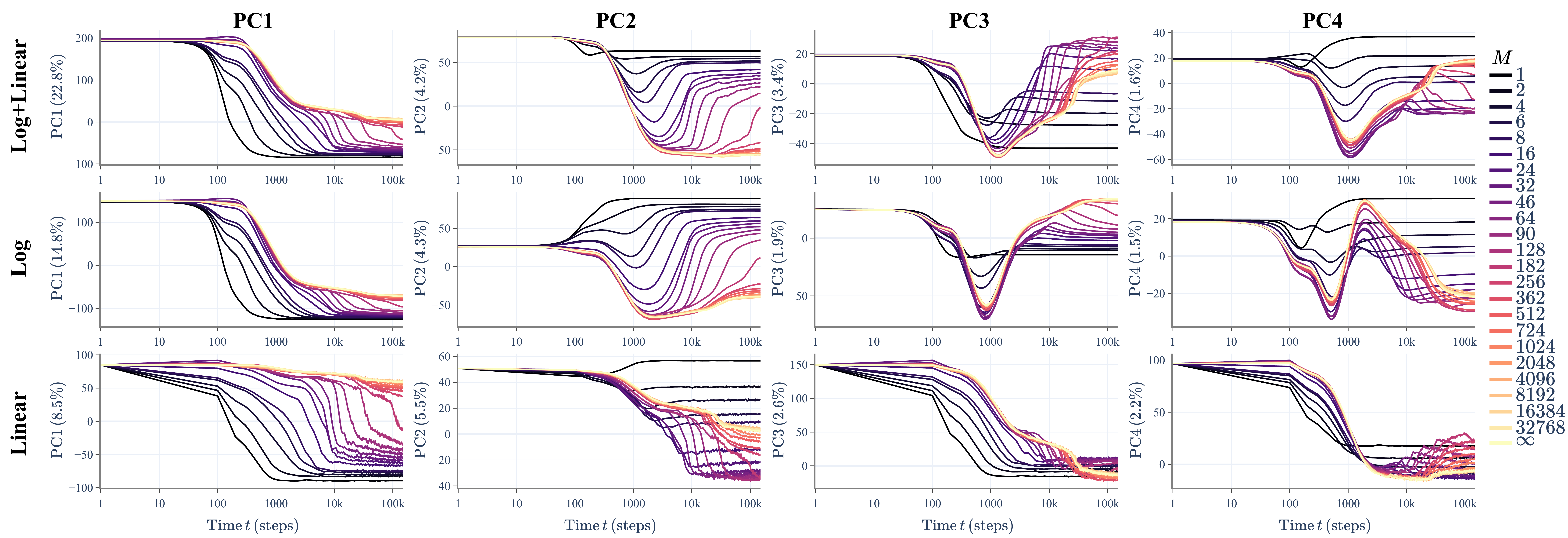}
    \caption{\textbf{Effect of checkpoint sampling.} Checkpoints sampling with $\mathcal{C}$ \emph{(top)}, $\mathcal{C}_{\mathrm{log}}$ \emph{(middle)} and $\mathcal{C}_{\mathrm{linear}}$ \emph{(bottom row)}. The first two rows are smoothed with a Gaussian kernel with variance $20^2$, whereas the last row is smoothed with variance $3^2$ due to the high variance over early checkpoints.}
    \label{fig:D00-TPCs-log-linear}
\end{figure}

In \cref{fig:D00-TPCs-log-linear} we test the effect on the PCA results if checkpoints are sampled only from $\mathcal{C}_{\mathrm{log}}$ versus $\mathcal{C}_{\mathrm{linear}}$ versus the combined $\mathcal{C}$ (cf.~\cref{appendix:checkpoint-sampling}).
Interestingly, the different sampling methods do yield fairly different results. The trajectory PCA using $\mathcal{C}$ are similar to those of $\mathcal{C}_{\mathrm{log}}$, indicating that log-sampling is perhaps sufficient to see the key features of the trajectories in the essential subspace. On the other hand, the curves obtained with $\mathcal{C}_{\mathrm{linear}}$ are different enough as to suggest that different modes of variation dominate over linear time.

\subsection{Number of contexts for PCA features} \label{section:PCA-batch-size}

\begin{figure}
    \centering
    \includegraphics[width=1\linewidth]{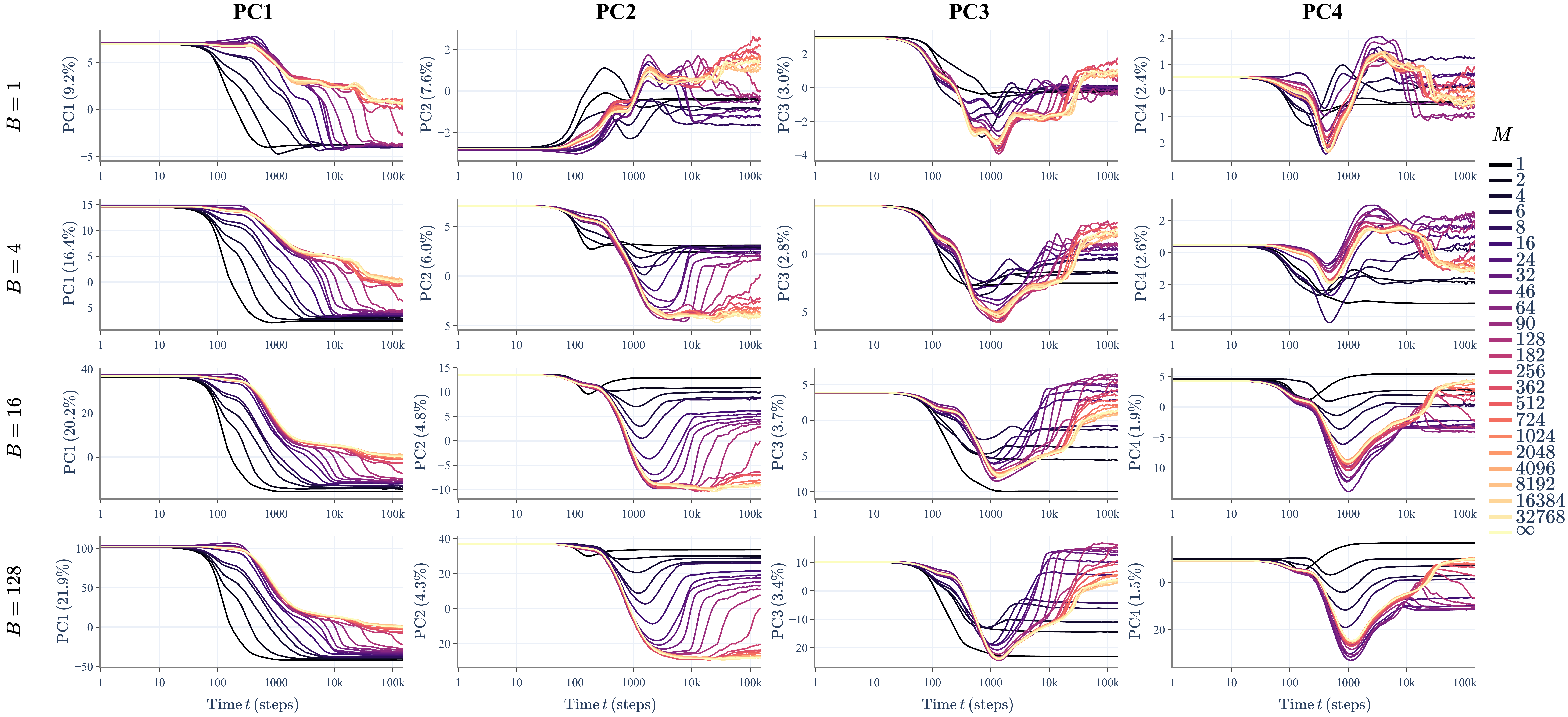}
    \caption{\textbf{Effect of number of context examples on PCA.} The first four PCs over time for $B \in \{1, 4, 16, 128\}$ (such that each dataset is a subset of the next, so $\mathcal{D}^{(1)}_1 \subset \mathcal{D}^{(1)}_4$, and so on). Note how there is relative stability in the trajectory PCA for $B \geq 16$ }
    \label{fig:D00-TPCs-batch_size}
\end{figure}

In the main body, the size $B$ of the input contexts in $\mathcal{D}^{(1)}$ defining the feature space of the trajectory PCA (see \cref{section:pca}) is set to $B=512$. In \cref{fig:D00-TPCs-batch_size} we show the convergence of the trajectories as $B$ increases. The convergence is quite fast, with $B=16$ already somewhat resembling those of $B=512$ (see \cref{fig:D00-TPCs-smoothed_unsmoothed}).

\clearpage
\section{Smoothing details for loss and PC curves}\label{section:gaussian_smoothing}

\begin{figure}[h!]
    \centering
    \includegraphics[width=0.88\linewidth,trim={0 5mm 0 47mm},clip]{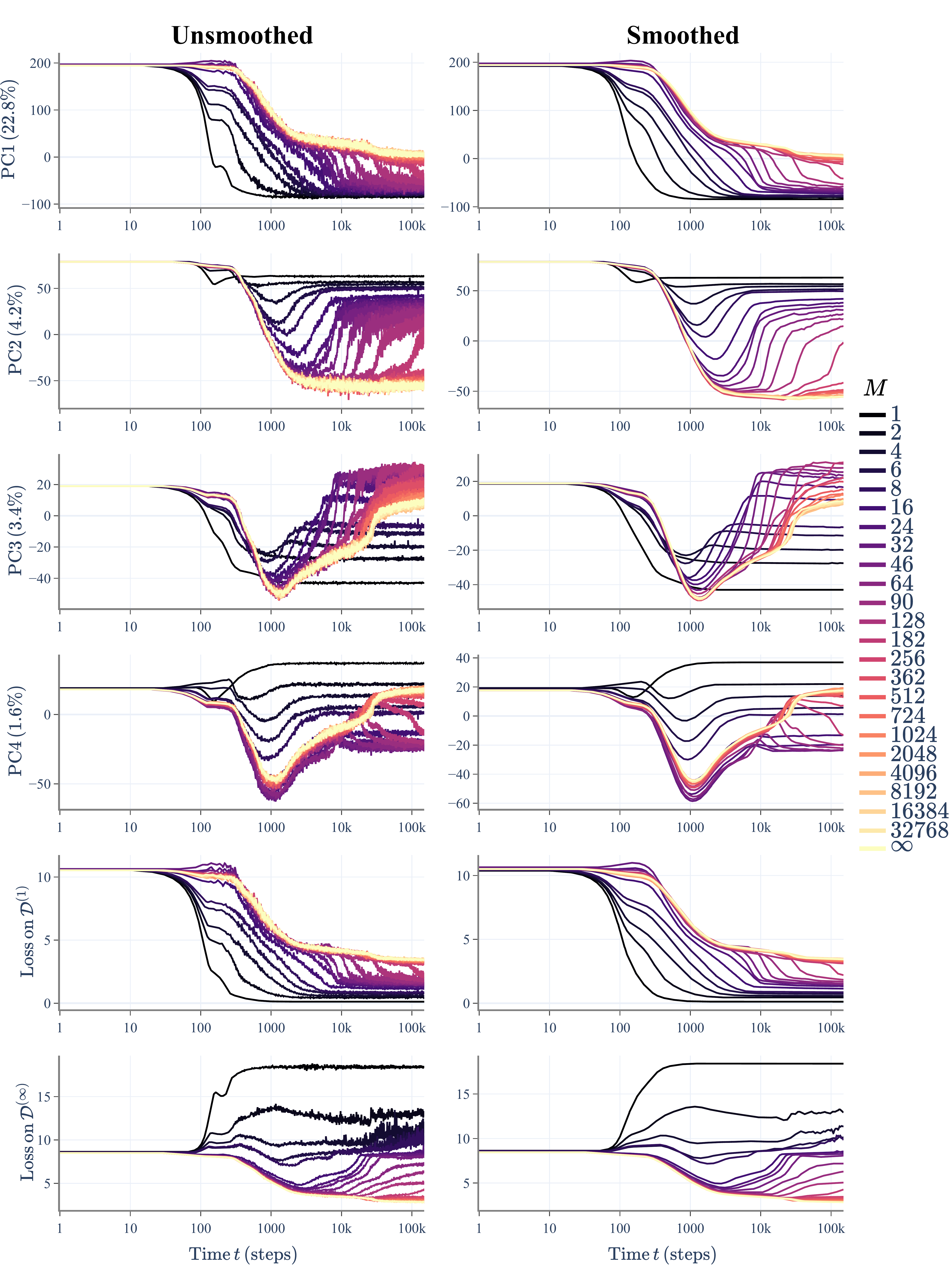}
    \caption{Raw \emph{(left)} vs.\ smoothed \emph{(right)} PC-over-time and loss curves.}
    \label{fig:D00-TPCs-smoothed_unsmoothed}
\end{figure}

We apply smoothing with a Gaussian kernel, with standard deviation 20, to each of the loss curves $\ell^M_B$ and PC-over-time curves $\gamma_M^j$ for all $M \in \mathcal{M}$, and $j \leq 4$. These smoothed curves are used to construct the data in \cref{fig:ood-and-pca} and associated figures in \cref{section:pca}.
\Cref{fig:D00-TPCs-smoothed_unsmoothed} shows the raw curves for comparison.
The motivation for applying the smoothing is to aid in visually distinguishing similar curves, as well as to obtain a reliable estimate of each $\tcrit[M]$ value.

\clearpage
\section{Per-token analysis}\label{appendix:per-token-retreat}

\begin{figure}
    \centering
    \begin{subfigure}[c]{0.49\linewidth}
        \centering
        \includegraphics[width=\linewidth]{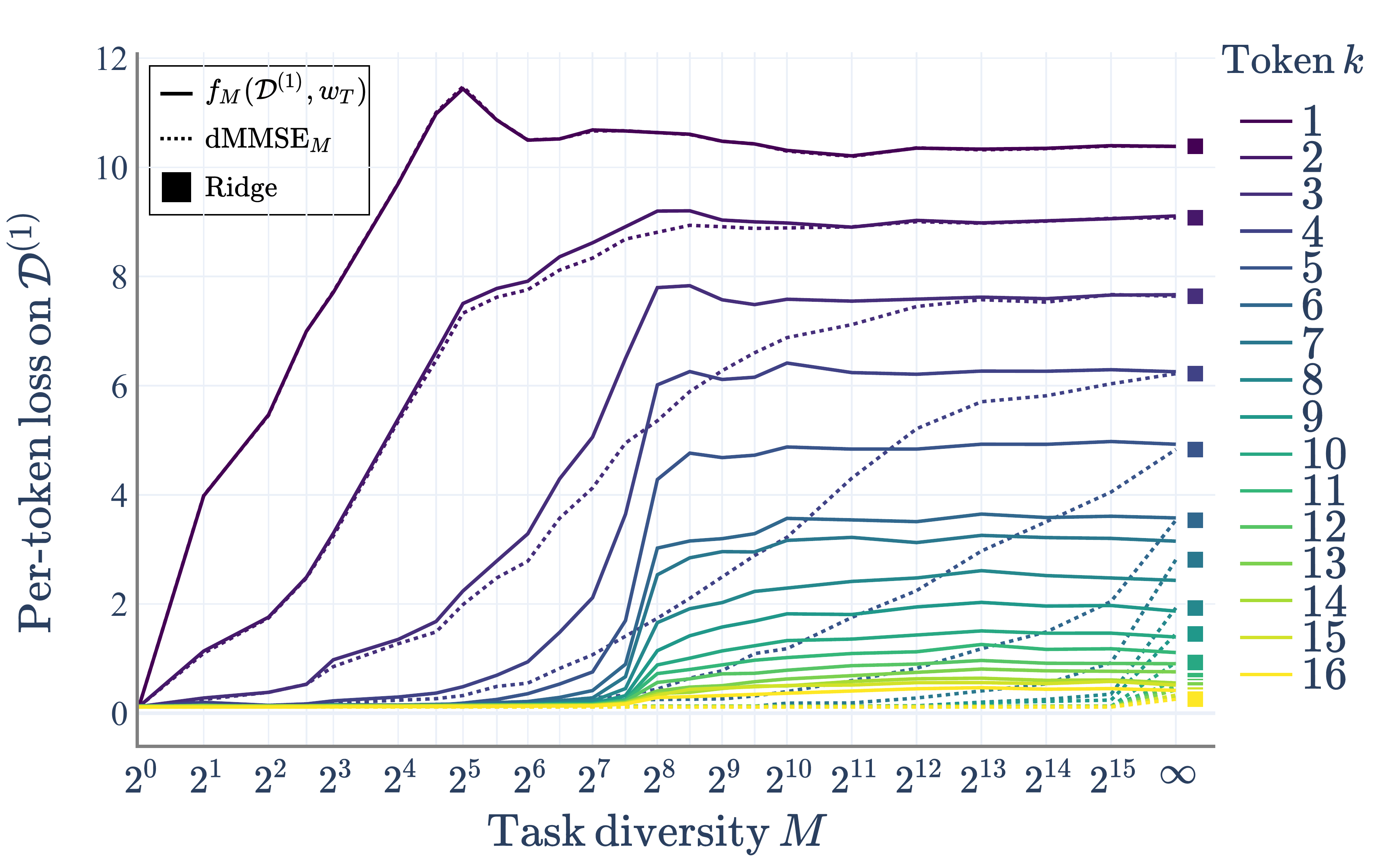}
        \caption{Per-token loss versus $M$ for models at end of training $f_M(\mathcal{D}^{(1)}, w_T)$ (solid line), $\mathrm{dMMSE}_M$ (dashed line), and ridge (squares). The loss is evaluated on $\mathcal{D}^{(1)}$, consistent with \cref{fig:ood-and-pca}. Note the greatest variation across $M$ is in the early tokens $k\leq 5$.}
        \label{fig:D00-pertoken-loss-vs-M}
    \end{subfigure}
    \hfill
    \begin{subfigure}[c]{0.49\linewidth}
        \centering
        \vspace{1.5ex}
        \includegraphics[width=\linewidth]{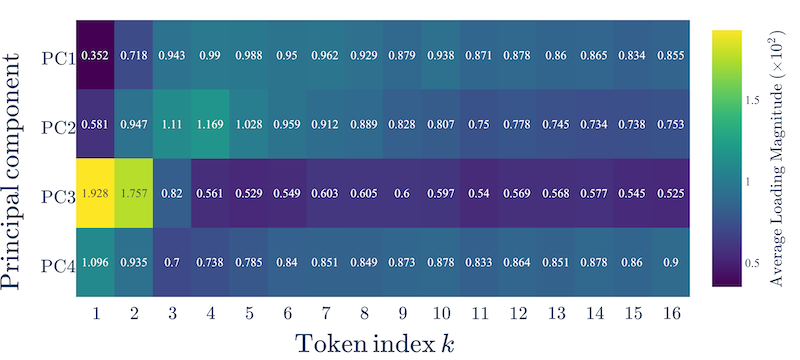}
        \vspace{1.5ex}
        \caption{Average loading magnitude $A$ (equation~\cref{eq:average-loading-magnitude}) across token positions. PC2 is most influenced by tokens $k=3,4,5$, PC3 by $k=1,2$, while later tokens are the main contributors to PC4, nearly inverting PC2's pattern.}
        \label{fig:D00-PC_token_heatmap}
    \end{subfigure}
    \caption{Token-wise analysis of loss and principal component contributions.}
\end{figure}

It is natural to ask, which features (columns of $F_{\mathcal{M}}$) are most responsible for the transient ridge phenomenon? Given the structure of our feature space, with predictions for each regression example $k \in \{1, \dots, K\}$ across $B$ contexts from $\taskdist[1]$, we expect an implicit structure dictated by token positions. As shown in \cref{fig:D00-pertoken-loss-vs-M}, most of the variance across $M$ is driven by early tokens $k \leq 5$, where dMMSE and ridge predictions differ most significantly. 

To verify this, \cref{fig:D00-PC_token_heatmap} examines the contribution of each token to each PC by calculating the average loading magnitude. For PC $i=1,\dots,v$ and token $k=1,\dots,K$, the average loading magnitude $A_{i,k}$ is 
\begin{align} \label{eq:average-loading-magnitude}
A_{i,k} = \frac{1}{B} \sum_{b=1}^B |V_{(b-1)K + k, i}|
\end{align}
where $V_{j,i}$ is the loading of PC $i$ against feature $j \in \{1,\dots, BK\}$, as defined in \cref{sec:PCA_of_F_M}. 

As predicted, we find that tokens $k =3,4,5$ contribute most to $\mathrm{PC}2$ and therefore are the features that drive the transient ridge phenomenon. Interestingly, $k=1,2$ contribute strongly to $\mathrm{PC}3$, while later tokens are the main contributors to $\mathrm{PC}4$, almost inverting the pattern of $\mathrm{PC}2$, aiding in the interpretation of the sharp retreat present in (PC2, PC4) space. 

This analysis suggests that the model improves its loss primarily by adjusting its computation on these early tokens, where the distinction between specialization (\dMMSE[M]) and generalization (ridge) is most pronounced.

\clearpage
\section{LLC estimation details}\label{appendix:llc}

We offer additional details on our use of SGLD for approximating the expectation in equation~\cref{eq:llc-estimator}, which we refer to as LLC estimation. Note that this appendix also pertains to estimation of LLC at parameters from throughout training, as analyzed in \cref{appendix:clean-v-messy}, not only fully-trained parameters, as analyzed in the main text.

\subsection{Interpreting LLC estimates}

For a full formal definition of the LLC, a derivation of the estimator, and experiments validating the soundness of the estimation technique in simpler models, we refer the reader to \citet{quantifdegen}.

For our purposes, it suffices to note that the form of the estimator can be intuitively understood as a degeneracy-aware effective parameter count---the more degenerate directions in the loss landscape near $w^*$, the more ways for the SGLD sampler to find points of low loss, the lower the estimate.

As another source of intuition, we note that equation~\cref{eq:llc-estimator} resembles a PAC-Bayes ``expected sharpness'' measure, as used by \citet{neyshabur2017exploring}. The LLC estimator uses a localized Gibbs posterior in place of the distribution of perturbations around the input parameter.

\subsection{Estimating the LLC with SGLD}

We adopt the approach of \citet{quantifdegen} in using stochastic gradient Langevin dynamics \citep[SGLD;][]{wellingBayesianLearningStochastic2011} to estimate the expectation value of the loss in the LLC estimator. For a given weight configuration $w^*$, we generate $C$ independent chains, each consisting of $T_\text{SGLD}$ steps. Each chain $c$ produces a sequence of weights $\{w_\tau^{(c)}\}_{\tau=1}^{T_\text{SGLD}}$. We then estimate the expectation $\mathbb{E}_\beta[\mathcal{O}(w)|w^*,\gamma]$ of an observable $\mathcal{O}$ using:

\begin{equation}
\frac{1}{CT_\text{SGLD}} \sum_{c=1}^C \sum_{\tau=1}^{T_\text{SGLD}} \mathcal{O}(w_\tau^{(c)}) \,.
\end{equation}

For the estimations studied here, we include a burn-in period. Within each chain (omitting the chain index $c$ for clarity), we generate samples as follows:

\begin{align}
w_{\tau+1} &= w_\tau + \Delta w_\tau, \\
w_1 &= w^*,
\end{align}

where the step $\Delta w_\tau$ is computed using the SGLD update:

\begin{equation}
\Delta w_\tau = \frac{\epsilon}{2}\left(\beta n \nabla \ell_m^{(\tau)}(w_\tau) + \frac{\gamma}{2}(w_\tau - w^*)\right) + \mathcal{N}(0,\epsilon)
\end{equation}

For each step $\tau$, we sample a mini-batch of size $m$ and use the associated empirical loss $\ell_m^{(\tau)}$ to compute the gradient. We follow \citet{quantifdegen} in recycling the mini-batch losses $\ell_m(w_\tau^{(c)})$ computed during SGLD for the expectation average. 

\subsection{SGLD hyperparameter tuning}

For local learning coefficient estimation, we sample $8$ independent chains with 4000 steps per chain, of which the first 2500 are discarded as a burn-in, after which we draw observations once per step, at a temperature $n\beta=30$, $\epsilon=0.00005$, and $\gamma=0.01$, over batches of size $m=1024$. Local learning coefficient estimation takes on the order of a single TPU-hour per training run.  

\begin{table}[h]
\centering
\caption{\textbf{LLC estimation hyperparameters}. A summary of the hyperparameters involved in estimating the local learning coefficient.}
\label{tab:lr-sgld-hyperparameters}
\begin{tabular}{cccc}
\toprule
\textbf{Hyperparameter} & \textbf{Category} & \textbf{Description/Notes} & \textbf{Values}\\
\midrule
C & Sampler & \# of chains & 8\\
$T_{\mathrm{SGLD}}$ & Sampler & Total \# of SGLD draws / chain& 1500\\
 $T_\text{burn-in}$& Sampler& \# of burn-in steps&2500\\
$\epsilon$ & SGLD & Step size & 0.00005\\
$\gamma$& SGLD & Localization strength & 0.01\\
$n\beta$& SGLD & Effective sample size (/inverse temperature)& 30\\
$m$ & SGLD & The size of each SGLD batch & 1024\\ 
\bottomrule
\end{tabular}
\end{table}

\begin{figure}
    \centering
    \includegraphics[width=\linewidth]{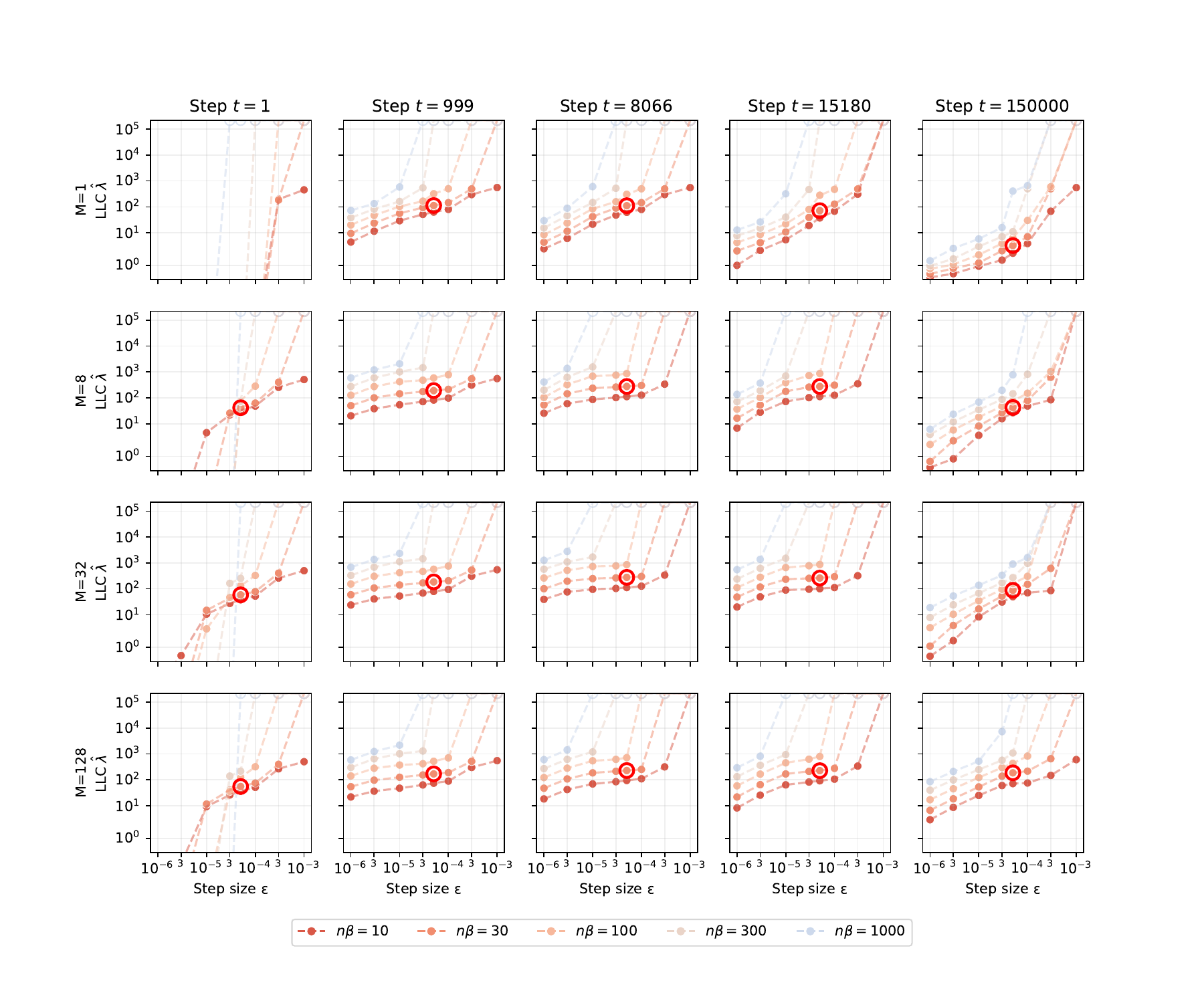}
    \caption{\textbf{LLC hyperparameter calibration sweep.}  Results of sweeping over $n\beta$ and $\epsilon$ for a selection of different training runs (rows) at different training steps (columns). The red circle indicates the final LLC estimation hyperparameters we ended up using. Open circles on the top y axis boundary indicate that those chains diverged. For these sweeps, we ran chains with 1000 draws after 500 burn-in steps. Final LLC estimates involve a longer chain length.}
    \label{fig:llc-calibration}
\end{figure}

To tune $\epsilon$ and $n\beta$, we perform a grid sweep over $\epsilon$, and $n\beta$ over a small subset of training runs and checkpoints, and plot the LLC estimate as a function of $\epsilon$, as depicted in \cref{fig:llc-calibration}. Within these plots, we look for a range of hyperparameters for which $\hat\lambda$ is independent of $\epsilon$ (which shows up as a characteristic plateau). This shows up as an ``elbow'' in Figure \ref{fig:llc-calibration} \citep{xu2024hyperparameter}. Subject to these constraints, we maximize $n\beta$ to get as close as possible to the ``optimal'' $n\beta=n/\log n$, and then maximize $\epsilon$ to reduce the necessary chain length. 

\begin{figure}
    \centering
    \includegraphics[width=\linewidth]{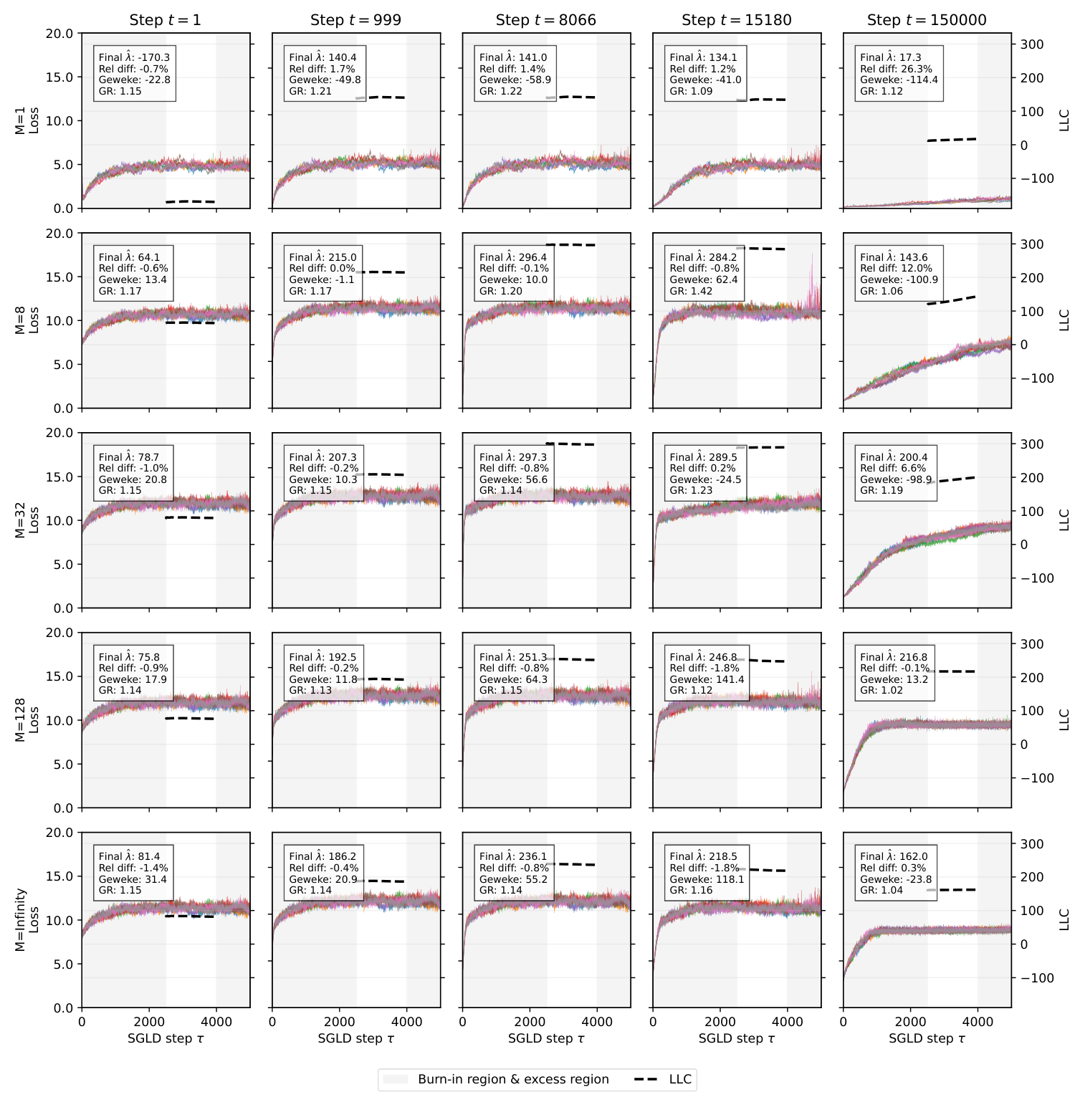}
    \caption{\textbf{SGLD loss traces.} For the hyperparameter values indicated by a red circle in \cref{fig:llc-calibration} ($n\beta=30$, $\epsilon=5e-5$), we plot SGLD loss traces to visually estimate a suitable burn-in period (indicated with the grey background) and minimum chain length (ultimately discarding an excess region, also in grey), and validate this choice with several diagnostics as described in \cref{sec:diagnostics}.}.
    \label{fig:loss-traces}
\end{figure}

\subsection{SGLD diagnostics}
\label{sec:diagnostics}
We employ multiple diagnostics to assess convergence and determine appropriate burn-in periods and numbers of samples for our SGLD chains.

\textbf{The Geweke diagnostic} \citep{geweke1992evaluating} compares the means of the initial 10\% and the last 50\% of the trajectory (after burn-in). Typically, values between $[-2, 2]$ indicate a suitable burn-in period. However, in our context, we found this diagnostic overly sensitive to minor trends, often producing large values despite minimal practical differences in LLC estimates.

To address this limitation, we introduce the \textbf{Relative Percentage Difference (RPD)}, calculated as:

\begin{equation}
\text{RPD} = \frac{\text{Mean}_{\text{last 50\%}} - \text{Mean}_{\text{first 10\%}}}{\text{Mean}_{\text{overall}}} \times 100\%
\end{equation}

The RPD provides a more intuitive measure of practical significance, typically showing differences of less than 1\% even when Geweke values are large.

We also use \textbf{the Gelman-Rubin statistic} to assess convergence across different chains. Values below 1.1 or 1.2 generally indicate suitable convergence \citep{gelmanrubin, gelman2003bayesian, vats2020revisiting}.

Representative loss traces for our chosen hyperparameters are shown in Figure \ref{fig:loss-traces}. Visual inspection suggests healthy, stabilized traces for most runs. The final checkpoints of low-$M$ training runs appear to be an exception and are still underconverged (top-right). The RPD confirms these observations, with values below 1\% for 16/25 checkpoints and below 2\% for 22/25 checkpoints. The exceptions align with visually less converged cases.

Interestingly, the Geweke diagnostic suggests most chains are far from converged, with values well outside the $[-2, 2]$ range. We attribute this to the diagnostic's sensitivity to minute trends when samples cluster tightly. Given the practical insignificance of these trends (as judged visually and by RPD), we opt to disregard these extreme Geweke values.

Our hyperparameter choice represents a compromise. While it is challenging to select parameters that perform optimally across all training runs, we aim for a consistent choice of hyperparameters to allow fair comparison. Based on visual inspection of loss traces, $\epsilon$-insensitivity criteria, and these diagnostic checks, we believe our chosen hyperparameters are appropriate even if, for example, individual checkpoints may be underconverged.

\clearpage
\section{Which solutions govern development?}
\label{appendix:clean-v-messy}

In \cref{section:slt}, we referred to \emph{solutions} $u_\infty$ and $v_M$, representing ridge and \dMMSE[M] respectively, and explained transient ridge in terms of the roles that the loss and LLC of these solutions play in the evolving loss/complexity tradeoff in Bayesian internal model selection.
In this appendix, we discuss two alternative candidates for the solutions that govern the development of our transformers.

\subsection{Candidate~1: End-of-training parameters}

The most straight-forward assumption, adopted in the main text, is that the development of our transformers is governed by the solutions to which our transformers converge. In particular, we assume there is a parameter $u_\infty \in \mathcal{W}$ implementing an approximation of ridge to which high-$M$ transformers converge, and there is a family of parameters $v_M \in \mathcal{W}$ for $M \in \mathbb{N}$ implementing approximations of \dMMSE[M] to which low- and intermediate-$M$ transformers converge (though high-$M$ transformers never do).

While we have shown that the in-distribution behavior of some of our transformers approximately matches that of dMMSE or ridge at convergence, it is an additional assumption to suppose that the loss and complexity of these parameters govern development.
The main text outlines how, under this assumption, the loss/complexity tradeoff explanation lines up with our observations (though, since the trajectories do not converge to \dMMSE[M] for high $M$, we have no obvious way to estimate $\lambda(v_M)$ in that case).

\subsection{Candidate~2: Intermediate parameters}

We now develop an alternative model of the development of our transformers. In brief, we first note that the emergence of \dMMSE[M] or ridge (as the case may be) is not necessarily the last significant internal development in our transformers. In particular, there may be some other parameter that governs the trajectory during the development of ridge, after which the development proceeds to be governed by the parameter to which the transformer eventually converges.

It is difficult to test this story, since we have no obvious way to access such an intermediate governing parameter and measure its LLC. However, we can gain some insight by applying a different methodology of monitoring the LLC of our transformers over the entirety of their development, on the assumption that we will see changes in the LLC playing out during the development.
This methodology has been pioneered by \citet{icl1}, though it is not without its own limitations as discussed by \citet{icl1}---in short, we note that while arbitrary transformer checkpoints may not correspond to local minima, the LLC \emph{estimator} of \citet{quantifdegen} is well-defined for any parameter, and with sufficient hyperparameter calibration (particularly the localization strength) we obtain stable estimates. See \cref{appendix:llc} for more details on LLC estimation calibration.

\begin{figure}[h!]
    \centering
    \includegraphics[width=\linewidth]{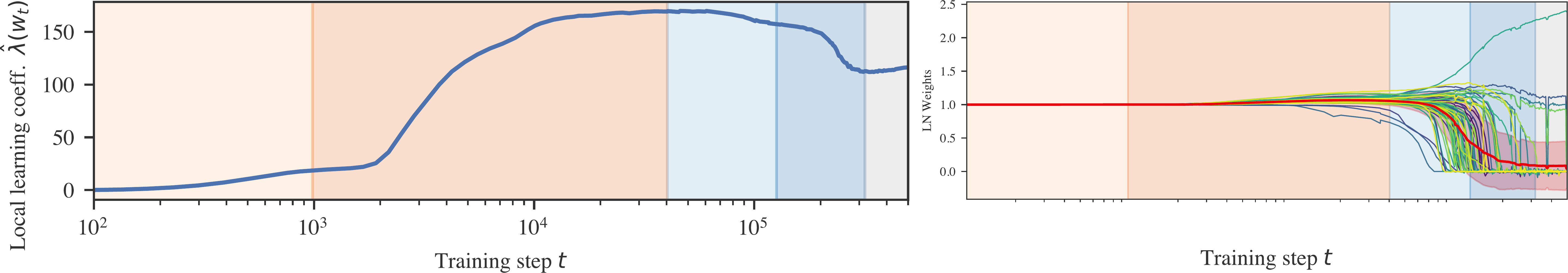}
    \caption{\label{fig:hoogland}%
        \textbf{Reproduced with permission from \citet{icl1}.}
        \emph{(Left):}
            Estimated LLC of $M=\infty$ transformer parameters over training time. Plateaus separate training into developmental stages. The stages are indicated with colored rectangles. The first two stages pertain to LLC increases, while the third and fourth pertain to LLC decreases.
        \emph{(Right):}
            Weights in a transformer layer normalization module over training time. The third and fourth stages coincide with a ``collapse'' of most of these weights to zero, concentrating the computation onto particular dimensions of the residual stream. A similar collapse also occurs for several other internal modules during the final stage.
    }
\end{figure}

Thus, we estimate the LLC over training for our family of models of varying $M$.
Before analyzing the LLC curves for our models, we recall in some detail the results from \citet{icl1}, who studied similar in-context linear regression models in the $M=\infty$ case, and found that their development can be divided into several \emph{developmental stages}, including the following:
\begin{itemize}
    \item 
        During the first developmental stage, the transformer rapidly learns to predict approximately zero for every context. Since $\hat\task_1^\infty = \mathbf{0}$, this is the optimal context-independent prediction.
    \item 
        During the second stage, the transformer begins to make use of the context for its predictions, coinciding with a significant LLC increase.
    \item 
        During the third and fourth stages, the transformer's behavior on in-distribution inputs remains roughly stable, but it specializes to Gaussian-distributed tasks, sacrificing its performance on extremely rare tasks with high magnitudes. These stages coincide with LLC decreases.
\end{itemize}
\citet{icl1} also observed that in the third and fourth stage, various internal modules in the transformer ``collapse,'' focusing their computation on a subset of dimensions. Therefore we refer to these stages as ``collapse'' stages.
We reproduce two indicative figures from \citet{icl1} for comparison with our own models in \cref{fig:hoogland}.
Note that, compared to our architecture, \citet{icl1} used a 2-layer transformer with a smaller embedding size and so a much smaller number of parameters, and they also used different training parameters including the total number of training steps.

\newpage

\begin{figure}[h!]
    \centering
    \includegraphics[width=0.6\linewidth]{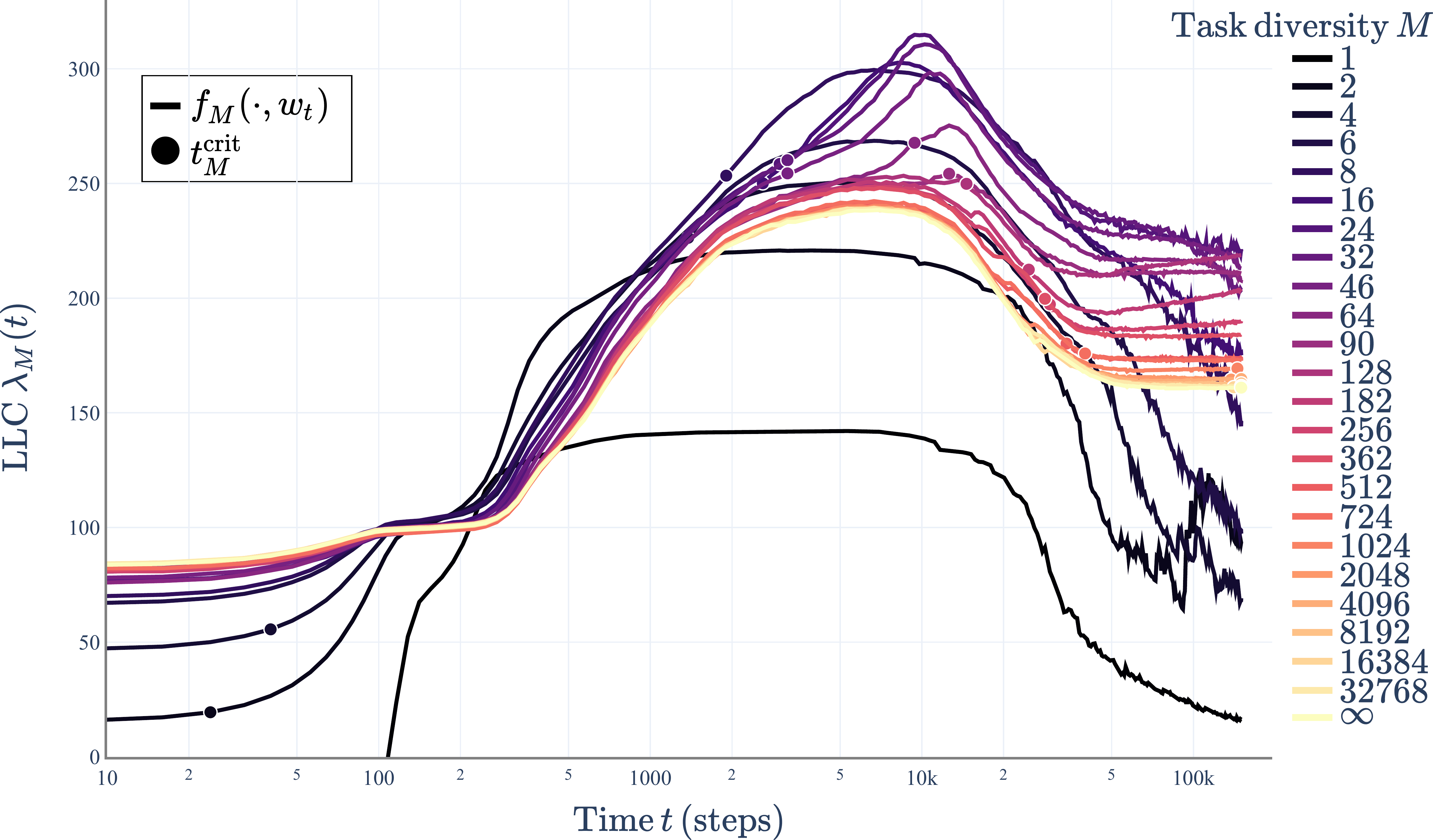}
    \caption{\label{fig:llc-over-time}%
        \textbf{Estimated LLC over time.} Following \citet{icl1}, we apply the LLC estimator to transformer checkpoints from each model's training run. We plot the result, indicating $\tcrit[M]$ (the step of minimum OOD loss) with a circle.
        Apart from very small $M$, the estimates follow a broadly similar trend to the $M=\infty$ case shown in yellow (cf., also, \cref{fig:hoogland}), with deviations for some intermediate $M$.
        See \cref{fig:llc_grid_per_M} for individual lines.
    }
\end{figure}

\Cref{fig:llc-over-time} shows the results of LLC estimation over time for our models.
As did \citet{icl1}, we observe that the LLC trends for our transformers can be characterized as a series of LLC changes (gradual increases or decreases) that are punctuated by brief plateaus.
Apart from very small $M$, all models have an early plateau, perhaps corresponding to learning an extremely simple context-independent predictor before moving on to more advanced predictors. All models also undergo a parallel decrease in LLC towards the end of training, possibly coinciding with a similar ``collapse'' of internal weights as studied by \citet{icl1}.
After that, for high $M$, the curves follow the development of the $M=\infty$ case shown in yellow (cf., also, \cref{fig:hoogland}) for the remainder of training.

The interesting trends are in the deviation for intermediate $M$ values from this $M=\infty$ baseline.
For some early intermediate $M$ values, at around the time $\tcrit[M]$ of minimum OOD loss, the LLC ``spikes.''
For these $M$, there is a neat interpretation of this increase as corresponding to a transition between a low-complexity proto-ridge solution and a high-complexity proto-\dMMSE[M] solution, following the same explanation assumed to cover the entirety of training in the main text.

For the later-intermediate $M$ values, these ``spikes'' in LLC appear less pronounced, or not at all (though the LLC still appears to be elevated with respect to the $M=\infty$ baseline). The final LLC still ends up at elevated compared to the final LLC from the $M=\infty$ case. However, there is no clear sign of a transition from a proto-ridge solution to a proto-\dMMSE[M] solution with an increased LLC to explain this.
In these cases, we note that $\tcrit[M]$ happens close to or within the ``collapse'' stages that affect a general downturn in the LLC for all $M$. It is possible that any competition between ridge and \dMMSE[M] as it effects the LLC is obscured from our analysis by other developments in the model.
Future work could attempt to understand and then prevent this collapse and study the within-training complexity dynamics more clearly.

\clearpage
\section{Why does transience stop?}
\label{appendix:end-of-retreat}

If the development of our transformers were to be governed exactly by Bayesian internal model selection, with increased training corresponding to increased samples, then the explanation predicts that for \emph{all} $M$, a higher-complexity but lower-loss solution should \emph{eventually} be preferred.
Instead, we see a \emph{task diversity threshold} \citetext{as originally observed by \citealp{raventós2023pretraining}} at which transience \emph{ends.}
In this section, we briefly consider several possible explanations for this phenomenon.
\subsection{Model capacity constraints}

We note that a \dMMSE[M] solution can only compete with the ridge solution if it is a faithful enough approximation of \dMMSE[M] that it actually achieves lower loss than the ridge solution.
There is certainly some $M$ sufficiently large such that the best realizable approximation of \dMMSE[M] for our architecture is insufficiently competitive, at which point, transience will end.
We note that \citet{raventós2023pretraining} and \citet{nguyen2024differential} draw similar conclusions about the potential role of model capacity in explaining the task diversity threshold.

In our case, this would require \dMMSE[M] to be unrealizable by $M=256$ or so. We do not rule this out, though our primary architecture has around 2.65 million parameters. We note that smaller model architectures show slightly lower task diversity thresholds (\cref{appendix:architecture-comparison}), indicating that model capacity may be somewhat involved.
    
\subsection{Under-training}

Bayesian internal model selection predicts that, under the stated assumptions, there will be \emph{some} amount of data $n$ beyond which the lower loss but more complex solution will be preferred, but the lower complexity will be preferred for all lower $n$. By analogy, it is possible that we have not trained our transformers for enough steps to incentivize them to specialize.

We do observe that the critical time $\tcrit[M]$ (the step of minimum OOD loss) increases with $M$ (see \cref{appendix:tcrit}). However, we note that the critical time is not near the end of training around the task diversity threshold.
        
\subsection{Neuroplasticity loss}

It's also possible that not all time-steps contribute equally to the ``effective amount of samples'' driving the tradeoff. In particular, it's possible that during the training of our transformers there is a brief window of opportunity or ``critical period'' during which they are governed by principles similar to Bayesian internal model selection, and after which they are either governed by similar principles on a slower timescale, or are resistant to further developments.

Some preliminary support for this hypothesis in our case can be seen in the LLC-over-training analysis provided in \cref{appendix:clean-v-messy}, wherein we compare the development of complexity in our model to the detailed analysis of the $M=\infty$ case conducted by \citet{icl1}. We observe that the task diversity threshold coincides very roughly with the overlap between the progressively increasing $\tcrit[M]$ and a general downturn in LLC estimates that affects all models. \citet{icl1} observed a similar LLC downturn at a similar stage of training and, analyzing the internals of their transformer (with smaller but somewhat similar architecture to ours) they found that many of the weights in various internal modules of the transformer ``collapse'' during this downturn, which could make further development difficult.
        
\subsection{Non-Bayesian pre-training}

We note that \citet{raventós2023pretraining} and \citet{panwar2024bayesianprism} cast the task diversity threshold as an example of ``non-Bayesian in-context learning,'' due to the failure of the transformer to adopt the optimal Bayesian solution derived from the pre-training data distribution for in-context learning.
In contrast, the perspective of Bayesian internal model selection casts the task diversity threshold as an interesting example of ``non-Bayesian pre-training,'' in which the transformer fails to learn following the dynamics that would coincide with the evolution of the Bayesian posterior with increasing data.

As we have discussed, it is unclear to what extent neural network development \emph{is} governed by principles similar to Bayesian internal model selection, which is a formal mathematical model of Bayesian inference in neural networks, not of stochastic gradient-based optimization.
It is plausible based on the perspective of nonlinear dynamics that degeneracy (such as measured by the LLC) should play a role in governing system trajectories during stochastic gradient-based optimization, and our work contributes empirical evidence that this is the case.
However, there are still various gaps in our understanding of the principles governing the development of neural networks.
The true ``dynamic internal model selection'' principles, once discovered, may reveal a natural explanation for the end of transience, leaving nothing further to be explained.

\clearpage
\section{Experiments with additional architectures}
\label{appendix:architecture-comparison}

\begin{figure}
    \centering
    \includegraphics[width=1.0\linewidth]{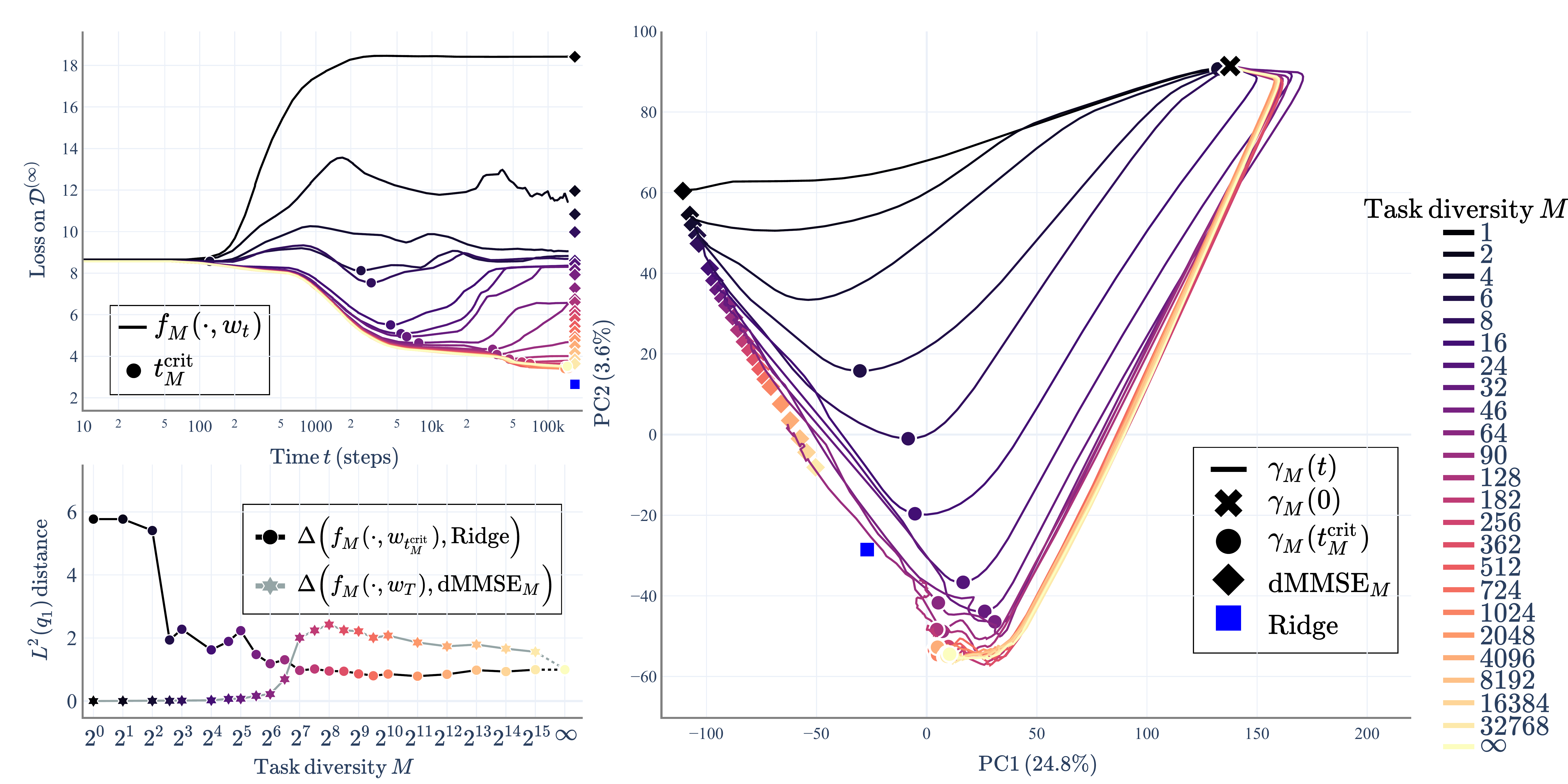}
    \caption{Reproduction of \cref{fig:ood-and-pca} for architecture with $L=2$, $H=1$, $d_{\text{embed}}=d_{\text{mlp}}=128$.}
    \label{fig:l2h1-de128-pca-ood}
\end{figure}

\begin{figure}
    \centering
    \includegraphics[width=1.0\linewidth]{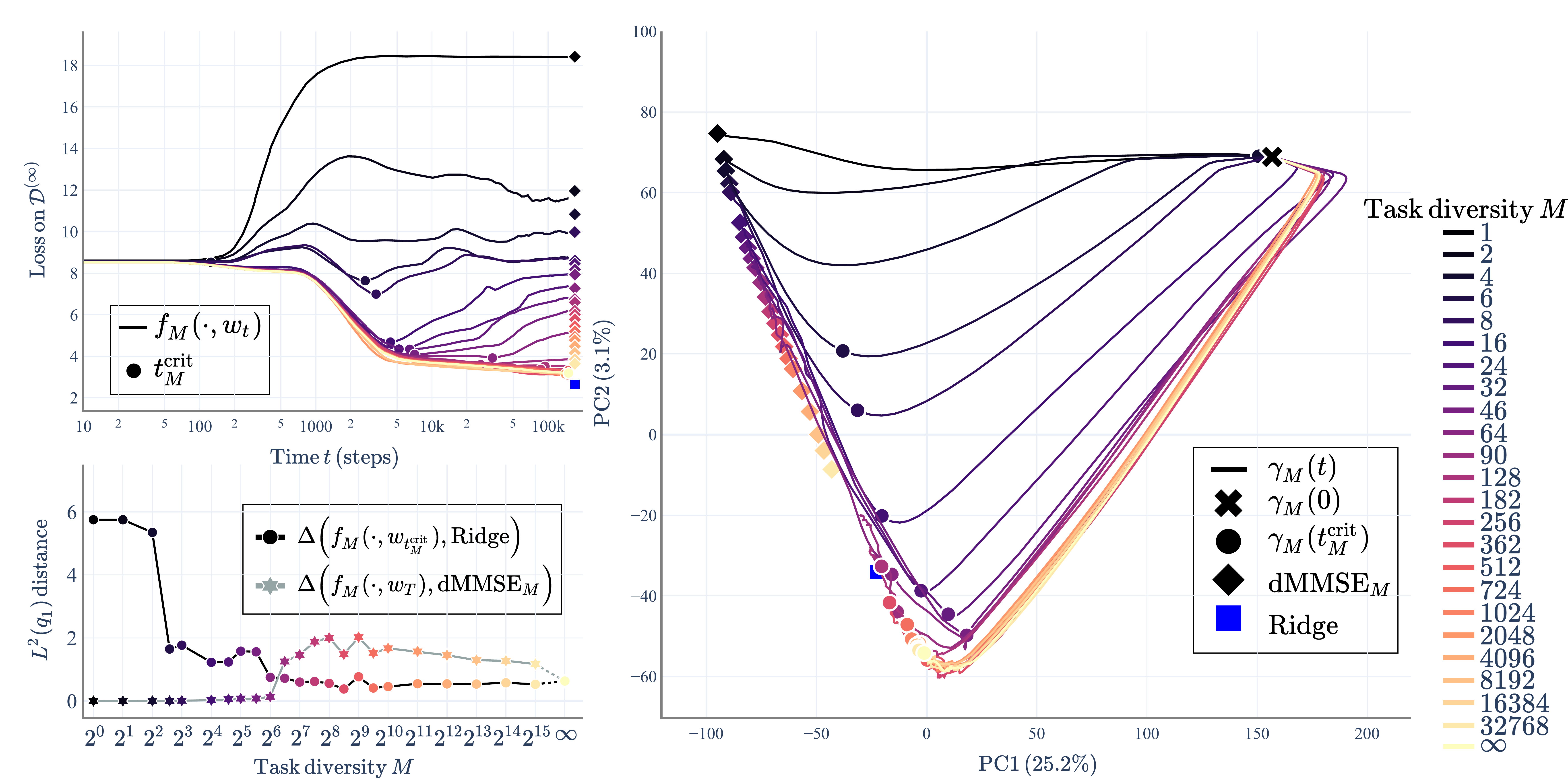}
    \caption{Reproduction of \cref{fig:ood-and-pca} for architecture with $L=2$, $H=4$, $d_{\text{embed}}=d_{\text{mlp}}=128$.}
    \label{fig:l2h4-de128-pca-ood}
\end{figure}

\begin{figure}
    \centering
    \includegraphics[width=1.0\linewidth]{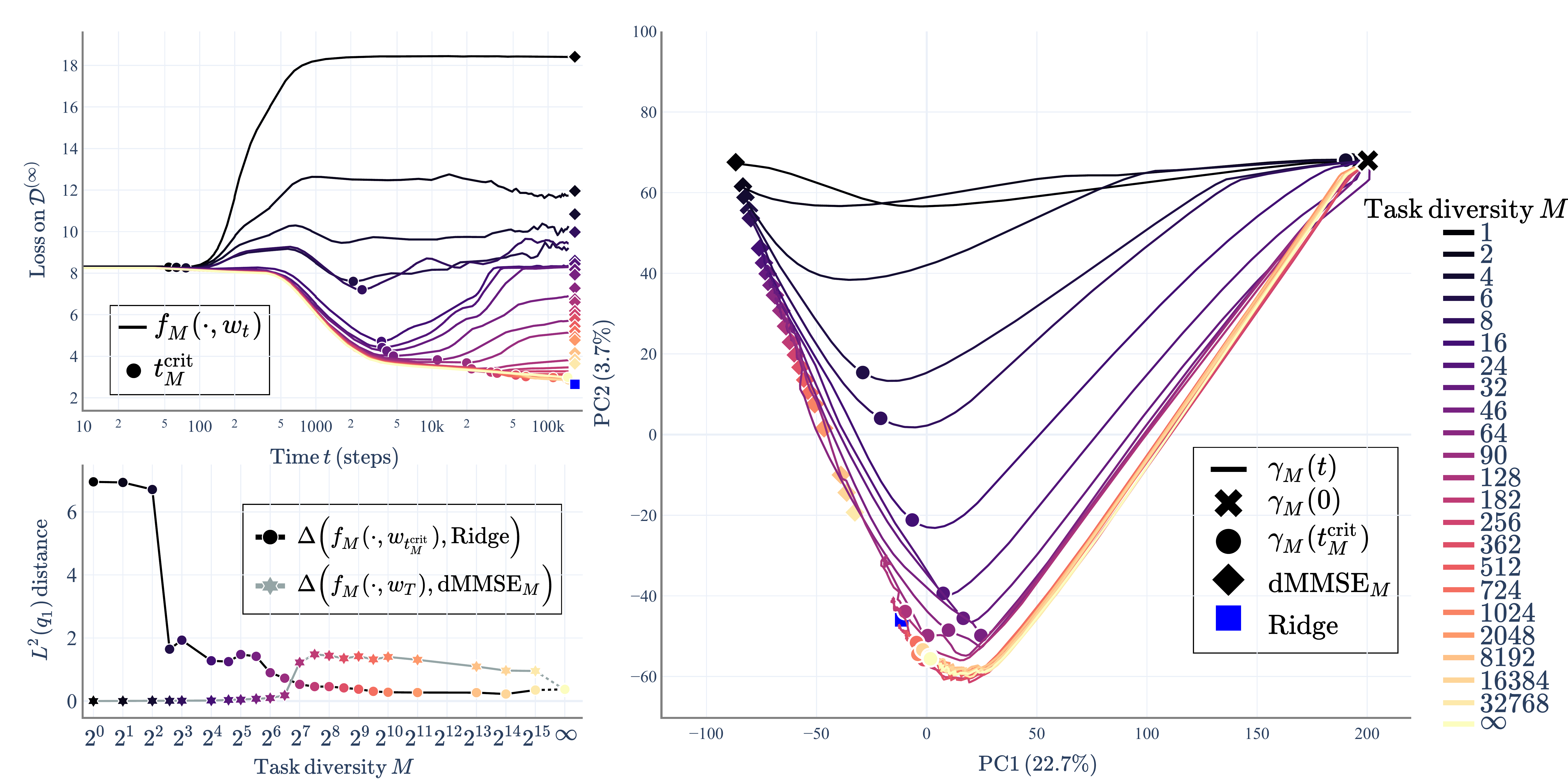}
    \caption{Reproduction of \cref{fig:ood-and-pca} for architecture with $L=2$, $H=4$, $d_{\text{embed}}=d_{\text{mlp}}=256$.}
    \label{fig:l2h4-de256-pca-ood}
\end{figure}

\begin{figure}
    \centering
    \includegraphics[width=1.0\linewidth]{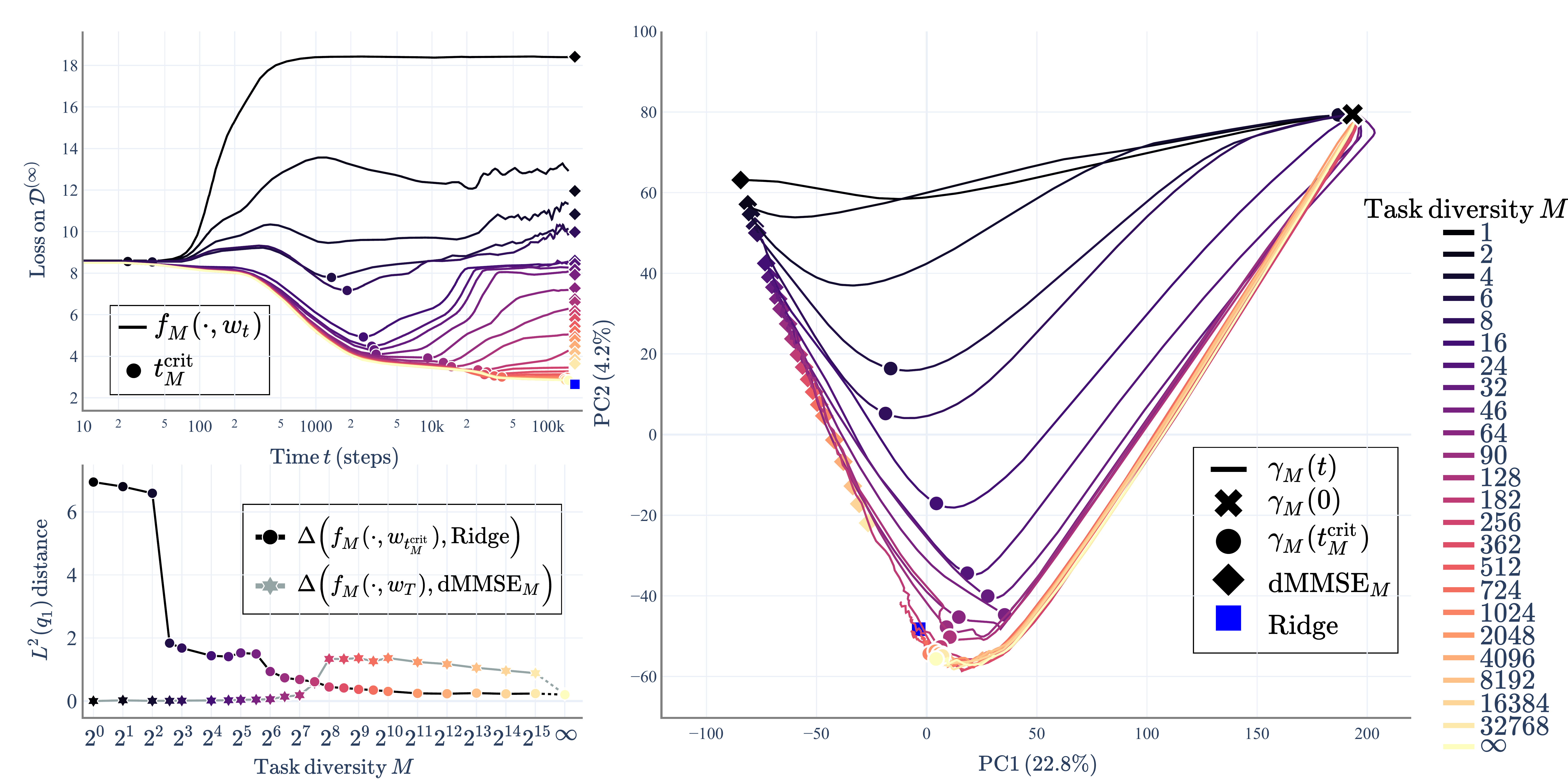}
    \caption{Reproduction of \cref{fig:ood-and-pca} for architecture with $L=2$, $H=4$, $d_{\text{embed}}=d_{\text{mlp}}=512$ (primary architecture).}
    \label{fig:l2h4-de512-pca-ood}
\end{figure}

\begin{figure}
    \centering
    \includegraphics[width=1.0\linewidth]{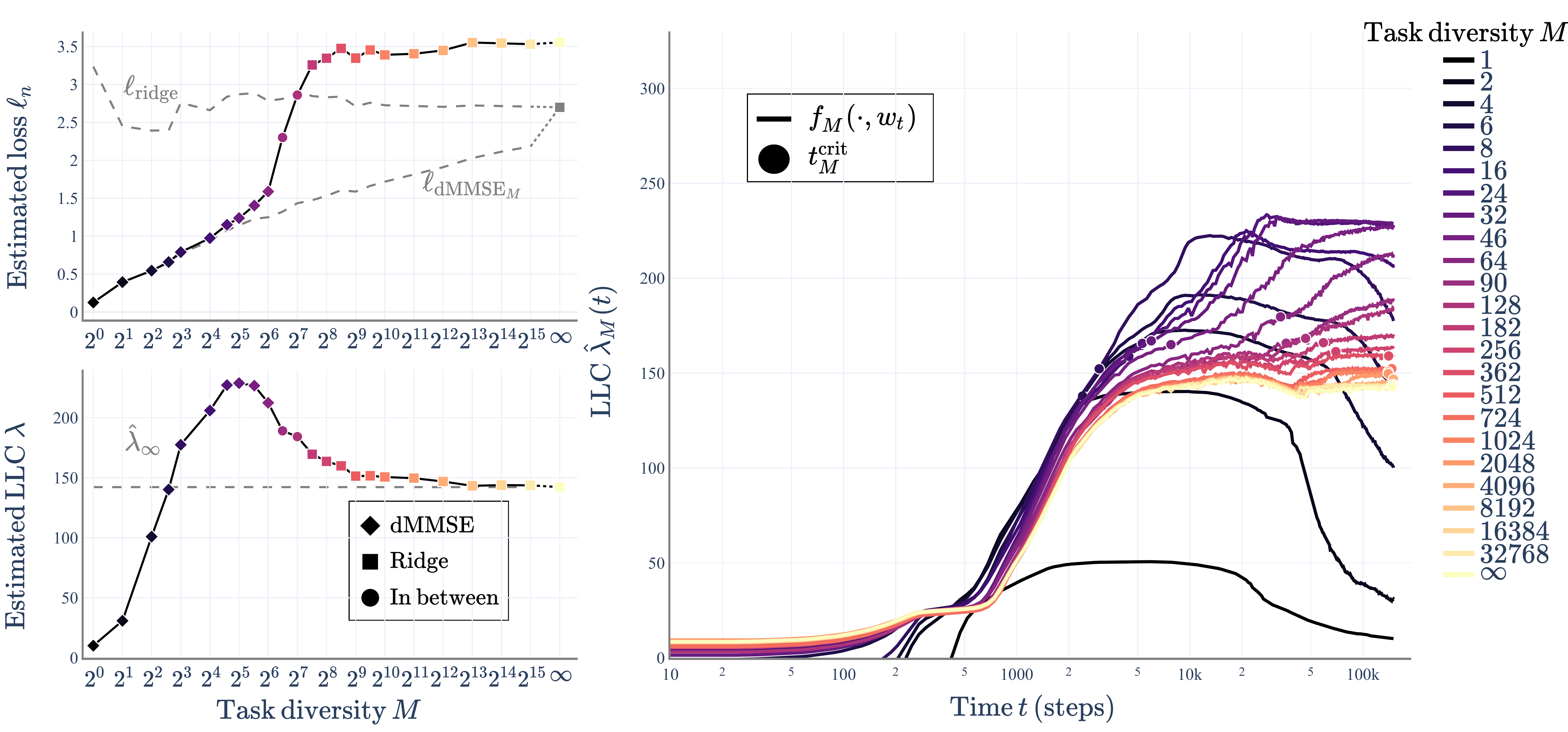}
    \caption{SLT results for architecture with $L=2$, $H=1$, $d_{\text{embed}}=d_{\text{mlp}}=128$.}
    \label{fig:l2h1-de128-slt}
\end{figure}

\begin{figure}
    \centering
    \includegraphics[width=1.0\linewidth]{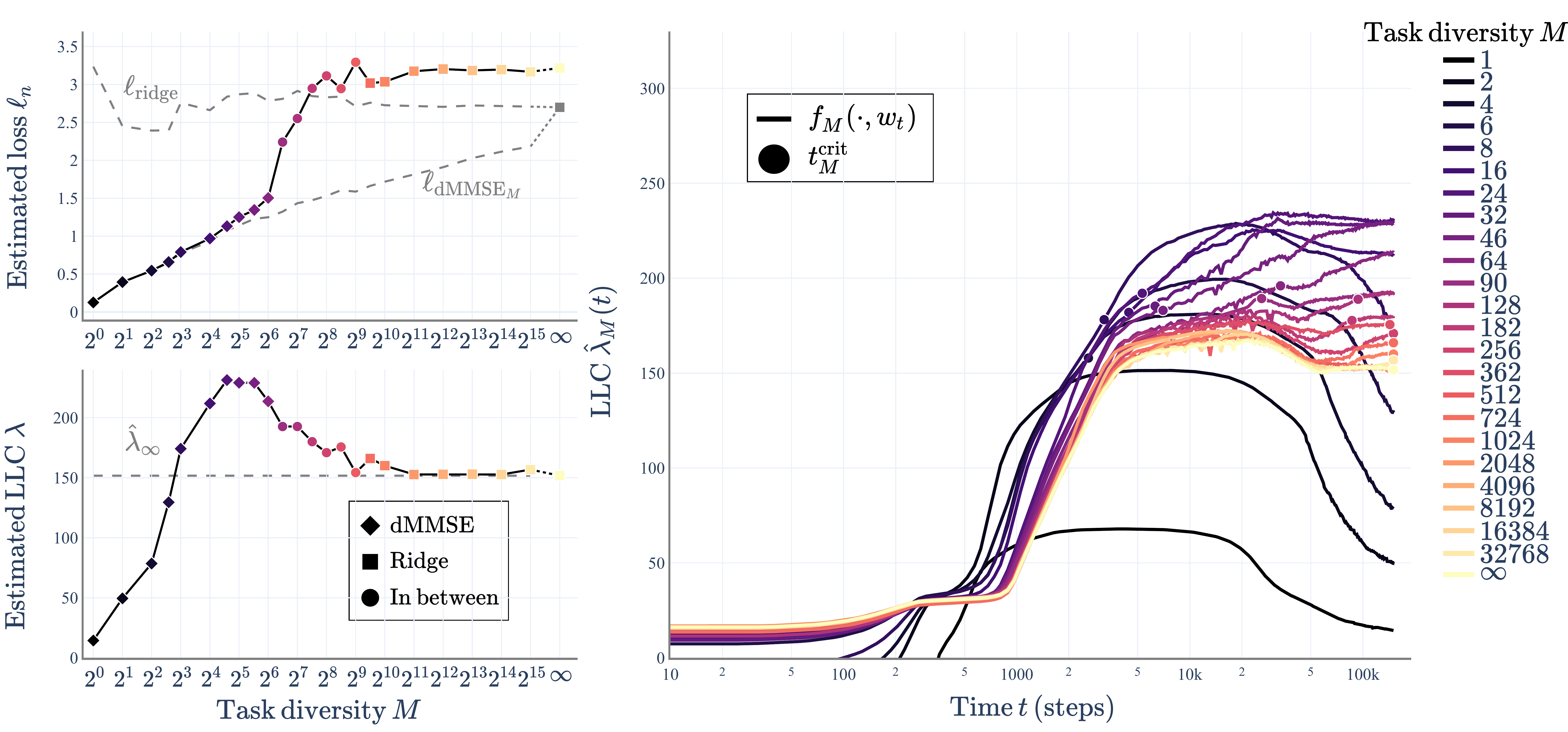}
    \caption{SLT results for architecture with $L=2$, $H=4$, $d_{\text{embed}}=d_{\text{mlp}}=128$.}
    \label{fig:l2h4-de128-slt}
\end{figure}

\begin{figure}
    \centering
    \includegraphics[width=1.0\linewidth]{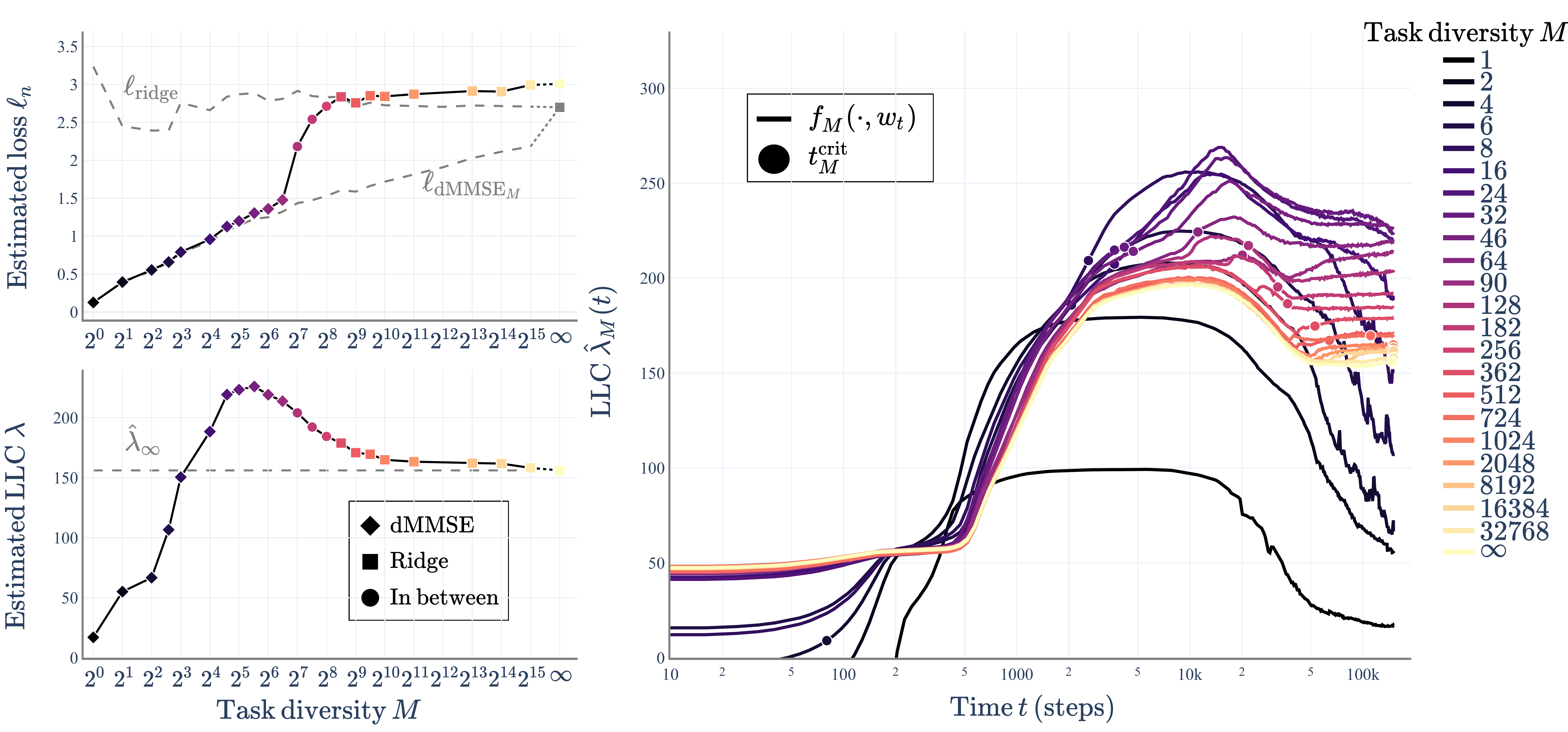}
    \caption{SLT results for architecture with $L=2$, $H=4$, $d_{\text{embed}}=d_{\text{mlp}}=256$.}
    \label{fig:l2h4-de256-slt}
\end{figure}

\begin{figure}
    \centering
    \includegraphics[width=1.0\linewidth]{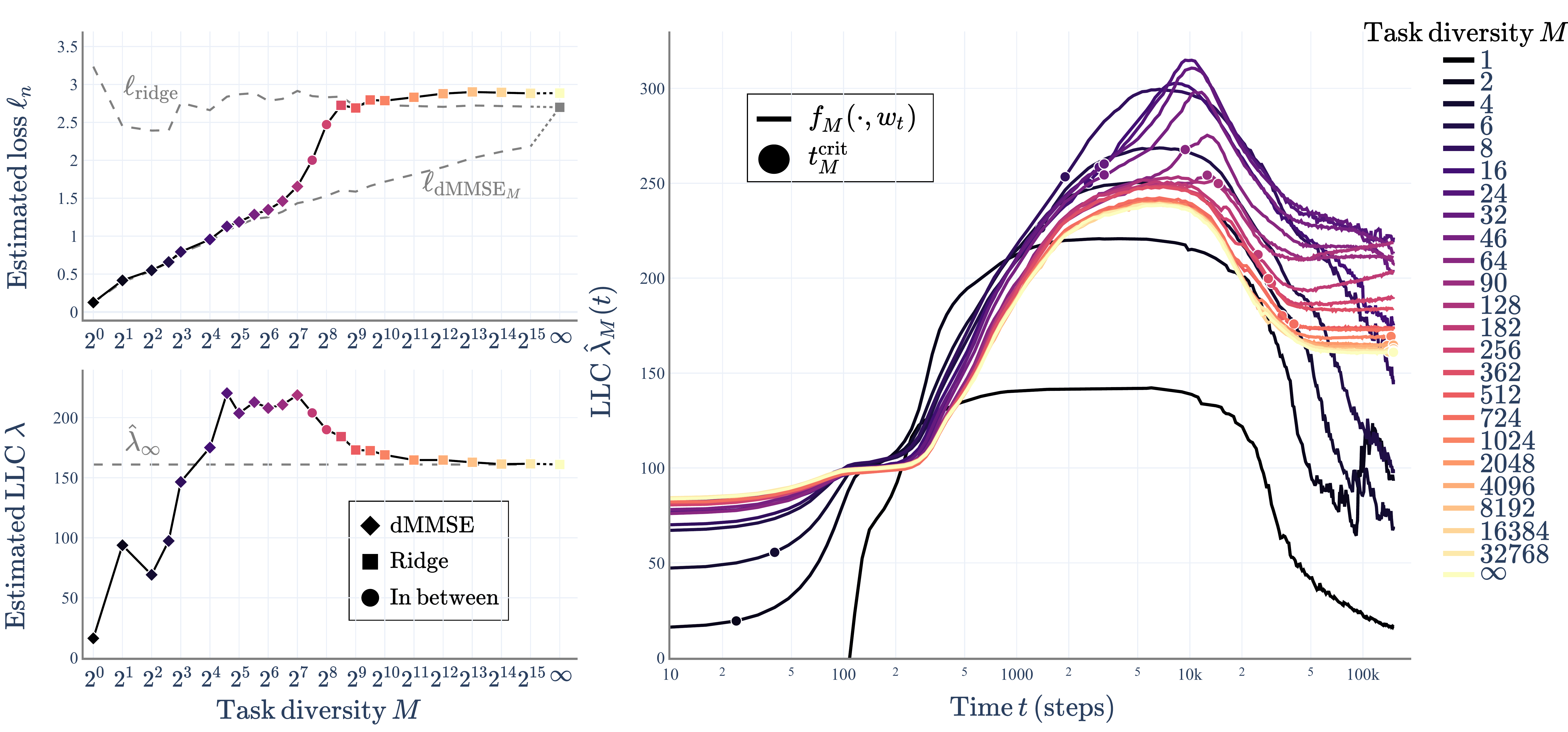}
    \caption{SLT results for architecture with $L=2$, $H=4$, $d_{\text{embed}}=d_{\text{mlp}}=512$ (primary architecture).}
    \label{fig:l2h4-de512-slt}
\end{figure}

\Cref{appendix:lr-architecture} details the architecture hyperparameters used for the primary architecture used in our main analysis. In this appendix, we report OOD loss, joint trajectory PCA, loss and LLC estimates and LLC-over-time analysis (see \cref{appendix:clean-v-messy}) for three additional transformer architectures (with smaller numbers of heads and/or smaller embedding/ML dimensions). The three additional architectures we study are as follows:
\begin{enumerate}
    \item
        $L=2$ layers, $H = 1$ head, $d_{\text{embed}}=d_{\text{mlp}}=128$ dimensional embeddings.
        The OOD loss and trajectory PCA results are in
            \cref{fig:l2h1-de128-pca-ood}.
        The loss, LLC, and LLC-over-time results are in
            \cref{fig:l2h1-de128-slt}.

    \item 
        $L=2$ layers, $H = 4$ heads, $d_{\text{embed}}=d_{\text{mlp}}=128$ dimensional embeddings.
        The OOD loss and trajectory PCA results are in
            \cref{fig:l2h4-de128-pca-ood}.
        The loss, LLC, and LLC-over-time results are in
            \cref{fig:l2h4-de128-slt}.
        
    \item 
        $L=2$ layers, $H = 4$ heads, $d_{\text{embed}}=d_{\text{mlp}}=256$ dimensional embeddings.
        The OOD loss and trajectory PCA results are in
            \cref{fig:l2h4-de256-pca-ood}.
        The loss, LLC, and LLC-over-time results are in
            \cref{fig:l2h4-de256-slt}.
\end{enumerate}
For comparison, we also reproduce \cref{fig:ood-and-pca,fig:clean-explanation,fig:llc-over-time} for the primary architecture with $L=2$ layers, $H = 4$ heads, $d_{\text{embed}}=d_{\text{mlp}}=512$ dimensional embeddings.
The OOD loss and trajectory PCA results are in
    \cref{fig:l2h4-de512-pca-ood}.
The loss, LLC, and LLC-over-time results are in
    \cref{fig:l2h4-de512-slt}.

\begin{figure}
    \centering
    \includegraphics[width=1.0\linewidth]{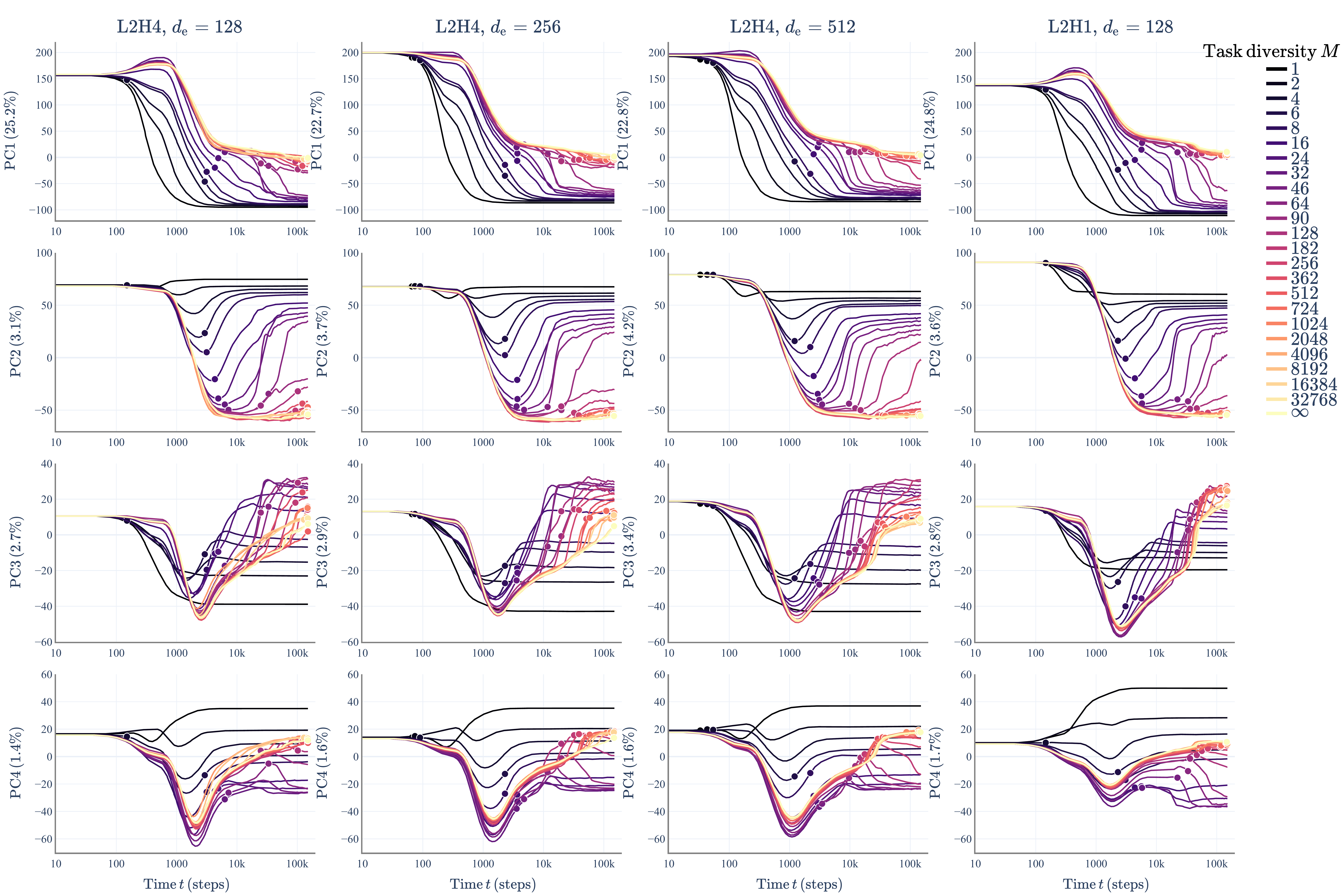}
    \caption{PCs over time for each of the four architectures, displaying remarkably similar behavior.}
    \label{fig:pc_over_time_architectures}
\end{figure}

We provide a comparison of the PC over time curves for each architecture in \cref{fig:pc_over_time_architectures}.

\begin{figure}
    \centering
    \begin{subfigure}[c]{0.48\textwidth}
        \centering
        \includegraphics[width=\textwidth]{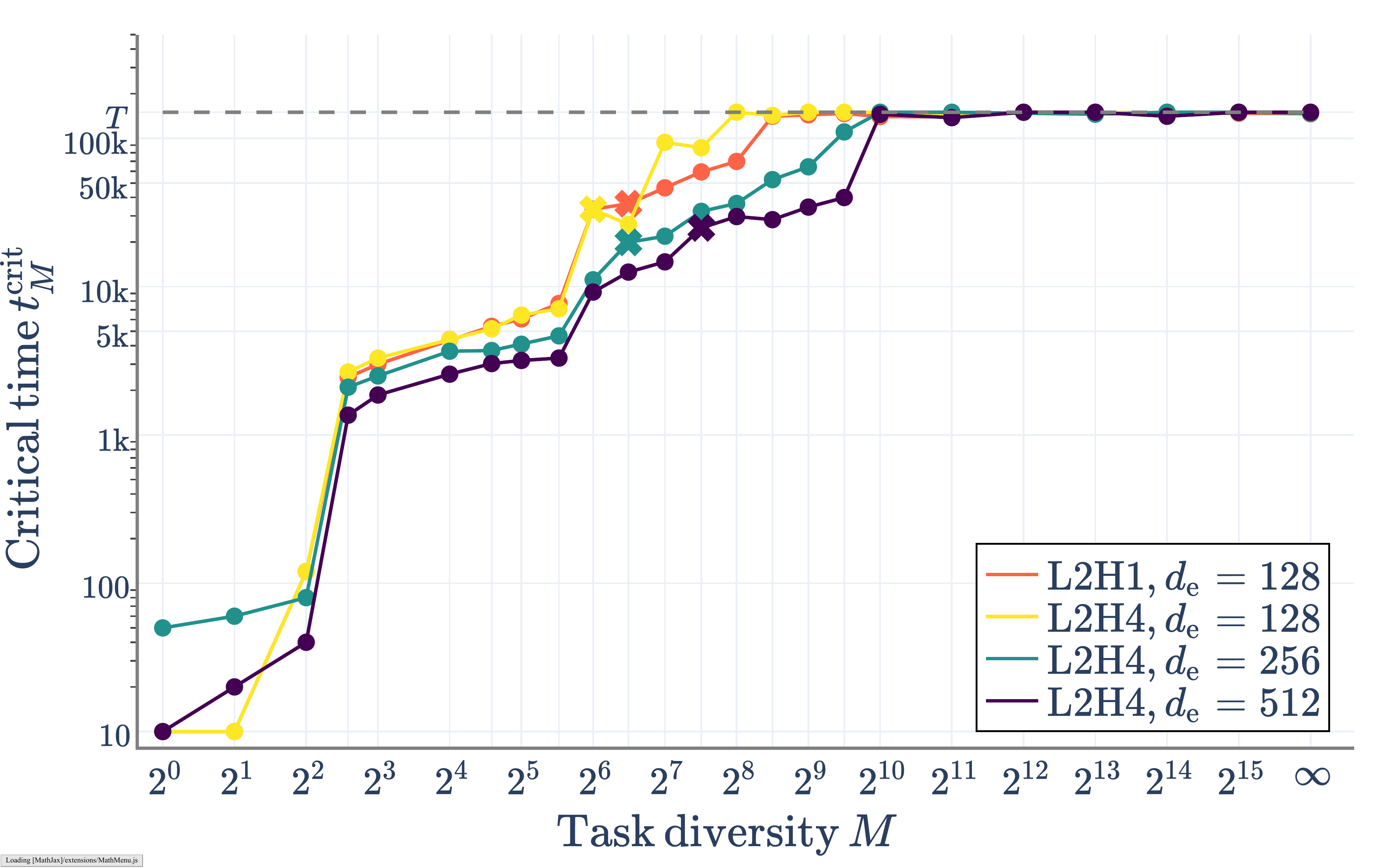}
        \caption{Comparison of $\tcrit[M]$ at each $M$ for each of the four architectures. The cross marks the largest $M$ value for which we determine the transient ridge phenomenon to occur for each architecture. Notice for the range of $M$ for which transient ridge occurs, $\tcrit[M]$ is increasing in $M$, though with different trends for different regions of $M$.}
        \label{fig:tcrit_comparison}
    \end{subfigure}
    \hfill
    \begin{subfigure}[c]{0.48\textwidth}
        \centering
        \includegraphics[width=\textwidth]{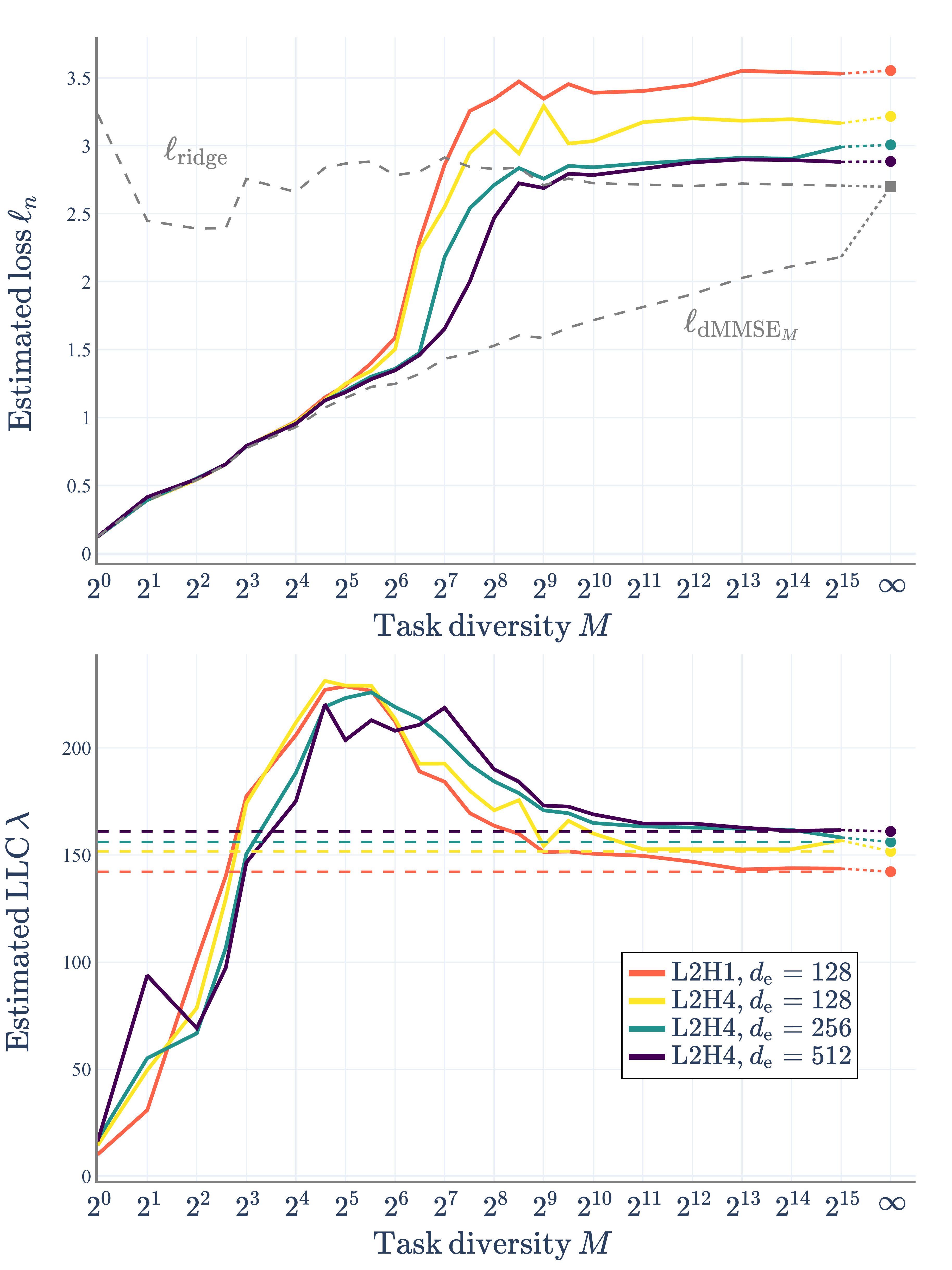}
        \caption{Comparison of final loss and final LLC for each architecture at each $M$. This shows the qualitative similarities across architectures, with the task diversity threshold increasing with model size. The dashed lines on the LLC plot represent $\hat{\lambda}_{\infty}$, the final trained LLC of the $M=\infty$ model, for each architecture.}
        \label{fig:final_loss_llc_all_archs}
    \end{subfigure}
    \caption{Comparisons between architectures.}
    \label{fig:combined_architecture_comparison}
\end{figure}

In \cref{fig:tcrit_comparison}, we plot $\tcrit[M]$ values (\cref{appendix:tcrit}) for all $M$ for all four architectures. We see that within each architecture, $\tcrit[M]$ increases with $M$. In \cref{fig:final_loss_llc_all_archs} we plot the final loss and final LLC of all architectures for comparison, which shows that the task diversity threshold increases slightly with architecture size. All four architectures exhibit qualitatively similar behaviors as shown by these two figures. 

\begin{figure}
    \centering
    \includegraphics[width=1.0\linewidth]{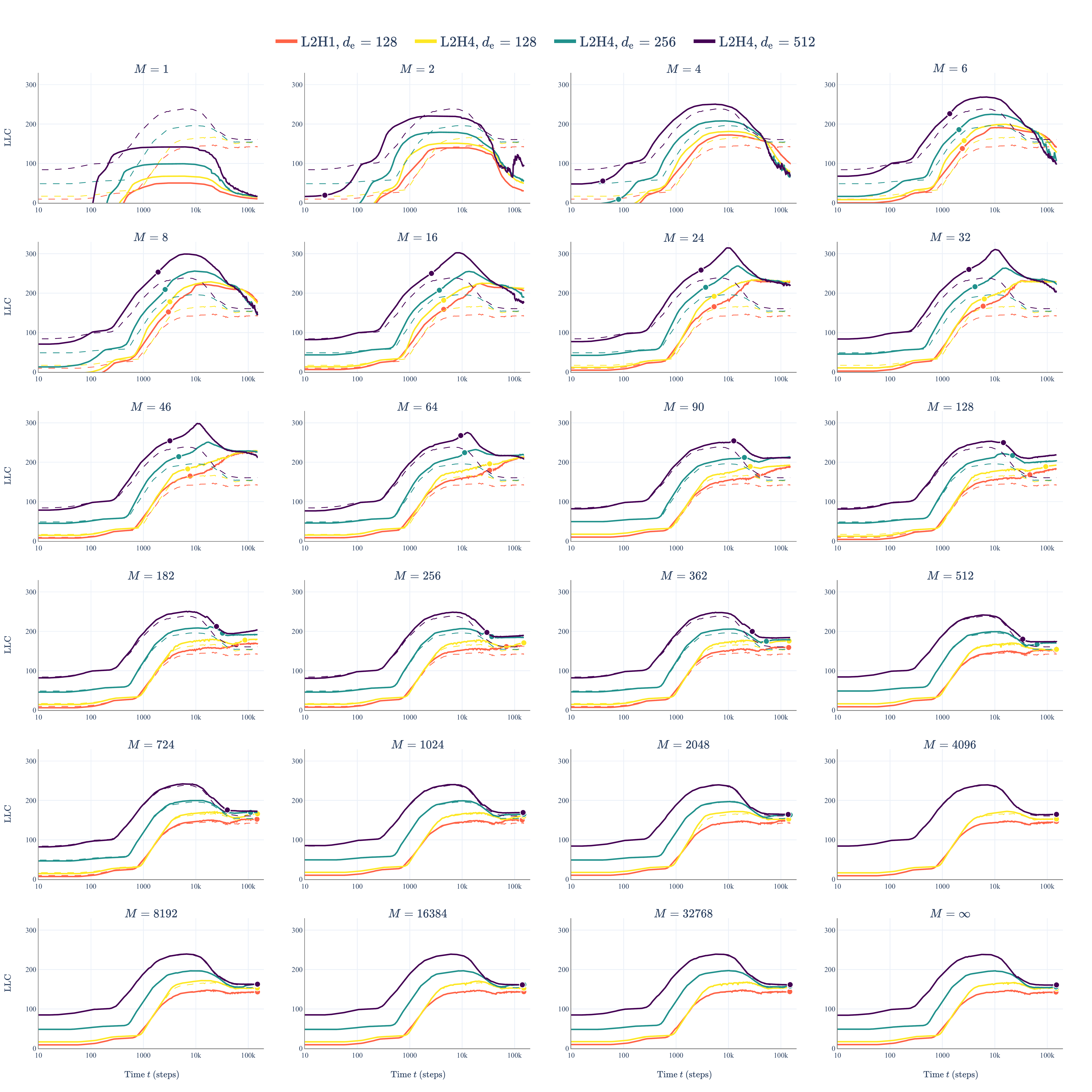}
    \caption{Comparison of LLC over time curves between architectures, where each subplot corresponds to a particular $M$ value. The dashed lines represent the $M=\infty$ baseline for each architecture. Circle represents $\tcrit[M]$.}
    \label{fig:llc_grid_per_M}
\end{figure}

In \cref{fig:llc_grid_per_M}, we plot LLC over time curves for easy comparison between architectures for each fixed $M$, as well as the respective ``baseline" of the $M=\infty$ model. While we mostly see similar behavior across architectures, there are subtle differences visible in these complexity dynamics, suggesting that specific architectural choices may influence the developmental trajectory of these models in meaningful ways.

\clearpage
\clearpage

\stopcontents[appendix]

\end{document}